\documentclass[lettersize,journal]{IEEEtran}

\usepackage{amsmath,amsfonts}
\usepackage{algorithmic}
\usepackage{algorithm}
\usepackage{array}
\usepackage[caption=false,font=normalsize,labelfont=sf,textfont=sf]{subfig}

\usepackage{caption}
\usepackage{textcomp}
\usepackage{stfloats}
\usepackage{url}
\usepackage{verbatim}
\usepackage{graphicx}

\usepackage{cite}
\usepackage{hyperref}
\usepackage{url}

\usepackage{amsmath}
\usepackage{amssymb}
\usepackage{mathtools}
\usepackage{amsthm}
\usepackage{bm}
\usepackage{bbm}

\newtheorem{prop}{Proposition}

\newtheorem{cor}{Corollary}

\theoremstyle{definition}

\newtheorem{defn}{Definition}

\newcommand{\qidx}{i}
\newcommand{\yidx}{j}
\newcommand{\imgidx}{l}

\newcommand{\cA}{\mathcal{A}}
\newcommand{\cC}{\mathcal{C}}
\newcommand{\cD}{\mathcal{Y}}

\newcommand{\cX}{\mathcal{X}}

\newcommand{\cQ}{\mathcal{Q}}

\newcommand{\defined}{\triangleq}

\usepackage{color}

\newcommand{\approach}{\mathsf{ContrastiveConf}}
\newcommand{\approachpos}{\mathsf{Conf}^+}
\newcommand{\approachneg}{\mathsf{Conf}^-}

\newtheorem{assumption}{Assumption}

\newcommand{\Ltotal}{\mathcal{L}_{\text{total}}}
\newcommand{\Lcls}{\mathcal{L}_{\text{cls}}}
\newcommand{\entropy}{\mathcal{H}}

\newcommand{\Lbox}{\mathcal{L}_{\text{box}}}
\newcommand{\cset}{\mathcal{C}}
\newcommand{\optmatch}{\sigma}

\usepackage{enumitem}
\usepackage{booktabs}
\usepackage{makecell}
\usepackage{multicol}
\usepackage{multirow}
\usepackage{courier}

\usepackage{tcolorbox}
\tcbuselibrary{skins}

\usepackage{soul}
\newenvironment{revision}{\color{black}}{}
\newenvironment{revision2}{\color{black}}{}
\newcommand{\rev}[1]{\textcolor{black}{#1}}
\DeclareRobustCommand{\revv}[1]{\textcolor{black}{#1}}
\DeclareRobustCommand{\revvv}[1]{\textcolor{black}{#1}}

\newcommand{\statement}[1]{\medskip\noindent
  \textcolor{black}{\textbf{#1}}\space
}

\newcommand{\citep}[1]{\cite{#1}}
\newcommand{\citet}[1]{\cite{#1}}

\usepackage[dvipsnames]{xcolor}
\usepackage{colortbl}
\definecolor{bestgreen}{rgb}{0.8,0.9,0.8} %
\definecolor{bestblue}{rgb}{0.678,0.847,0.902} %
\definecolor{lightgray}{rgb}{0.9,0.9,0.9}

\usepackage{comment}

\usepackage[switch]{lineno}

\usepackage{fancyhdr}
\fancypagestyle{appendixstyle}{%
    \fancyhf{}
    \fancyhead[R]{\thepage}
    
}

\title
{Uncertainty Quantification in Detection Transformers: Object-Level Calibration and Image-Level Reliability}

\author{Young-Jin Park\IEEEauthorrefmark{1}, Carson Sobolewski\IEEEauthorrefmark{1}, \textnormal{and} Navid Azizan \thanks{$^*$Contributed equally and share co-first authorship.}\\
Massachusetts Institute of Technology\\
{\tt\small \{youngp,csobo,azizan\}@mit.edu}
}

\begin{document}
\maketitle

\begin{abstract}
DEtection TRansformer (DETR) and its variants have emerged as promising architectures for object detection, offering an end-to-end prediction pipeline. In practice, however, DETRs generate hundreds of predictions that far outnumber the actual objects present in an image. \rev{This raises a critical question: which of these predictions could be trusted?
This is particularly important for safety-critical applications, such as in autonomous vehicles.}
Addressing this concern, we provide empirical and \rev{theoretical} evidence that predictions within the same image play distinct roles, resulting in varying reliability levels. \begin{revision}Our analysis reveals that DETRs employ an optimal ``specialist strategy'': one prediction per object is trained to be well-calibrated%
, while the remaining predictions %
are trained to suppress their foreground confidence to near zero, even when maintaining accurate localization. We show that this strategy emerges as the loss-minimizing solution to the Hungarian matching algorithm, fundamentally shaping DETRs' outputs.
While selecting the well-calibrated predictions is ideal, they are unidentifiable at inference time.    
This means that any post-processing algorithm---used to identify trustworthy predictions---poses a risk of outputting a set of predictions with mixed calibration levels.
Therefore, practical deployment necessitates a joint evaluation of both the model's calibration quality and the effectiveness of the post-processing algorithm.
However, we demonstrate that existing metrics like average precision and expected calibration error are inadequate for this task.
To address this issue, we further introduce Object-level Calibration Error (OCE), which evaluates calibration by aggregating predictions per ground-truth object rather than per prediction. This object-centric design penalizes both retaining suppressed predictions and missed ground truth foreground objects\end{revision}, making OCE suitable for both evaluating models and identifying reliable prediction subsets. Finally, we present a post hoc uncertainty quantification (UQ) framework that predicts per-image model accuracy.
\revvv{The code is available at: \url{https://github.com/azizanlab/uq-detr}.}
\end{abstract}

\section{Introduction} \label{sec:intro}

Object detection is an essential task in computer vision, with applications that span various domains including autonomous driving, warehousing, and medical image analysis.
Existing object detection methods predominantly utilize convolutional neural networks (CNNs) \citep{girshick2014rich, ren2017faster, redmon2016you, he2020mask, sun2023sparse, cai2021cascade} to identify and locate objects within images.
More recently, DEtection TRansformer (DETR) \cite{carion2020end} has revolutionized the field by utilizing a Transformer encoder-decoder architecture to offer a scalable end-to-end prediction pipeline, where the model predicts a set of bounding boxes and class probabilities.
This paradigm shift has led to the exploration of various DETR variants, positioning them as potential foundation models for object detection tasks.
While notable progress has been made, the reliability of these predictions remains under-investigated.

\rev{Quantifying the reliability of these models' predictions is critical across different scenarios.
For instance, when building an auto-labeling system using DETRs, it is crucial to consider that the model may not always provide accurate predictions.
Images where the model's reliability is questionable may need to be reviewed by human labelers.
In safety-critical applications, such as autonomous vehicles, knowledge of the reliability of a prediction in the perception pipeline for a given scene is crucial for the downstream decision-making. Without properly understanding this reliability, the system may, for example, overlook a pedestrian, leading to catastrophically wrong decisions.
However, owing to DETR’s unique set-prediction mechanism, quantifying its reliability is not straightforward and remains largely underexplored.}

In particular, DETR outputs a fixed number of predictions, typically in the hundreds; consequently, the central concern is \emph{which predictions can be trusted and used}.
Practitioners often employ heuristic approaches to select a subset of predictions, such as by setting a user-defined threshold to retain only a small number of high-confidence (e.g., $0.7$) outputs, as seen in the official demo~\citep{detr_demo}. 
Similarly, previous studies exploring model reliability, such as \citep{kuppers2020multivariate,munir2022towards,munir2023bridging,pathiraja2023multiclass,munir2024cal,munir2023domain}, have also applied a user-defined confidence threshold (e.g., 0.3) to retain a subset of predictions for evaluating calibration quality, rather than using the entire set.
On the other hand, the published implementations of several DETR variants select the top-$k$ outputs (e.g., 100 out of 300 for Deformable-DETR \citep{zhu2020deformable} and 300 out of 900 for DINO \citep{zhang2022dino}) based on confidence scores.
Nonetheless, how different subset-selection schemes affect model reliability, as well as the process of choosing appropriate configurations for each scheme, remains under-explored, and their significance has yet to be fully understood.

\begin{revision}
To address this gap and move beyond heuristic subset selection, 
we first seek a fundamental characterization of DETRs' predictions based on the training objective used.
Our theoretical analysis shows that the Hungarian loss incentivizes a ``specialist strategy'': having a single well-calibrated prediction (which we refer to as the \emph{primary} prediction) to represent each ground-truth object, while suppressing all remaining predictions (referred to as the \emph{secondary} predictions).
This specialist strategy presents a fundamental challenge for practical deployment: while those primary predictions are trained to be reliable, they are unidentifiable at test time. 
This means that any post-processing algorithm poses a risk of outputting a set of predictions with mixed calibration levels.
Consequently, evaluating model reliability requires jointly assessing the model's calibration quality and the effectiveness of the post-processing algorithm in recovering the well-calibrated subset. 

To facilitate this joint evaluation, we introduce \textbf{object-level calibration error (OCE)}, which evaluates calibration along ground-truth \emph{objects} rather than \emph{predictions} (Section~\ref{sec:uq_obj}).
OCE penalizes both missing ground truth objects and retaining unreliable secondary predictions, enabling proper assessment of model + post-processing combinations, something that common existing metrics are fundamentally incapable of.
For instance, average precision (AP) \citep{salton1983introduction, everingham2010pascal,lin2014microsoft} favors retaining all predictions, while expected calibration error (ECE) variants \citep{kuppers2020multivariate, oksuz2023towards, kuzucu2024calibration} favor discarding nearly all predictions to achieve artificially low error by ignoring missed objects.

Finally, we leverage both our theoretical characterization of DETRs' specialist strategy as well as OCE's ability to identify primary and secondary predictions and develop a method for \textbf{image-level uncertainty quantification}. 
More specifically, we first demonstrate that the foreground confidence contrast between primary and secondary predictions is strongly correlated with image-level reliability.
This motivates our novel framework for quantifying image-level reliability by measuring the confidence contrast between the ``positive predictions'' retained by an OCE post-processing and the remaining ``negative'' ones.
We conduct numerical experiments across in-distribution, near out-of-distribution, and far out-of-distribution scenarios, and show the effectiveness of our approach (Section~\ref{sec:uq}).

In short, our key contributions are:
\begin{itemize}[leftmargin=10pt]
    \setlength\itemsep{0.25em}
    \item We theoretically characterize and empirically validate the optimal strategy that DETR takes to minimize the Hungarian loss: it creates a single well-calibrated prediction per object while suppressing all others to near-zero foreground confidence while maintaining reasonable bounding box accuracy.
    \item We introduce OCE, a metric suitable for both evaluating model calibration and identifying reliable post-processing by measuring calibration per ground-truth object, avoiding pitfalls of existing metrics.
    \item We propose an algorithm to quantify image-level reliability by measuring the confidence contrast between positive and negative predictions identified by OCE. We demonstrate the effectiveness of our approach with numerical experiments on the \texttt{COCO} and \texttt{Cityscapes} datasets.
\end{itemize}
\end{revision}

\section{Related Work} \label{sec:related}

\statement{Calibration.}
Model calibration refers to how well a model’s predicted confidence scores align with the actual likelihood of those predictions being correct. In other words, low-confidence samples should exhibit low accuracy, and high-confidence samples should demonstrate high accuracy. For example, if among all instances to which the model assigns an 80\% probability, the event occurs in about 80\% of those instances, the model is well‑calibrated. 
\revvv{Achieving good calibration is crucial because it improves trust in the model’s predictions, guides more informed decision-making, and enables better risk assessment in real-world applications~\cite{park2025know,yun2025atom,park2026tractable,park2025probabilistic}.}

To evaluate the alignment, one can measure the expected calibration error by binning predictions based on their confidence scores and computing the mean absolute error between the average confidence and the corresponding accuracy within each bin.
However, in object detection tasks, measuring accuracy is not straightforward because predictions comprise both class probabilities and bounding boxes. Additionally, due to the set-prediction nature of object detection models, it is unclear which ground-truth object corresponds to each prediction.
To address these challenges, detection expected calibration error (D-ECE) \citep{kuppers2020multivariate} defines precision as the accuracy metric and matches each prediction to a ground-truth object based on an intersection over union (IoU) threshold (commonly set at $0.5$ or $0.75$).
On the other hand, localisation-aware expected calibration error (LaECE) \citep{oksuz2023towards} defines accuracy as the product of precision and IoU, thereby accounting for localization errors as well.

Yet, when model calibration is evaluated across the entire set of predictions, which includes the secondary predictions, the resulting evaluation becomes highly biased and therefore unreliable.
To avoid this issue, previous studies have often measured D-ECE only on predictions with confidence scores exceeding $0.3$.
As detailed in later sections, applying a threshold of $0.3$ selects the well-calibrated predictions in some DETR variants; thus, using such a fixed threshold may occasionally be acceptable.
However, the optimal threshold is not guaranteed to be always $0.3$ for other models, like UP-DETR, which introduces potential risks.
In response, our paper proposes OCE, which measures the model's calibration quality alongside the employed post-processing scheme and thus can adaptively identify the reliable subset.

\statement{Uncertainty Quantification.}
Several studies focus on out-of-distribution (OOD) identification in object detection models. For example, \citet{li2022out} proposes a built-in OOD detector to isolate OOD data for human review, including those of unknown and uncertain classes (i.e., epistemic but not aleatoric uncertainty), by modeling the distribution of training data and assessing whether samples belong to any of the training class distributions. \citet{du2022vos} generates outlier data from class-conditional distribution estimations derived from in-distribution data, training the model to assign high OOD scores to this generated data and low OOD scores to the original in-distribution data. Similarly, \citet{oksuz2023towards} employs an auxiliary detection model capable of expressing its confidence. Other works, including \citet{du2022siren} and \citet{wilson2023safe}, investigate the latent representations generated by object detection models to identify the OOD nature of the input.

To the best of our knowledge, the aforementioned existing UQ techniques primarily focus on prediction-level analysis. Moreover, they predominantly address CNN-based models and explore the methodological way to better quantify the uncertainty in object detection models. In contrast, our paper emphasizes the significance of identifying a reliable subset within the entire set of predictions for uncertainty quantification, particularly in DETRs. Further, another novelty of our work lies in investigating an appropriate methodology to integrate different predictions’ confidence estimates to quantify image-level reliability.

\section{Preliminaries}

\begin{figure}
    \centering
    \includegraphics[width=\linewidth]{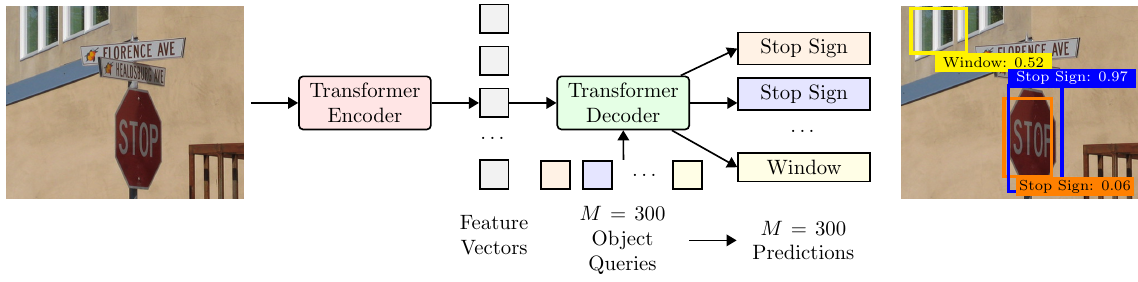}
    \caption{
    A diagram of the DETR architecture. An input image is first processed through a CNN backbone to generate a 2D feature representation. This representation is then passed to the Transformer encoder, which extracts feature vectors. These feature vectors are fed to the decoder, which receives $M$ learned object queries together. The decoder outputs $M$ prediction sets, each containing a bounding box and corresponding class probabilities.
    }
    \label{fig:detr}
\end{figure}

\subsection{Detection Transformer (DETR)} \label{sec:detr}
Consider a test image $x$ and denote the set of ground-truth objects present in the image by $\cD$.
Analogously, the set of predictions generated by DETR, parameterized by $\theta$, is denoted by $\hat{\cD}_\theta(x)$.
Each prediction $\hat{y}_\qidx \in \hat{\cD}_\theta(x)$ is characterized by a bounding box $\hat{b}_\qidx$ and an associated class probability distribution $\hat{p}_\qidx$ over $\cset \cup \{\emptyset\}$, where $\cset$ is the set of foreground classes and $\emptyset$ represents the background class.

The structure of DETR is composed of two main components: the Transformer encoder, which extracts a collection of features from the given image; and the Transformer decoder, which uses these features to make predictions.
In addition to the features extracted by the encoder, the decoder's input consists of $M$ (typically several hundred) learnable embeddings, also known as \emph{object queries}.
Each decoder layer is composed of a self-attention module among object queries and a cross-attention module between each object query and the features.
After processing the queries through several decoder layers, the model produces the $M$ final representation vectors that are converted into bounding boxes and class labels via a shared feedforward network, $f_\phi$.
Together, these predictions form the final outputs, making DETR’s predictions essentially an $M$-element set.
See Figure~\ref{fig:detr} for an illustration.

\subsection{Hungarian Loss in DETR} \label{sec:hungarian}

\rev{Let $\hat{\cD}_\theta(x) = \{ \hat{y}_1, ..., \hat{y}_M \}$, where $\hat{y}_\qidx = (\hat{p}_\qidx, \hat{b}_\qidx)$. This set is compared against $\cD = \{ y_1, ..., y_N \}$, where $N$ is the number of ground-truth objects. To evaluate the loss, a one-to-one correspondence between the two sets is first established. The ground-truth set is padded with $M-N$ background objects, and an optimal matching permutation $\optmatch$ is found by minimizing a pairwise matching cost $\mathcal{L}_{\text{match}}$ using the Hungarian algorithm~\citep{kuhn1955hungarian, carion2020end}.}

\begin{revision}
Once the optimal matching $\optmatch(\cdot)$ is determined, the actual training objective, the Hungarian Loss, is computed over these pairs. The loss for the entire set is the sum of the losses for each ground-truth object $y_\yidx = (c_\yidx, b_\yidx)$---having the target class label $c_\yidx$ and bounding box $b_\yidx$---and its matched prediction $\hat{y}_{\optmatch(\yidx)}$:
\begin{equation} \label{eq:hungarian_loss}
\mathcal{L}(\hat{y}_{\optmatch(\yidx)}, y_\yidx) = \sum_{\yidx=1}^{M} w_\yidx\left[ -\log \hat{p}_{\optmatch(\yidx)}(c_\yidx) + \mathbbm{1}_{\{c_\yidx \neq \emptyset\}} \mathcal{L}_{\text{box}}(b_\yidx, \hat{b}_{\optmatch(\yidx)}) \right] ,
\end{equation}
where $\hat{p}_\qidx(c)$ is the predicted probability for class $c$ of the $\qidx$-th prediction. $\mathcal{L}_{\text{box}}$ is a linear combination of the IoU and L1 loss between the two bounding boxes. This process naturally partitions the $M$ predictions into two groups for each image: the \emph{optimal positive} predictions, which are matched to ground-truth foreground objects, and \emph{optimal negative} predictions, which are matched to background objects. In practice, the log-probability term for negative predictions (where $c_\yidx = \emptyset$) is down-weighted (e.g., by a factor of $w_{\emptyset}=0.1$):
$$w_\yidx = \begin{cases} 
    w_{\emptyset} & \text{if } c_\yidx = \emptyset \\
    1 & \text{if } c_\yidx \neq \emptyset .
\end{cases}$$
This structure, which penalizes high-confidence foreground predictions from negative predictions, forms the basis of our subsequent analysis.
\end{revision}

\begin{figure*}[t]
    \centering
    \subfloat[Accurate boxes. Calibrated.]{
        \includegraphics[width=0.3\linewidth]{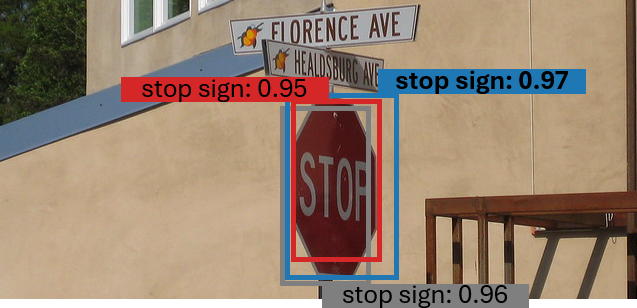}
        \label{fig:hypothesis_a}
    }
    \subfloat[Random boxes. Low Confidence.]{
        \includegraphics[width=0.3\linewidth]{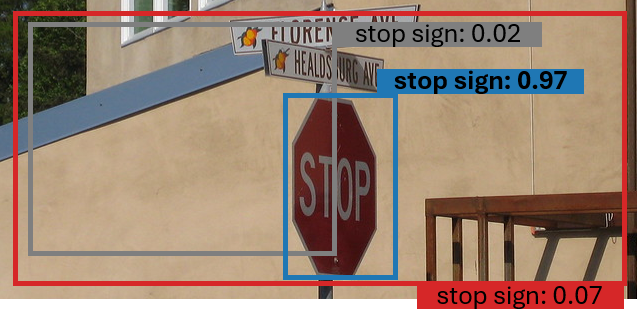}
        \label{fig:hypothesis_b}
    }
    \subfloat[Accurate boxes. Uncalibrated.]{
        \includegraphics[width=0.3\linewidth]{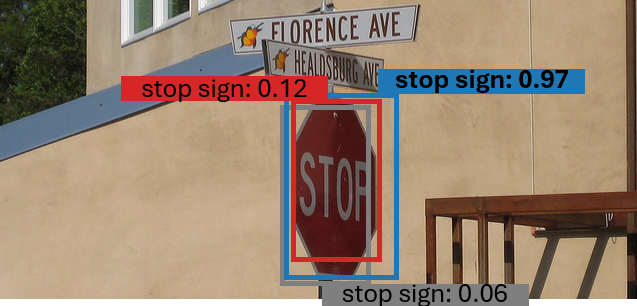}
        \label{fig:hypothesis_c}
    }
    \label{fig:hypotheses}
    \caption{
    DETR generates hundreds of predictions for each image, resulting in multiple predictions per object, with at least one (i.e., \textcolor{Blue}{\textbf{blue boxes}}) being well-calibrated.
    This figure illustrates how DETR handles the remaining predictions (i.e., \textcolor{Red}{\textbf{red}} and \textcolor{Gray}{\textbf{gray}} boxes). 
    DETR could (a) assign equally high confidence scores with accurate bounding boxes to all predictions; \rev{however, this strategy fails to optimize the Hungarian loss, as high-confidence unmatched predictions incur substantial penalties. Instead, DETR could assign low confidence to unmatched predictions, either with (b) random bounding boxes or (c) accurate bounding boxes.} Our analysis reveals that DETR follows the third strategy, which explains the varying levels of reliability observed across predictions.}
\end{figure*}

\section{\rev{Optimal Prediction Strategies in DETR}}
\label{sec:impact_reliability}

\begin{revision}
In this work, we evaluate a model's reliability using the concept of calibration. 
We define a prediction as perfectly calibrated if its predicted probability distribution matches the true posterior probability distribution of the object it is matched to.
We formalize this by considering an object whose true, and potentially ambiguous, class identity is represented by a ground-truth probability distribution $p_{\text{gt}}$. 
A prediction is \emph{perfectly calibrated} with respect to this object if its predicted distribution $\hat{p}$ exactly matches the true distribution $p_{\text{gt}}$, i.e.,
\begin{equation*}
    \hat{p}(c) = \revvv{p_{\text{gt}}}(c) \quad \forall c \in \cset \cup \{\emptyset\}.
\end{equation*}
\end{revision}

\subsection{Motivation and Scope}
During the inference stage at test-time, ground-truth annotations are unavailable, meaning the optimal positive predictions remain unknown.
This raises the question of \textbf{Which of the predictions can be trusted and used?}
If all of the model's predictions were generated independently and were well-calibrated (e.g., Figure~\ref{fig:hypothesis_a}), the large number of predictions would not be a concern. We could simply apply well-known algorithms like NMS to remove duplicates and resolve redundancy.

\begin{revision}
However, empirical evidence suggests that such independence is unlikely to happen, as it appears to conflict with the minimization of the Hungarian loss in Equation~\eqref{eq:hungarian_loss}. In practice, we observe that DETR tends to assign a well-calibrated confidence score to only a single prediction per object, while suppressing remaining predictions with low foreground confidence. Intriguingly, although the Hungarian loss imposes no penalty on the localization of these background-matched optimal negative predictions—implying they could potentially output random bounding boxes (Figure~\ref{fig:hypothesis_b})—we find that DETR exhibits a strong preference for accurate localization (Figure~\ref{fig:hypothesis_c}), consistently producing precise boxes even for optimal negative predictions.
In the following sections, we investigate this behavior in depth.
\end{revision}

\begin{figure*}[t]
    \centering
    
    \subfloat[]{
        \includegraphics[width=1.0\textwidth]{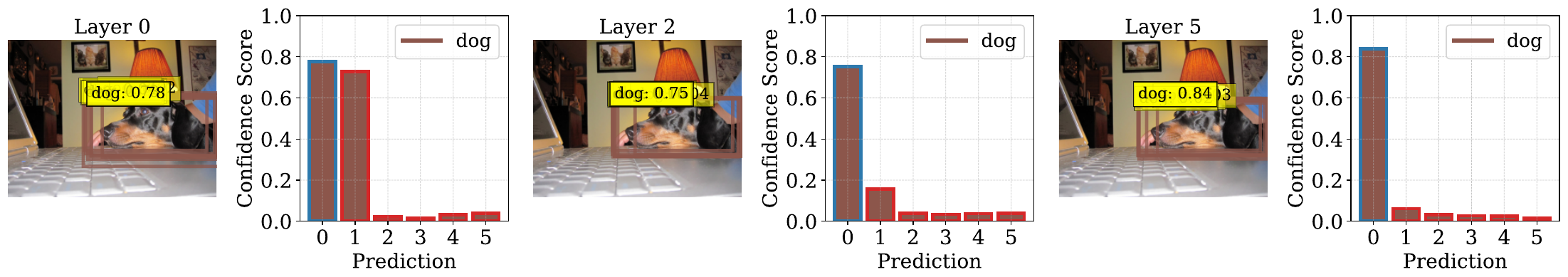}
        \label{fig:insight-reli-dog}
    }
    
    \subfloat[]{
        \includegraphics[width=1.0\textwidth]{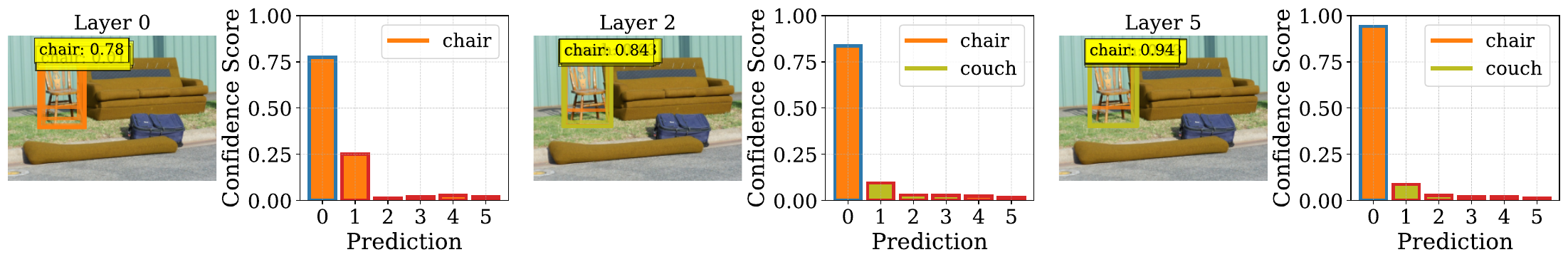}
        \label{fig:insight-reli-chair}
    }
    
    \caption{Visualizations of the predictions generated by Cal-DETR. The optimal positive prediction (indexed by 0 and \textcolor{Blue}{bordered in blue}) and the five optimal negative predictions (indexed by 1-5 and \textcolor{BrickRed}{bordered in red}) with the largest IoU are presented.
    For each prediction and layer, the maximum confidence score and its corresponding label are visualized.
    \rev{The model's attention mechanism enables it to learn a ``specialist strategy.'' When an object is confidently identified, a single primary query (the intended optimal positive prediction) is assigned a (high) calibrated confidence score. In parallel, all other remaining queries (the intended optimal negative predictions) have their confidence scores suppressed to near-zero, even while they maintain accurate bounding box predictions. This specialist strategy---where the model designates one query as the sole reliable predictor and the rest as background specialists---is demonstrably favored by the loss function over having multiple competing, moderately-confident predictions for the same object.}
    }
    \label{fig:insight}
\end{figure*}

\subsection{Exploring the Anatomy of DETR's Predictions} \label{sec:insight}
\begin{figure*}[t]
    \centering
    \subfloat[High-reliability Image]{
        \includegraphics[width=1.0\textwidth]{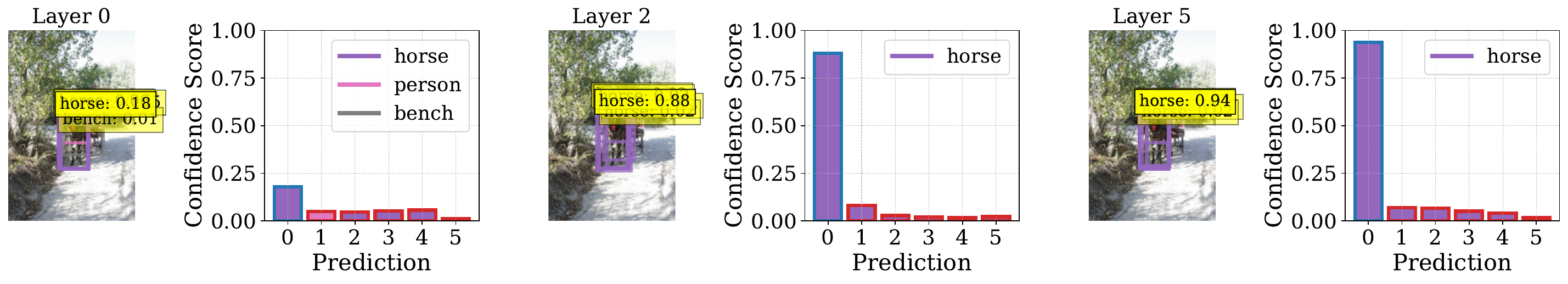}
        \label{fig:insight-reli-horse}
    }
    
    \subfloat[Low-reliability Image]{
        \includegraphics[width=1.0\textwidth]{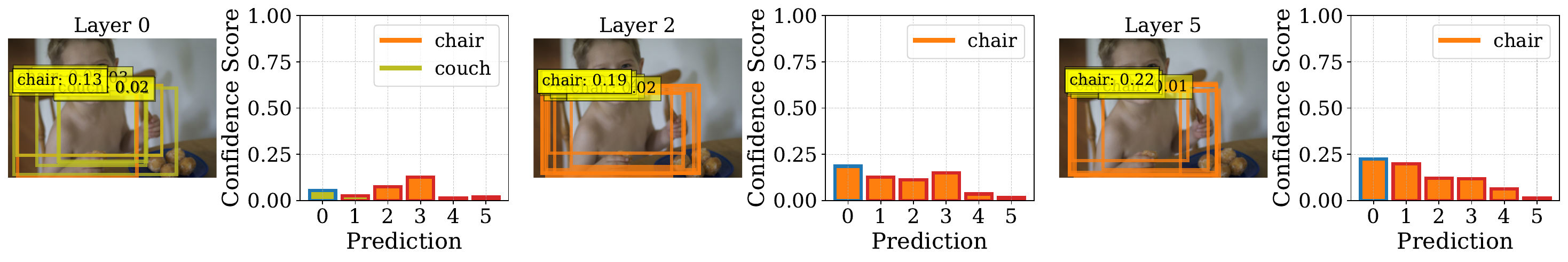}
        \label{fig:insight-unreli-chair}
    }
    \caption{When the model is confident, it assigns a high confidence score to the optimal positive prediction by either increasing or remaining high across the decoder layers via cross-attention mechanism---thus be calibrated. While those of remaining optimal negative queries' confidence scores are pushed to be near-zero while having accurate bounding box predictions.
    On the other hand, when the model is uncertain, DETR assigns a lower confidence score to the optimal positive prediction, thereby maintaining good calibration.
    However, \textbf{conversely}, it slightly increases or maintains the confidence scores for remaining predictions.
    }
    \label{fig:insight_contrast}
\end{figure*}

We begin our qualitative analysis by visualizing and examining the outputs generated from the DETR decoders.
Since the Transformer decoder outputs only representation vectors, investigating their evolution across layers is not straightforward.
We address this by reapplying the final feedforward network that operates on the last layer, $f_\phi$, to the intermediate layers.
This allows us to transform each representation vector into its associated bounding box and class label.
This is feasible due to the alignment of intermediate representations, facilitated by residual connections between decoder layers~\citep{chuang2023dola}.
Sample visualizations are in Figures~\ref{fig:insight} and \ref{fig:insight_contrast}. %

In the first decoder layer, the model appears to explore the encoded image features, producing varied queries that result in various plausible predictions.
In this early stage, the distinction between optimal positive and negative queries can be ambiguous (e.g., Figure~\ref{fig:insight-reli-dog} and Figure~\ref{fig:insight-reli-horse}).
However, the attentions through the subsequent decoder layers progressively refine these predictions.
By the final layer, the model selects a single prediction and assigns a confidence score based on its understanding of the image and the object.
In contrast, the confidence scores for remaining queries do not increase to the same extent as the positives and even decrease. %
On the other hand, in images of low reliability (where the model is uncertain), the confidence score of the optimal positive prediction does not significantly increase, while the scores of the optimal negative ones are either slightly raised or unchanged. 
Based on this observation, we present our main claim:

\rev{
\emph{DETR learns to assign multiple predictions for each object within a given image and to designate a single prediction that is calibrated, while the remaining predictions are suppressed to have low foreground confidence, even when maintaining accurate localization.
When the model's uncertainty about its prediction (and thus about the matching process) is high, the confidence separation between those predictions diminishes.}
}

\begin{revision}
\subsection{Theoretical Analysis of the Optimal Prediction Strategy} \label{sec:specialist}

To provide theoretical support for the claim, we analyze the optimal prediction strategy in a stochastic setting. We define a random variable $y_{\text{gt}} = (c_{\text{gt}}, b_{\text{gt}})$ representing the ground-truth object with distribution $Y$, where $c_{\text{gt}} \in \cset$ represents the class that is sampled from the class distribution $p_{\text{gt}}$, and $b_{\text{gt}}$ represents the bounding box sampled from some distribution $\mathcal{B}$.

The padded set of $M$ targets is denoted by $\{y_\yidx\}_{\yidx=1}^{M}$, where $y_1 = y_{\text{gt}}$ (the sampled object) and $y_\yidx = \emptyset$ for $\yidx \neq 1$. The model provides a set of predictions $\{\hat{y}_\qidx\}_{\qidx=1}^M$. The objective is to find predictions that minimize the expected total loss over the distribution $Y$:
\begin{equation*}
    \mathbb{E}\left[ \Ltotal \right] = \mathbb{E}_{y_{\text{gt}} \sim Y} \left[ \sum_{\yidx=1}^M \mathcal{L}(\hat{y}_{\optmatch(\yidx)}, y_\yidx) \right].
\end{equation*}
To formalize the roles of the predictions, we define the index of the optimal positive query as $p = \optmatch(1)$, while the remaining $M-1$ indices, $\{n\}_{n \ne p}$, are the optimal negative queries.
Given this definition, we can decompose the total loss into terms for the optimal positive and optimal negative indices:
\begin{equation*}
    \mathbb{E}\left[ \Ltotal \right] = \mathbb{E}_{y_{\text{gt}} \sim Y} \left[ \mathcal{L}(\hat{y}_p, y_{\text{gt}}) \right] + \sum_{n \ne p} \mathcal{L}(\hat{y}_n, \emptyset).
\end{equation*}
The loss for the optimal positive prediction $\hat{y}_p = (\hat{p}_p, \hat{b}_p)$ matched to the $y_{\text{gt}}$ is:
\begin{align} \label{eq:rev_pos_loss}
\mathbb{E}\left[ \mathcal{L}(\hat{y}_p, y_{\text{gt}}) \right] &= \mathbb{E}_{y_{\text{gt}} \sim Y}  \left[ \Lbox(\hat{b}_p, b_{\text{gt}}) - \log(\hat{p}_{p}(c_{\text{gt}})) \right] \notag \\
&= \mathbb{E}_{b_{\text{gt}} \sim \mathcal{B}}  \left[ \Lbox(\hat{b}_p, b_{\text{gt}}) \right] - \sum_{c \in \cset} p_{\text{gt}}(c) \log(\hat{p}_{p}(c)),
\end{align}
where $\hat{p}_p(c)$ is the predicted probability for class $c$, as defined before.
The loss for optimal negative prediction $\hat{y}_n$ matched to $\emptyset$ (one of the optimal negatives) is the down-weighted negative log-probability of the background class:
\begin{equation} \label{eq:rev_neg_loss}
\mathcal{L}(\hat{y}_n, \emptyset) = -w_{\emptyset} \log(\hat{p}_{n}(\emptyset))
.\end{equation}
Here, $w_{\emptyset}$ is the down-weighting factor (e.g., 0.1) for the background class.
The following result establishes a lower bound on the expected total loss.

\begin{prop} \label{prop:bayes}
The minimum achievable expected total loss is lower bounded by the entropy of the ground-truth class distribution, i.e., $$\mathbb{E}\left[ \Ltotal \right] \ge \entropy(p_{\text{gt}}) + \min_{\hat{b}} \mathbb{E}[\Lbox(\hat{b}, b_{\text{gt}})],$$ where $\entropy(p_{\text{gt}}) = -\sum_{c \in \cset} p_{\text{gt}}(c) \log(p_{\text{gt}}(c))$.
\end{prop}
\begin{proof}
See Section~\ref{proof:bayes}.
\end{proof}

Next, we characterize the optimal prediction set that uniquely achieves this bound.

\begin{prop}[Optimal Prediction Set] \label{prop:specialist}
\revv{Let $\{(\hat{p}^*_\qidx, \hat{b}^*_\qidx)\}_{\qidx=1}^{M}$ denote a globally optimal prediction set, i.e., that minimizes $\mathbb{E}\left[ \Ltotal \right]$. Then we have}
\begin{enumerate}
    \item for the optimal positive prediction ($\qidx=p$): $\hat{p}^*_\qidx = p_{\text{gt}}$ (i.e., $\hat{p}^*_{\qidx}(c) = p_{\text{gt}}(c)$ $\forall c \in \cset$ and $\hat{p}^*_{\qidx}(\emptyset)=0$) and the box $\hat{b}^*_\qidx$ minimizes $\mathbb{E}[\Lbox(\hat{b}_\qidx, b_{\text{gt}})]$, and
    \item for any optimal negative prediction ($\qidx \ne p$): $\hat{p}^*_{\qidx}(c)=0$ $\forall c \in \cset$ and $\hat{p}^*_{\qidx}(\emptyset)=1$.
\end{enumerate}
\end{prop}
\begin{proof}
See Section~\ref{proof:specialist}.
\end{proof}

\statement{\rev{From Optimal Set to Strategy under Ambiguity.}}
The previous propositions formally show that the Hungarian loss incentivizes a deterministic specialization. 
The optimal state is not a symmetric hedge where multiple queries compete, but for one query to emerge as a specialist that becomes a perfectly calibrated predictor of the object's true class distribution ($p_{\text{gt}}$).
All other queries are simultaneously optimized to predict the background class, minimizing their own loss to zero.

This finding implies that the model is incentivized to develop an internal strategy to achieve this state. 
Ideally, the model would learn to designate a specific object query (which we conceptually refer to as the \emph{primary} query) to become the optimal positive prediction $p$, and the remaining ones (\emph{secondary queries}) to become the optimal negatives.
As we discussed in Section~\ref{sec:insight}, this is architecturally feasible, as DETR can leverage its attention mechanism to orchestrate predictions: Throughout the decoder layers, the model can identify the most promising object query and progressively specialize it, while simultaneously suppressing the others.

However, in practice, a critical ambiguity arises. While the Hungarian matching algorithm $\optmatch$ is deterministic for a deterministic annotation, the ground truth itself is often ambiguous.
Consequently, the assignment of queries to targets becomes a stochastic process.
We formalize this by defining $P_\qidx$ as the probability that the $\qidx$-th query is matched to the ground-truth object:
\begin{equation*}
P_\qidx \defined \mathbb{P}_{y_\text{gt} \sim Y}(\optmatch(1) = \qidx).
\end{equation*}

In this setting, the extreme specialization is now suboptimal because it incurs an infinite loss if the primary prediction is matched to the background. %
Therefore, to minimize its expected loss under this uncertainty, each query must find an optimal prediction $\hat{p}_\qidx^*$ that ``hedges'' against this matching uncertainty.
The expected loss $J_\qidx$ for query $\qidx$ is decomposed as:
\begin{align*}
J_\qidx &= \underbrace{P_\qidx \cdot \mathbb{E}[\Lbox(\hat{b}_\qidx, b_{\text{gt}}) \mid \optmatch(1)=\qidx]}_{\text{Expected Box Loss Term}} \notag \\
&\quad \underbrace{- P_\qidx \sum_{c \in \cset} p_{\text{gt}}(c) \log(\hat{p}_\qidx(c)) - (1-P_\qidx) w_{\emptyset} \log(\hat{p}_\qidx(\emptyset))}_{\text{Expected Classification Objective } J(\hat{p}_\qidx)}.
\end{align*}
Minimizing $J_\qidx$ with respect to $\hat{p}_\qidx$ is equivalent to minimizing the expected classification objective $J(\hat{p}_\qidx)$.
Proposition \ref{prop:undercal} characterizes the optimal strategy to minimize the objective.

\begin{prop}[Optimal Strategy under Uncertainty] \label{prop:undercal}
For a query with a matching probability $P_\qidx$ to the ground-truth object whose class distribution is $p_{\text{gt}}$, the class prediction that minimizes the expected loss is $\hat{p}_\qidx^*$, where
\begin{enumerate}
    \item the prediction for any foreground class $c \in \cset$ is:
    \begin{equation*} %
    \hat{p}_\qidx^*(c) = p_{\text{fg}}^*(P_\qidx) p_{\text{gt}}(c)
    .\end{equation*}
    \item the prediction for the background class is $$\hat{p}_\qidx^*(\emptyset) = 1 - p_{\text{fg}}^*(P_\qidx),$$
\end{enumerate}
where $p_{\text{fg}}^*(P_\qidx)$ is the optimal foreground probability, defined as
\begin{equation*} %
p_{\text{fg}}^*(P_\qidx) = \frac{P_\qidx}{P_\qidx + w_{\emptyset}(1-P_\qidx)}.
\end{equation*}
\end{prop}
\begin{proof}
See Section~\ref{proof:undercal}.
\end{proof}

Based on the result from Proposition~\ref{prop:undercal}, we analyze two characteristics in their prediction strategy:

\begin{itemize}[leftmargin=10pt]
    \setlength\itemsep{0.25em}
    \item \textbf{Perfect Specialization:} 
    Consider the scenario where the model is highly certain about the image context and thus about which query will be matched. In this scenario, the matching probability for the primary---which will eventually become the optimal positive---query index is perceived as 1 (i.e., $P_{p} \to 1$), while it vanishes for all others (i.e., $P_{\qidx} \to 0$ for $\qidx \ne p$).
    In this scenario, the strategy recovers the optimal prediction set (Proposition~\ref{prop:specialist}), where solely the $p$-th query becomes a perfectly calibrated predictor of the object ($p_{\text{gt}}$).
    
    \item \textbf{Hedging:}
    When the model is uncertain about the matching query (i.e., $P_\qidx$ is dispersed), it optimally ``hedges.'' The foreground probability for the primary query is suppressed ($p_{\text{fg}}^*(P_p) < 1$), balancing the risk between foreground and background assignments.
\end{itemize}

In practice, matching probabilities are skewed: $P_p$ is close to 1 for the primary query, and $P_\qidx$ is close to 0 for secondary queries: $\qidx \neq p$. Thus, DETRs prioritize specialization while retaining a hedging mechanism proportional to their uncertainty. We term this the \emph{Specialist Strategy}. This \revvv{jibes} with the empirical trends observed in Figures~\ref{fig:insight} and \ref{fig:insight_contrast}.
\end{revision}

\begin{revision2}
\statement{Extension to Multi-Object Scenarios.}
Our theoretical analysis focuses on the single-object case and therefore does not fully characterize general multi-object scenes. Nevertheless, the result suggests a natural extension: when objects can be treated as independent in the matching process, the Hungarian objective still encourages one query to specialize to each foreground object, while the remaining unmatched queries are driven toward background predictions. In this sense, the specialist strategy is expected to apply on a per-object basis.
We extend Propositions~\ref{prop:bayes} and \ref{prop:specialist} into the multiple object case and present them in Appendix~\ref{app:multi}.

The situation becomes more complex when multiple objects compete strongly during matching---for example, due to spatial overlap with similar appearance---and encourages high uncertainty in matching. In other words, the matching uncertainty for each query increases, and the optimal behavior is expected to shift from specialization toward hedging. We provide qualitative evidence for this conjecture in  Appendix~\ref{app:multi}, through multi-object inference examples. %
\end{revision2}

\subsection{Numerical Analysis}

\statement{Setup.}
To provide quantitative support for our claim, we conducted experiments using four DETR variants: UP-DETR \citep{dai2021up}, Deformable-DETR (D-DETR) \citep{zhu2020deformable}, Cal-DETR \citep{munir2024cal}, and DINO \citep{zhang2022dino}.
\rev{We also investigate with Faster R-CNN \citep{ren2017faster} and YOLOv3 \citep{redmon2018yolov3, redmon2016you}.}
\revv{Each model is trained on the \texttt{COCO} training dataset (train2017), and, for validation, we use $1,000$ held-out images (i.e., $20\%$) of the \texttt{COCO} validation set (val2017).
For testing, we use the remaining $4,000$ images (i.e., the remaining $80\%$) of the \texttt{COCO} validation set (val2017). For out-of-distribution tests, we use the test datasets of \texttt{Cityscapes} \citep{cordts2016cityscapes} and \texttt{Foggy Cityscapes} \citep{sakaridis2018semantic}.}

\begin{figure*}
    \centering
    \subfloat[\label{fig:ecdf_cal_opt_}]{\includegraphics[width=0.23\linewidth]{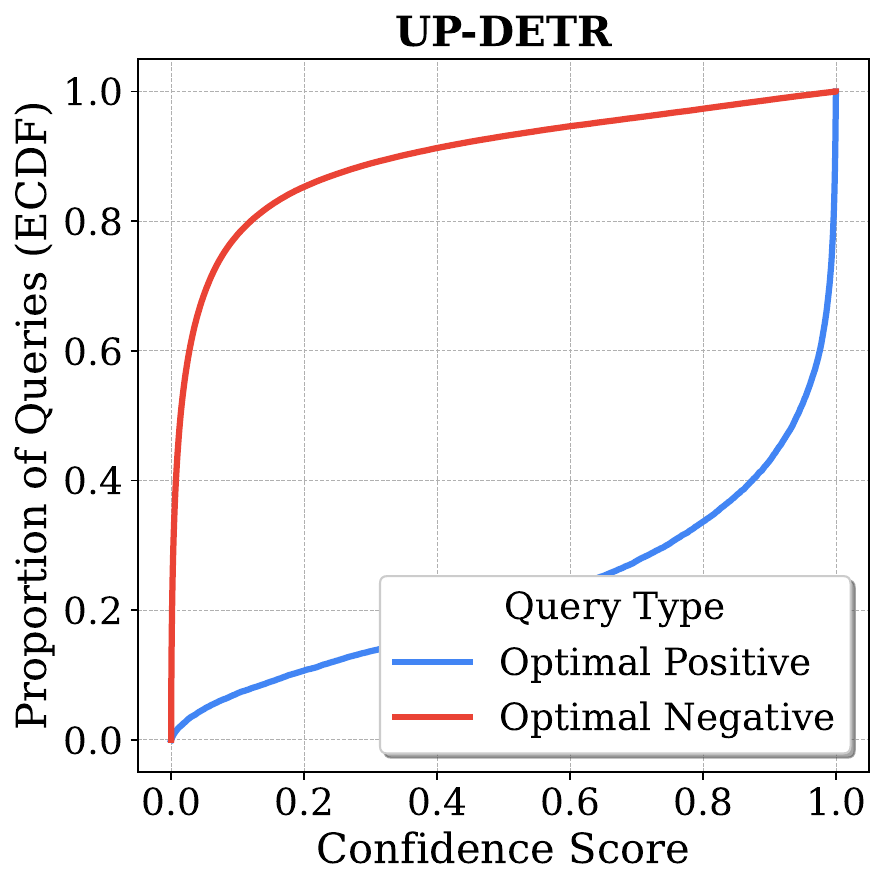}}
    \hspace{1em}
    \subfloat[\label{fig:ecdf_cal_opt}]{\includegraphics[width=0.23\linewidth]{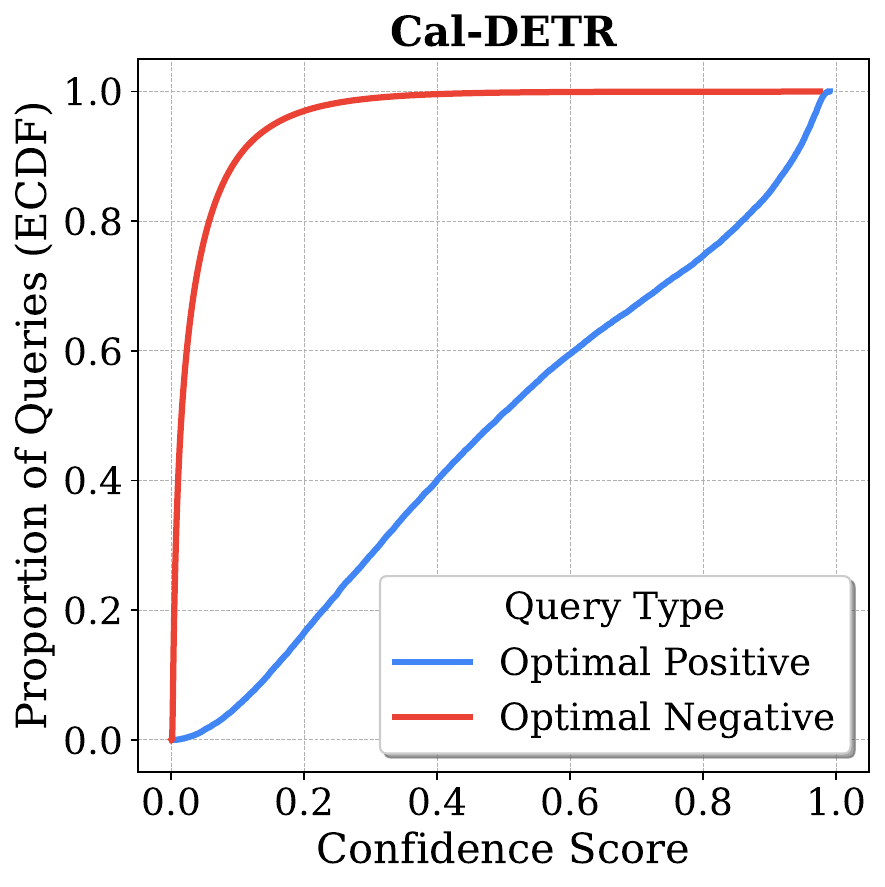}}
    \hspace{1em}
    \subfloat[\label{fig:ecdf_bbox_cal_opt_}]{\includegraphics[width=0.23\linewidth]{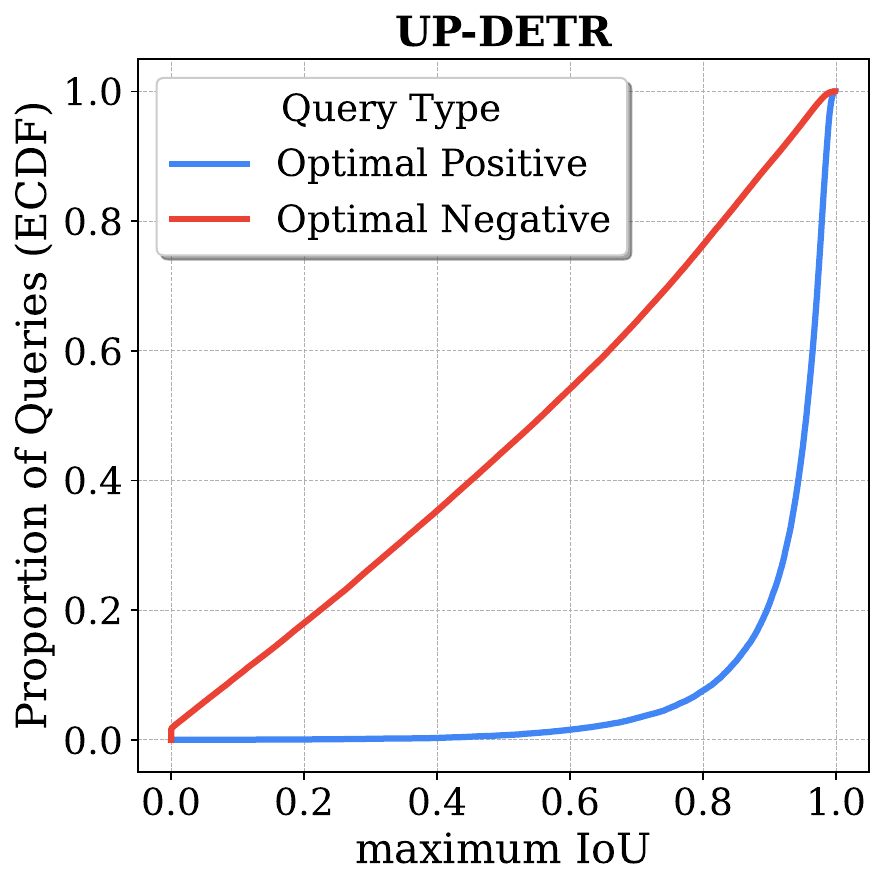}}
    \hspace{1em}
    \subfloat[\label{fig:ecdf_bbox_cal_opt}]{\includegraphics[width=0.23\linewidth]{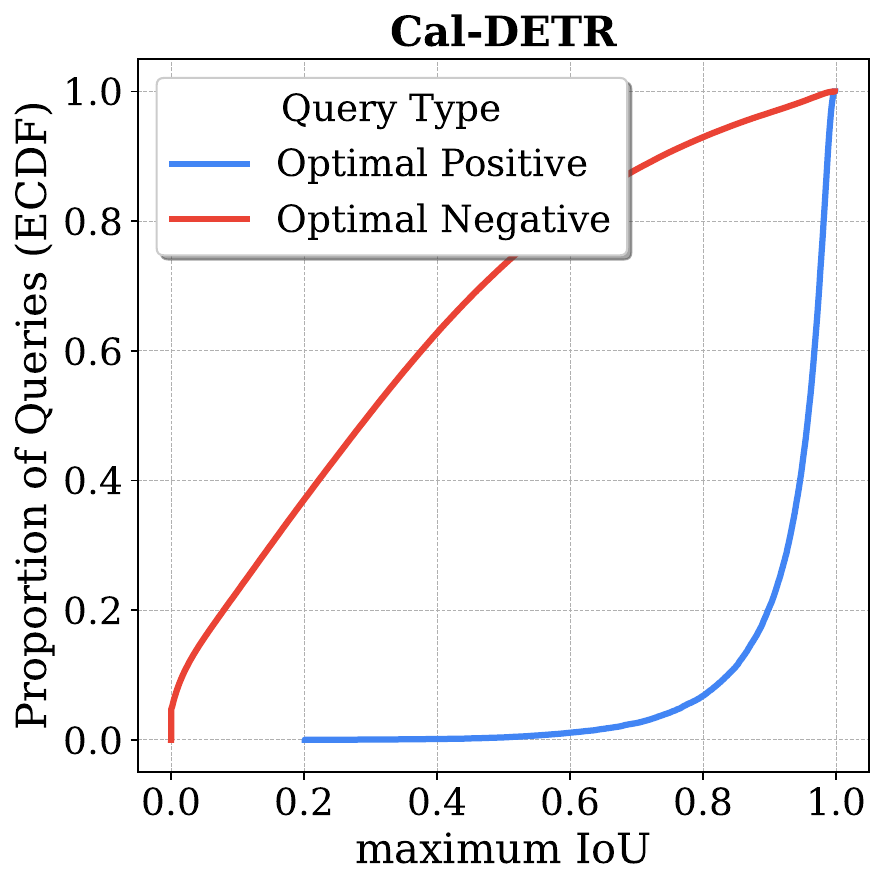}}
    
    \caption{
    \rev{Empirical cumulative distribution function (ECDF) plots for maximum confidence score over foreground classes and maximum IoU over ground truth objects, comparing optimal positive predictions (blue) and optimal negative predictions (red), in UP-DETR and Cal-DETR on the \texttt{COCO} dataset. (a, b) Confidence: The optimal negative prediction confidence is overwhelmingly clustered at zero. Optimal positive predictions span the full confidence range, acting as the system's calibrated predictors. (c, d) Localization: Optimal positive predictions are highly accurate (high IoU), which is the necessary condition for the model to designate a primary query. Optimal negative predictions maintain decent localization, which is not a failure but the optimal risk-averse strategy.}
    } \label{fig:ecdf}
\end{figure*}

\begin{revision}

\statement{\rev{Empirical Analysis of Prediction-Level Distributions.}}
We now empirically analyze the distributions of both confidence and localization accuracy to validate our theoretical findings.
Ideally, this analysis would compare the primary prediction (the query intended to be the optimal positive) against secondary predictions.
However, since the model's internal designation is latent and not directly observable, we utilize the outcomes of the Hungarian matching as effective proxies.
Specifically, we analyze optimal positive predictions (those matched to the ground truth) as the observable realization of primary prediction, and optimal negative predictions (those matched to the background) as the proxy for secondary predictions.

First, we investigate the confidence scores. We plot the empirical cumulative distribution function (ECDF) of the maximum foreground probability ($\max_{c\neq\emptyset}\hat{p}_\qidx({c})$) in Figures~\ref{fig:ecdf_cal_opt_} and \ref{fig:ecdf_cal_opt}.
This indicates that the optimal negative predictions' confidence scores are heavily concentrated in the near-zero regime, effectively being suppressed, in contrast to the widely dispersed confidence distribution observed for optimal positive predictions.

Next, we perform a parallel analysis on localization accuracy in Figures~\ref{fig:ecdf_bbox_cal_opt_} and \ref{fig:ecdf_bbox_cal_opt}. 
More specifically, for each prediction, we measure the maximum IoU with respect to the ground truth objects in the image.
The results for optimal positive predictions confirm that the vast majority exhibit high accuracy. 
On the other hand, the analysis of optimal negative predictions reveals a critical distinction: despite their suppressed confidence, they frequently maintain substantial localization accuracy.
It reflects a risk-averse strategy where the model maintains precise backup bounding boxes to mitigate the high penalty of potential matching errors, ensuring training stability.
As shown in Figure~\ref{fig:ecdf_appendix}, these observed trends are consistent across various DETR variants and other detectors.

We remark that this is not a model failure, but rather a risk-averse strategy induced by the loss landscape. 
In a stochastic setting where $y_{\text{gt}}$ varies, there is always a non-zero probability that secondary queries get matched to the ground-truth object.
If the secondary query predicts a random box, its matching to the foreground object elevates the expected box loss.
Thus, to minimize the expected total loss, the model is incentivized to set secondary queries to predict a decent bounding box rather than a random box.

\end{revision}

\begin{revision}

\begin{figure*}[t]
    \centering
    \includegraphics[width=0.95\textwidth]{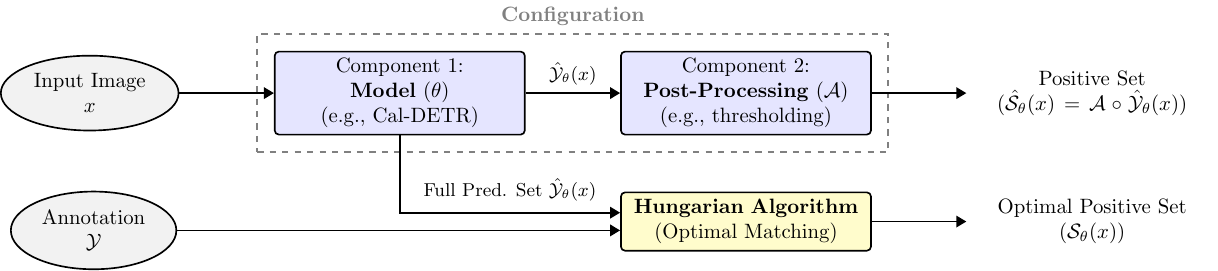}
    \caption{
     \rev{Flowchart of our evaluation framework. Two components vary across configurations:
     (1) the model ($\theta$), which generates a full set of predictions ($\hat{\cD}_\theta(x)$), and
     (2) the post-processing algorithm ($\mathcal{A}$), which selects a positive set ($\hat{\mathcal{S}}_\theta(x)$) from these predictions. In addition, we can obtain the optimal positive sets ($\mathcal{S}_\theta(x)$) by passing the full prediction set ($\hat{\cD}_\theta(x)$) and ground-truth object annotations ($\cD$) to the Hungarian Algorithm.
     }
    }
    \label{fig:joint_evaluation_framework}
\end{figure*}

\begin{figure}[t]
    \centering
    \includegraphics[width=0.95\columnwidth]{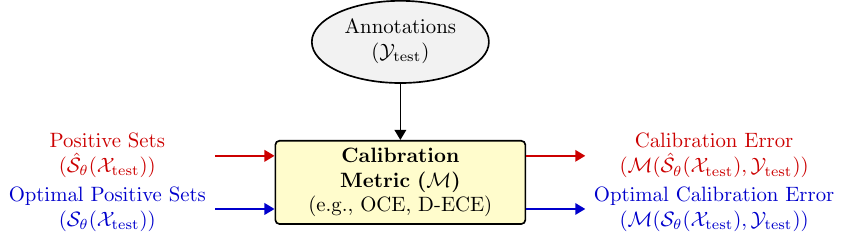}
    \caption{
    \rev{Flowchart of our method for evaluating calibration error. A calibration metric ($\mathcal{M}$) compares either given positive sets ($\hat{\mathcal{S}}_\theta(\mathcal{X}_{\text{test}})$) or optimal positive sets (${\mathcal{S}}_\theta(\mathcal{X}_{\text{test}})$) with ground-truth object annotations ($\cD_{\text{test}}$) to get calibration error ($\mathcal{M}(\hat{\mathcal{S}}_\theta(\mathcal{X}_{\text{test}}), \cD_{\text{test}})$) or optimal calibration error ($\mathcal{M}({\mathcal{S}}_\theta(\mathcal{X}_{\text{test}}), \cD_{\text{test}})$), respectively.
    }
    }
    \label{fig:calibration_metric}
\end{figure}

\section{Quantifying Calibration in DETR} \label{sec:uq_obj}

\subsection{Implications of the Specialist Strategy for Evaluation}
Our analysis reveals that DETR's training objective incentivizes a specialist strategy. 
This strategy assigns two fundamentally different roles to the predictions:
\begin{enumerate}
    \item A small subset of primary predictions (intended to be matched with foreground objects) is trained to be \emph{calibrated predictors}, matching the true class distribution $p_{\text{gt}}$.
    \item The vast majority of secondary predictions are trained to be \emph{background classifiers}, intentionally suppressing their foreground confidence to near-zero.
\end{enumerate}
This specialist strategy introduces a fundamental ambiguity in downstream decision-making. 
When a prediction exhibits low confidence, it becomes indistinguishable: is it a positive prediction correctly expressing high uncertainty about an ambiguous object, or is it a secondary prediction whose confidence has been artificially suppressed? This distinction is paramount, particularly in safety-critical domains.
\begin{tcolorbox}[
    colback=white,
    colframe=blue!70!black!80,
    boxrule=1pt,
    arc=0mm,
    title=\textbf{Safety-Critical Implication}
]
\rev{For an autonomous vehicle, a low-confidence ``person'' prediction could be a truly ambiguous object (like a person depicted on a billboard or on a vehicle), which the model is correctly uncertain about.
However, it could also be a definite pedestrian identified by a secondary prediction, whose confidence was artificially suppressed to near-zero as part of the model's behavior, resulting from optimizing the Hungarian loss (as discussed in \rev{Section~\ref{sec:impact_reliability}}).
In the latter case, relying on this suppressed score could lead the system to catastrophically misjudge the risk posed by a real person.}
\end{tcolorbox}

Ideally, at test time, we aim to filter out the secondary predictions and retain only the primary ones.
However, identifying the primary predictions is non-trivial because the model's internal designation is latent.
Consequently, practitioners rely on a post-processing algorithm (e.g., confidence thresholding) to select a subset of predictions serving as an estimate of the primary set.

While one might consider evaluating calibration directly on the optimal positive predictions, identifying this set requires ground-truth labels, making it infeasible at test time.
Evaluating against this oracle set is useful for analysis, but it does not perfectly reflect the actual deployment conditions, where the system must rely solely on a post-processed subset.

These issues create a critical challenge for practical deployment: we must evaluate calibration quality using the same post-processing algorithm that will be applied at test time.
Therefore, \textbf{practitioners should assess the entire (model + post-processing) pipeline jointly to identify configurations that produce reliable positive sets}, as illustrated in Figures~\ref{fig:joint_evaluation_framework} and \ref{fig:calibration_metric}, rather than evaluating the model in isolation.
To address this, we propose a new calibration metric that takes as input any positive set produced by a post-processing algorithm, enabling such joint evaluation.

\end{revision}

\subsection{Proposed Metric: Object-Level Calibration Error} \label{sec:oce}

\statement{Notation.} For each image $x_\imgidx$ having $N_\imgidx$ objects, we consider a set of ground-truth object annotations $\cD_\imgidx = \{  y_{\imgidx, \yidx} = (c_{\imgidx, \yidx}, b_{\imgidx, \yidx}) \}_{\yidx=1}^{N_\imgidx}$, and a set of DETR predictions: $\hat{\cD}_\theta(x_\imgidx) = \{ \hat{y}_{\imgidx, \qidx} = (\hat{p}_{\imgidx, \qidx}, \hat{b}_{\imgidx, \qidx}) \}_{\qidx=1}^M$ where $M$ represents the number of object queries.
Let $\cD_{\text{test}} = \big(\cD_{\imgidx}\big)_{\imgidx=1}^{N_{\text{test}}}$ denote the set of all ground-truth objects across all $N_{\text{test}}$ test images $\mathcal{X}_{\text{test}} = \big(x_\imgidx\big)_{\imgidx=1}^{N_{\text{test}}}$.

\begin{defn} \label{defn::cal}
Consider a subset of predictions, $\hat{\mathcal{S}}_\theta(x_\imgidx) \subseteq \hat{\cD}_\theta(x_\imgidx)$, that is generated by post-processing algorithm $\mathcal{A}$ from the full prediction set: $\hat{\mathcal{S}}_\theta = \mathcal{A} \circ \hat{\cD}_\theta$. We define an object-level calibration error (OCE) as the average Brier score per object:
\begin{gather}
    \mathrm{OCE}_{\tau} \big( \hat{\mathcal{S}}_\theta, \cD_{\text{test}} \big) = \frac{1}{|\cD_{\text{test}}|} \sum_{y_{\imgidx, \yidx} \in \cD_{\text{test}}} \mathrm{Brier}_{\tau} \big( \hat{\mathcal{S}}_\theta(x_\imgidx),  y_{\imgidx, \yidx} \big), \nonumber \\ %
    \mathrm{Brier}_{\tau} \big( \hat{\mathcal{S}}_\theta(x_\imgidx),  y_{\imgidx, \yidx} \big) = \sum_{c=1}^C \big( \mathbbm{1}(c=c_{\imgidx, \yidx}) - \bar{p}^{(\tau)}_{\imgidx, \yidx}(c) \big)^2,  \nonumber \\
    \bar{p}^{(\tau)}_{\imgidx, \yidx}(c) = \frac{1}{|\cQ^{(\tau)}_{\imgidx, \yidx}|} \sum_{q \in \cQ^{(\tau)}_{\imgidx, \yidx}} \hat{p}_{\imgidx, \qidx}(c), \label{eq:oce_agg}
\end{gather}
where $\cQ^{(\tau)}_{\imgidx, \yidx}$ is a set of prediction indices that has bounding box overlap with the ground-truth object $y_{\imgidx, \yidx}$:
\begin{equation}
    \cQ^{(\tau)}_{\imgidx, \yidx} \defined \{ \qidx \mid \text{IoU} (b_{\imgidx, \yidx}, \hat{b}_{\imgidx, \qidx}) \ge \tau \}. \nonumber
\end{equation}
When $|\cQ^{(\tau)}_{\imgidx, \yidx}| = 0$, we consider the predicted probability to be zero, thus the corresponding Brier score is estimated as $1.0$.
Following \citep{lin2014microsoft, kuzucu2024calibration}, we use IoU thresholds of $\tau = 0.5, 0.75$ and report the average score as $\mathrm{OCE}(\cdot) = \mathrm{OCE}_{0.5}(\cdot) + \mathrm{OCE}_{0.75}(\cdot)$. \revv{The ablation study on the choice of aggregation methods and IoU thresholds can be found in Section~\ref{sec:oce_abl}.}
\end{defn}

\rev{The introduced object-level calibration error (OCE) has two desirable characteristics derived from its object-centric design.
Unlike prediction-level metrics (e.g., D-ECE) that average errors over the selected predictions, OCE averages errors over the ground-truth objects.
Consequently, the metric effectively penalizes two critical failure modes in recovering the primary predictions.
(1) Retaining secondary predictions: OCE penalizes subsets containing secondary predictions---those that maintain systematically low confidence despite accurate localization---as they fail to contribute correctly to the assigned objects.
(2) Missing primary predictions: OCE penalizes subsets that are too aggressive and discard primary predictions, leaving ground-truth objects unmatched and incurring a maximum penalty.
Thus, unlike existing prediction-level ECE metrics, OCE effectively measures how well the post-processed subset approximates the set of reliable predictions.}

\statement{\rev{Remark (complexity).}}
\rev{Computing OCE only requires IoU checks between GT boxes and predicted boxes; precomputing an $N_i \times M$ IoU matrix per image makes the additional cost negligible relative to standard evaluation.
}

\begin{revision}
\subsection{Numerical Analysis}

\begin{table*}[t!]
\centering
\small

\caption{
\begin{revision}
The Pearson correlation coefficient between the proposed OCE and established calibration metrics, D-ECE and LA-ECE$_0$, on the optimal positive predictions ($\mathbf{^*}$) across six models---UP-DETR, D-DETR, Cal-DETR, DINO, Faster R-CNN, and YOLOv3---and three data distributions. 
The consistently very strong correlations confirm that OCE effectively captures the same underlying concept of calibration quality as D-ECE and LA-ECE$_0$.
\end{revision}
}
\resizebox{1.0\linewidth}{!}{
\begin{tabular}{ll|ccc|ccc|ccc|c}
\toprule
\multirow{2}{*}{\textbf{Predictions}}  & \multirow{2}{*}{\textbf{Metrics}} & \multicolumn{3}{c}{\texttt{\textbf{COCO (in-distribution)}}} & \multicolumn{3}{c}{\textbf{\texttt{Cityscapes  (near OOD)}}} & \multicolumn{3}{c|}{\textbf{\texttt{Foggy Cityscapes (OOD)}}} & \multirow{2}{*}{\textbf{\makecell[c]{Average \\ $\pm$ Std.}}} \\
\cmidrule(lr){3-5} \cmidrule(lr){6-8} \cmidrule(lr){9-11}
& & \textbf{D-ECE$^*$} & \textbf{LA-ECE$_0^*$} & \textbf{OCE$^*$} & \textbf{D-ECE$^*$} & \textbf{LA-ECE$_0^*$} & \textbf{OCE$^*$} & \textbf{D-ECE$^*$} & \textbf{LA-ECE$_0^*$} & \textbf{OCE$^*$} \\
\midrule
\midrule
\multirow{3}{*}{\makecell[l]{Optimal \\ Positives}} & D-ECE$^*$ & 1.000 & 0.983 & 0.791 & 1.000 & 0.963 & 0.967 & 1.000 & 0.890 & 0.989 & \textcolor{black}{0.954 $\pm$ 0.066} \\
 & LA-ECE$_0^*$ & 0.983 & 1.000 & 0.859 & 0.963 & 1.000 & 0.873 & 0.890 & 1.000 & 0.854 & \textcolor{black}{0.936 $\pm$ 0.061} \\
 & $\mathsf{OCE}^*$ & 0.791 & 0.859 & 1.000 & 0.967 & 0.873 & 1.000 & 0.989 & 0.854 & 1.000 & \textcolor{black}{0.926 $\pm$ 0.076} \\
\bottomrule
\end{tabular}
}
\label{tab:cal_ranking_opt}

\end{table*}

\subsubsection{Validating OCE as a Calibration Metric}
\label{sec:rank_opt}

Our first step is to validate that OCE is a rightful measure of calibration. We first perform a sanity check to ensure OCE fundamentally measures the same concept of calibration as established metrics. 
To do this, we isolate the metrics' behaviors from the confounding effects of different post-processing decisions. 
We therefore choose the optimal positive set. This set best represents the predictions that each model was \emph{specifically trained to produce as a calibrated predictor}.

\statement{\rev{Hypothesis.}} If OCE is a valid calibration metric, its ranking of models on this oracle optimal positive set should strongly correlate with the rankings produced by established metrics (D-ECE, LA-ECE$_0$). A lack of correlation would suggest OCE is measuring a completely different, unrelated property.

\statement{\rev{Method.}} We evaluate the calibration errors on the optimal positive set from our six models---UP-DETR, D-DETR, Cal-DETR, DINO, Faster-RCNN, and YOLOv3---using OCE, D-ECE, and LA-ECE$_0$. We then evaluate the models using each metric and compute the Pearson correlation coefficient (PCC) between metrics.

As shown in the first three rows in Table~\ref{tab:cal_ranking_opt}, the correlations are all extremely high. This robust finding provides a critical validation: \textbf{OCE captures the same underlying concept of calibration quality as established metrics.} This confirms OCE is a legitimate calibration metric, providing the necessary foundation for the next section, where we will demonstrate its superiority in the practical scenario where optimal positive set is not accessible.

\subsubsection{Pitfalls of Existing Metrics as a Reliable Set Selection Criteria}
\label{sec:metric_pitfalls}

We now analyze the behavior of existing metrics and the proposed OCE in a practical scenario: selecting a confidence threshold (i.e., post-processing configuration) that achieves the lowest calibration error for a given model.

To this end, this section analyzes the effectiveness of existing calibration metrics (D-ECE, LA-ECE$_0$, and our OCE) as well as performance metrics (AP \citep{salton1983introduction, everingham2010pascal,lin2014microsoft} and Localization recall precision (LRP) \citep{oksuz2018localization, oksuz2022one}) for this task.
Specifically, for each of the four different DETR models, we sweep the confidence score threshold from $0.0$ to $1.0$. This generates different positive subsets of predictions. We then plot the score of each metric for these subsets to determine the threshold that each metric prefers (i.e., the one that yields its optimal score).
Results on Cal-DETR are illustrated in Figure~\ref{fig:thr_metrics}.
For definitions of metrics and additional results on other DETR models, please refer to the Appendix~\ref{app:results}.
Our empirical findings reveal the following behaviors:

\statement{\rev{Pitfall 1: Average Precision Prefer No Filtering.}}
As noted in several studies \citep{oksuz2018localization, oksuz2023towards, kuzucu2024calibration, oksuz2022one}, the optimal AP is achieved when the threshold is set close to $0.0$. This is because AP does not harshly penalize low-confidence predictions.
However, as discussed so far, using the full prediction set (hundreds of predictions) is practically \revvv{nonsensical}. It carries a high risk of including uncalibrated negative predictions, diminishes interpretability, and leads to unreliable downstream decisions.

\statement{\rev{Pitfall 2: ECEs Prefer Discarding Almost All Predictions.}}
More importantly, the optimal ECEs (D-ECE and LA-ECE$_0$) are achieved when the threshold is set close to $1.0$. This behavior, which favors retaining only a tiny subset of high-confidence predictions, is a structural pitfall of prediction-level ECEs \citep{kuzucu2024calibration}.
The cause is simple: ECEs do not penalize missed predictions (i.e., false negatives).
As a result, they can achieve a near-zero (perfect) error by evaluating a set containing only a few highly accurate and confident predictions, while ignoring all the ground-truth objects the model missed at that high threshold.

Therefore, this pitfall makes ECEs fundamentally unsuitable for the goal we defined: finding a reliable (model + post-processing) combination. If a practitioner uses D-ECE to select a threshold, the metric will misleadingly guide them to select a threshold near 1.0, which is nonsensical as it discards almost all predictions to achieve a low error. 

\begin{figure}[t]
    \centering
    \includegraphics[width=\linewidth]{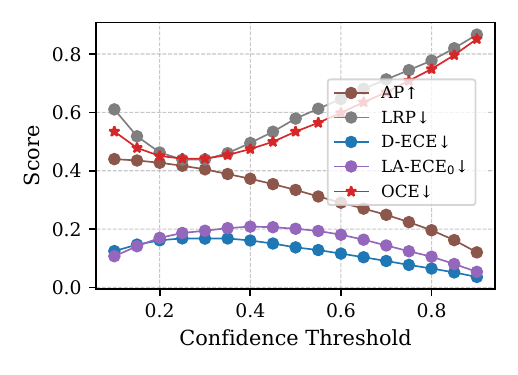}
    \caption{Impact of confidence threshold selection on various performance and error metrics in Cal-DETR evaluated on \texttt{COCO}. Higher scores are preferable ($\uparrow$), while lower scores are preferable ($\downarrow$).}
    \label{fig:thr_metrics}
\end{figure}

\statement{\rev{A Note on LRP: Good for Localization, Not Calibration.}}
On the other hand, LRP, a localization-focused performance metric, does appear to find a sensible, non-extreme optimal threshold.
This is rational: optimal positives are trained to be well-localized predictors, while optimal negatives are relatively not. 
LRP, by focusing purely on localization quality, thus correctly identifies a threshold that filters out the poorly-localized predictions.
However, LRP is not designed to measure calibration and does not consider the confidence quality of the predictions at all. 
As a result, a model ranking scored by LRP reflects localization performance, which is not necessarily aligned with the model's confidence calibration qualities.
\end{revision}

\subsubsection{The effectiveness of OCE}
The biggest advantage of using OCE over other metrics is that \textbf{OCE is capable of measuring the model’s calibration quality alongside the post-processing scheme employed.} 
Resultingly, as illustrated in Figures~\ref{fig:thr_metrics} and \ref{fig:thr_appendix}, the confidence threshold vs.\ OCE curve is U-shaped and the optimal OCE is achieved around a threshold of $0.3$.
This is particularly notable because it reflects the practical choices in many calibration studies to date \citep{kuppers2020multivariate,munir2022towards,munir2023bridging,pathiraja2023multiclass,munir2024cal,munir2023domain}, confirming the reliability of using OCE.

Using a fixed separation threshold of $0.3$ might serve as a reasonable approximation for distinguishing between optimal positives and negatives. However, trained or post-calibrated DETR models often have varying optimal confidence thresholds. For example, UP-DETR has an optimal threshold of approximately $0.5$, and low-temperature calibrated models are likely to require even higher thresholds. Consequently, employing a fixed threshold poses potential risks.

\begin{figure*}[t!]
    \centering
    \subfloat[Threshold Optimizing AP]{%
    \hspace{3em}
        \includegraphics[width=0.2\textwidth, trim={0 0 0 60pt}, clip]{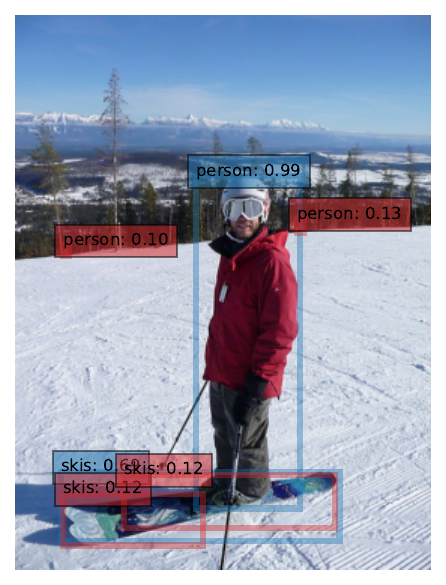}
    \hspace{2em}
    }
    \subfloat[Threshold Optimizing OCE]{%
    \hspace{2em}
        \includegraphics[width=0.2\textwidth, trim={0 0 0 60pt}, clip]{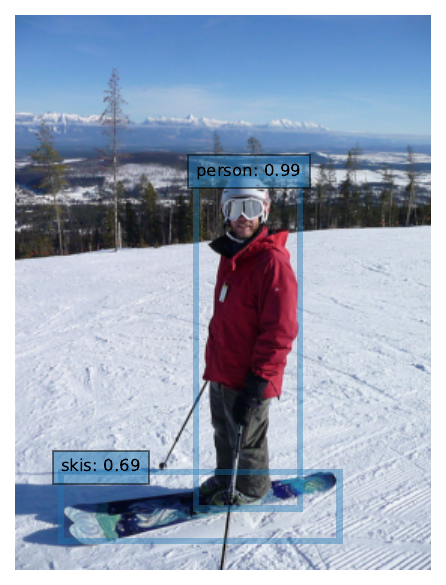}
    \hspace{2em}
    }
    \subfloat[Threshold Optimizing D-ECE]{%
    \hspace{2em}
        \includegraphics[width=0.2\textwidth, trim={0 0 0 60pt}, clip]{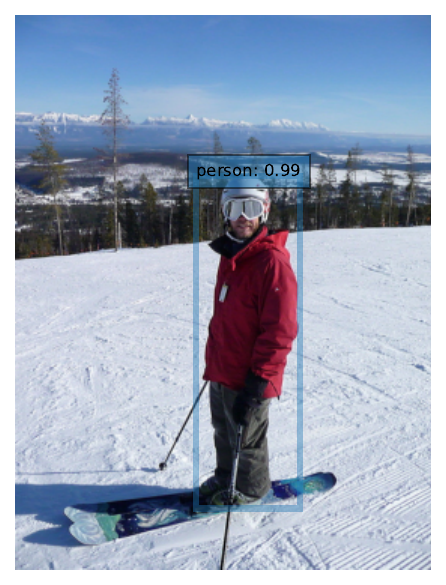}
    \hspace{3em}
    }
    
    \caption{\rev{Exemplary visualization demonstrating the impact of confidence threshold ($\rho$) selection on the final prediction subset in Cal-DETR across different post-processing configurations. Optimal positive and negative predictions are highlighted with blue and red boxes, respectively. The threshold selected by AP (e.g., $\rho=0.1$) retains redundant predictions, mixing calibrated positives with uncalibrated negatives. The threshold selected by D-ECE (e.g., $\rho=0.9$) discards nearly all detections, including the optimal positive prediction with moderate confidence, resulting in a missed detection. The proposed OCE identifies a practical threshold (e.g., $\rho=0.27$) that balances precision and recall while maintaining calibrated predictions.}}
    \label{fig:separation_visualization_calibration}
\end{figure*}

\begin{figure}[t]
    \centering
    \includegraphics[width=\linewidth]{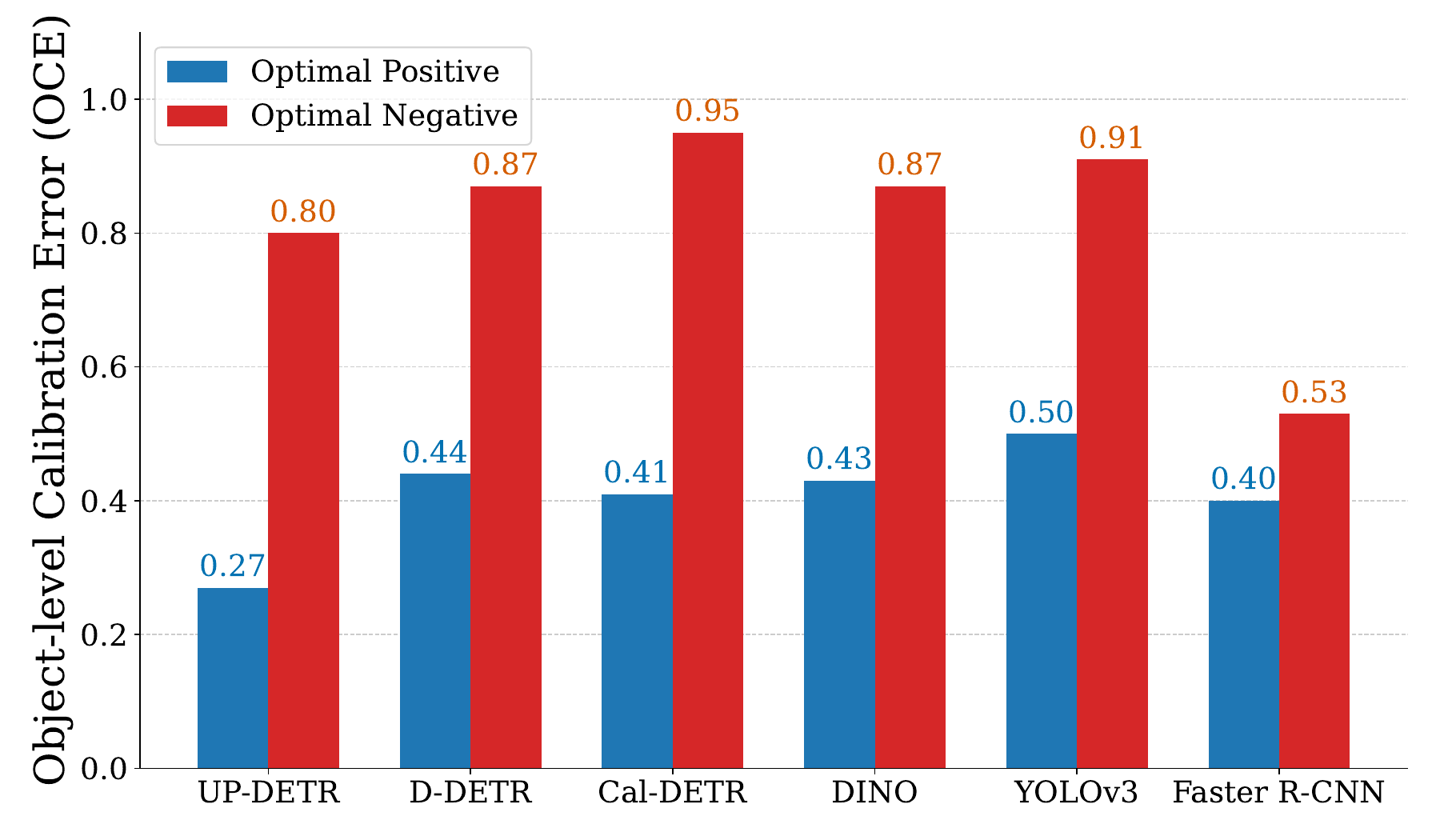}
    \caption{
    A visualization of the difference in calibration between optimal positive and negative predictions on the \texttt{COCO} dataset. Object-level Calibration Error (OCE) value of each set is shown, where a lower score represents better-calibrated predictions. As shown, OCE varies significantly between optimal positive and negative predictions, indicating that OCE is a useful metric for assessing the reliability of post-processing schemes.}
    \label{fig:obj_cal_opt}
\end{figure}

\begin{figure}[t]
    \centering
    \includegraphics[width=\linewidth]{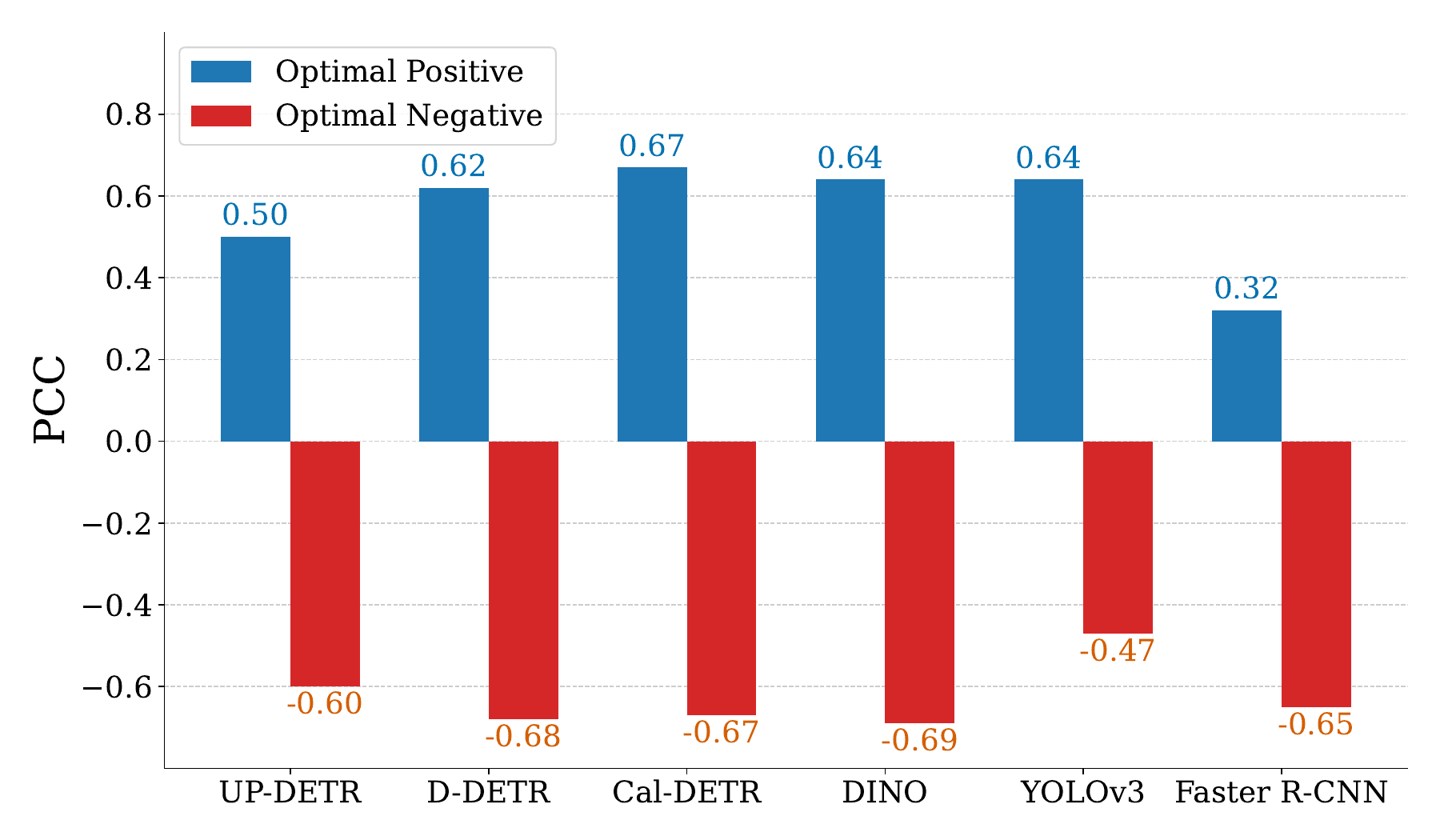}
    \caption{
    Pearson Correlation Coefficient (PCC) between per-image AP scores and per-image average foreground confidence scores on optimal positive and negative predictions (higher values indicate stronger correlation). As expected, optimal positive predictions show high positive correlation between confidence and performance. More interestingly, optimal negative predictions exhibit an \emph{inverse correlation} with image-level reliability.}
    \label{fig:imreli_opt}
\end{figure}

\begin{revision2}
\subsubsection{Ablation studies on OCE}
\label{sec:oce_abl}
We first conduct an ablation study on the choice of aggregation methods in OCE. Specifically, we compare different aggregation strategies as alternatives to averaging in Equation~\eqref{eq:oce_agg}.
First, the Max-IoU aggregation selects the prediction that has the largest overlap with the ground-truth box:
$$
q^{*} = \arg\max_{q \in Q} \text{IoU}(b_{\text{gt}}, \hat{b}_q), \quad \bar{p}(c) = \hat{p}_{q^{*}}(c) .
$$

Second, we consider an IoU-weighted averaging:
$$
w_q = \frac{\text{IoU}(b_{\text{gt}}, \hat{b}_q)}{\sum_{q'} \text{IoU}(b_{\text{gt}}, \hat{b}_{q'})}, \quad \bar{p}(c) = \sum_{q \in Q} w_q \cdot \hat{p}_q(c) .
$$

After computing these OCE variants across six models and three data distributions, we compare their Pearson correlation coefficients. As shown in the Table~\ref{tab:oce_abl} in the Appendix, all three aggregation variants are nearly perfectly correlated with one another. This confirms that OCE is robust to the specific choice of aggregation rule for overlapping predictions.

We then conduct an ablation study on the choice of IoU threshold $\tau \in [0.1, 0.9]$ using the default averaging method. The results in Figure~\ref{fig:oce_iou_abl} in the Appendix demonstrate that 1) adjacent thresholds are highly correlated with each other as expected, and 2) they generally correlate well except for extreme values (e.g., $\tau = 0.9$). This empirically validates the overall robustness to the choice of IoU threshold and confirms that ensembling $\tau \in \{0.5, 0.75\}$ serves as a reliable default. We note that the ensembling strategy aligns with well-established evaluation protocols, which often average over IoU thresholds of $0.5$ and $0.75$ for metrics such as AP \citep{lin2014microsoft} and D-ECE \citep{kuzucu2024calibration}.
\end{revision2}

\section{Quantifying Image-level Reliability} \label{sec:uq}

\begin{revision}
In this section, we revisit the importance of separation between primary and secondary predictions at test-time, now from the perspective of image-level reliability estimation.

Recall that our theoretical analysis (Proposition~\ref{prop:undercal}) shows that when the model is uncertain about its prediction---expressed by matching probability $P_i<1$---it suppresses the predictive distribution. Reflecting this observation to the secondary prediction, we find that its foreground confidence score is no longer zero. Consequently, we expect the separation between these predictions to become blurred under uncertainty, and our empirical observations confirm this.

Specifically, for \textbf{reliable} images, we observe that the confidence of optimal positive predictions increases across decoder layers while optimal negative predictions remain low, resulting in a \textbf{large gap between optimal positive and negative predictions} (e.g., Figure~\ref{fig:insight-reli-horse}). In contrast, for \textbf{unreliable} instances, optimal positive prediction confidence fails to increase across layers, while optimal negative prediction scores are either slightly elevated or unchanged. This produces a \textbf{small gap between optimal positive and negative predictions} (e.g., Figure~\ref{fig:insight-unreli-chair}).

To further empirically verify this, we first obtain the image-wise AP score by using the same COCO evaluator \citep{lin2014microsoft}, passing the predictions and annotations for each image individually rather than the entire image set.
We use this value as a proxy of image-level reliability (See Definition~\ref{defn::reli} for more formal interpretation.)
Then, we measure the Pearson correlation coefficient between the image-wise AP and the average confidence score of the optimal positives and the optimal negatives, respectively.

As shown in Figure~\ref{fig:imreli_opt}, the average confidence scores of the optimal positive predictions exhibit a moderately strong positive correlation with image-level reliability (i.e., the model's actual average precision on that image) across different models.
Notably, the average confidence scores for negative predictions are \emph{inversely} correlated with image-level reliability.
This finding reinforces our claim, highlighting the importance of identifying primary and secondary predictions, when conducting image-level uncertainty quantification (UQ) in DETR. 
\end{revision}

\subsection{Image-Level Reliability}
We introduce a formal definition of \emph{image-level reliability} by examining the model's overall object detection performance on the image, extending \citep{ardeshir2022embedding,park2024quantifying}.

\begin{defn} \label{defn::reli}
We define image-level reliability as a measure of how closely the predictions align with the ground-truth annotations for a given test image $x$:
\begin{equation} \label{eq:imgreli}
\mathsf{ImReli}(x; \theta) \defined \mathsf{Perf} \big( \hat{\cD}_\theta(x), ~ \cD \big) ,
\end{equation}
where one can use any standard performance metric, such as average precision, for $\mathsf{Perf}$ depending on the user's requirements.
\end{defn}

By its definition, image-level reliability directly addresses the model’s applicability to a given test instance.
However, since image-level reliability requires ground-truth annotations for its determination, obtaining it during inference is not feasible. 
Therefore, this paper's objective is to develop a method that assigns a quantitative score to each image instance, closely aligning with the model's image-level reliability.

\subsection{Proposed Method: Quantifying Reliability by Contrasting} 

Based on our findings, we propose a post hoc UQ approach by contrasting the confidence scores of positives and negatives:
\begin{gather*}
    \approach(x) = \approachpos(x) - \lambda \approachneg(x) \label{eq:ContrastiveReli}, \\ 
    \approachpos(x) = \frac{1}{|\hat{\mathcal{S}}_\theta(x)|} \sum\nolimits_{(\hat{p}, \hat{b}) \in \hat{\mathcal{S}}_\theta(x)} \max_{c \in \cC} \hat{p}(c), \\
    \approachneg(x) = \frac{1}{|\hat{\mathcal{S}}_\theta^{-}(x)|} \sum\nolimits_{(\hat{p}, \hat{b}) \in \hat{\mathcal{S}}_\theta^{-}(x)} \max_{c \in \cC} \hat{p}(c),
\end{gather*}

where $\hat{\mathcal{S}}_\theta(x)$ and $\hat{\mathcal{S}}_\theta^{-}(x) \defined \hat{\cD}_\theta(x)\setminus \hat{\mathcal{S}}_\theta(x)$ are predicted sets, which we refer to \emph{positive} and \emph{negative} predictions, respectively.
$\lambda$ is a scaling factor that can be determined on the validation set; we include an ablation study of the scaling factor in the subsequent section. 
Our results demonstrate that the proposed method is robust to the choice of scaling factor and consistently outperforms baseline approaches, achieving the robust performance with a scaling factor of $5.0$ - $10.0$.

\begin{revision}
\statement{Selecting positives at test time.}
Ground-truth matches are unavailable at test time, so we approximate the optimal positive set $\mathcal{S}_\theta(x)$ by applying a post-processing operator $\cA$ (e.g., confidence thresholding, Top-$k$, or NMS).
We choose $\cA^*$ on the validation set to \emph{minimize} OCE:

\begin{gather*}
    \cA^* = \arg\min_{\cA}~\mathrm{OCE}\big( \cA \!\circ\! \hat{\cD}_\theta(\cX_{\text{val}}),\, \cD_{\text{val}}\big),
\end{gather*}
where $\cD_{\text{val}}$ is the set of all ground-truth objects across the images $\cX_{\text{val}}$ in validation dataset.
\end{revision}

\begin{table*}[t!]
\centering
\small
\caption{
Comparison of the proposed method ($\approach$) with baseline methods on their Pearson correlation coefficient with the image-level reliability in Equation~\eqref{eq:imgreli}, with different models.
$\mathsf{Perf}$ is evaluated using AP. %
For each model, we apply the optimal matching, the standard post-processing scheme (Top-$100$ and confidence thresholding by $0.3$), or the proposed OCE-based post-processing scheme, as indicated within the parentheses. 
The strongest correlations are highlighted in bold, while negative correlations appear in red.
First, the proposed contrasting method, $\approach$, achieves the highest correlation with image-level reliability.
In addition, it is noteworthy that the average confidence of negative predictions ($\approachneg$) exhibits a negative correlation with image-level reliability, in contrast to the positive correlation observed when using positive predictions ($\approachpos$).
}
\renewcommand{\arraystretch}{1.0}
\resizebox{1.0\linewidth}{!}{
\begin{tabular}{ll|cccc|cccc|cccc}
\toprule
& \multirow{2}{*}{\textbf{Methods}} & \multicolumn{4}{c}{\texttt{\textbf{COCO (in-distribution)}}} & \multicolumn{4}{c}{\textbf{\texttt{Cityscapes  (near OOD)}}} & \multicolumn{4}{c}{\textbf{\texttt{Foggy Cityscapes (OOD)}}}  \\
\cmidrule(lr){3-6} \cmidrule(lr){7-10} \cmidrule(lr){11-14}
& & \textbf{UP-DETR} & \textbf{D-DETR} & \textbf{Cal-DETR} & \textbf{DINO} & \textbf{UP-DETR} & \textbf{D-DETR} & \textbf{Cal-DETR} & \textbf{DINO} & \textbf{UP-DETR} & \textbf{D-DETR} & \textbf{Cal-DETR} & \textbf{DINO} \\
\midrule
\midrule
\multirow{3}{*}{Oracles} & $\approachpos$ (Optimal) &  0.503 &  0.618 &  0.670 &  0.635 &  0.555 &  0.647 &  0.649 &  0.633 &  0.561 &  0.642 &  0.662 &  0.634 \\
 & $\approachneg$ (Optimal) &  \textcolor{red}{-0.608} &  \textcolor{red}{-0.584} &  \textcolor{red}{-0.572} &  \textcolor{red}{-0.586} &  \textcolor{red}{-0.293} &  \textcolor{red}{-0.296} &  \textcolor{red}{-0.330} &  \textcolor{red}{-0.311} &  \textcolor{red}{-0.177} &  \textcolor{red}{-0.235} &  \textcolor{red}{-0.212} &  \textcolor{red}{-0.196} \\
 & $\approach$ (Optimal) & \bf 0.700 & \bf 0.648 & \bf 0.684 & \bf 0.662 & \bf 0.601 & \bf 0.659 & \bf 0.656 & \bf 0.644 & \bf 0.612 & \bf 0.660 & \bf 0.667 & \bf 0.647 \\
\midrule
\midrule
\multirow{2}{*}{Baselines} & $\approachpos$ (Top-100) &  \textcolor{red}{-0.619} &  \textcolor{red}{-0.603} &  \textcolor{red}{-0.581} &  \textcolor{red}{-0.601} &  \textcolor{red}{-0.409} &  \textcolor{red}{-0.374} &  \textcolor{red}{-0.372} &  \textcolor{red}{-0.391} &  \textcolor{red}{-0.262} &  \textcolor{red}{-0.270} &  \textcolor{red}{-0.239} &  \textcolor{red}{-0.229} \\
 & $\approachpos$ (0.3) &  0.476 &  0.464 &  0.539 &  0.504 &  0.229 &  0.353 &  0.385 &  0.399 &  0.195 &  0.256 &  0.258 &  0.258 \\
\midrule
\multirow{3}{*}{Ours} & $\approachpos$ (OCE) &  0.477 &  0.478 &  0.547 &  0.512 &  0.252 &  0.330 &  0.393 &  0.370 &  0.191 &  0.276 &  0.270 &  0.274 \\
 & $\approachneg$ (OCE) &  \textcolor{red}{-0.640} &  \textcolor{red}{-0.575} &  \textcolor{red}{-0.571} &  \textcolor{red}{-0.585} &  \textcolor{red}{-0.349} &  \textcolor{red}{-0.382} &  \textcolor{red}{-0.379} &  \textcolor{red}{-0.394} &  \textcolor{red}{-0.256} &  \textcolor{red}{-0.322} &  \textcolor{red}{-0.253} &  \textcolor{red}{-0.283} \\
 & $\approach$ (OCE) & \bf 0.656 & \bf 0.588 & \bf 0.628 & \bf 0.613 & \bf 0.359 & \bf 0.407 & \bf 0.441 & \bf 0.440 & \bf 0.275 & \bf 0.350 & \bf 0.317 & \bf 0.327 \\
 \bottomrule
\end{tabular}
}

\label{tab:imreli_full}
\end{table*}

\subsection{Numerical Analysis} \label{sec:uq-img}

The model is tested on three datasets with varying levels of out-of-distribution (OOD) characteristics: \texttt{COCO} (in-distribution), \texttt{Cityscapes} (near OOD), and \texttt{Foggy Cityscapes} (OOD).

We compare different methods for quantifying per-image reliability based on different separation methods.
For separation methods, we apply the optimal matching (as a reference) as well as the practical post-processing schemes such as fixed and adaptive confidence thresholding.
We measure the Pearson correlation coefficient of each method with the $\mathsf{ImReli}$ computed using the standard AP metric.

Table~\ref{tab:imreli_full} shows that the proposed $\approach$ consistently achieves the best correlation with image-level reliability. Interestingly, the absolute value of $\approachneg$ often surpasses that of $\approachpos$ when a non-optimal separation scheme is applied. This observation highlights how our contrasting approach, which leverages the strengths of both $\approachpos$ and $\approachneg$, achieves robust performance across diverse settings.

\begin{figure*}[t]
    \centering
    \subfloat[\texttt{COCO}]{
        \includegraphics[width=0.32\textwidth]{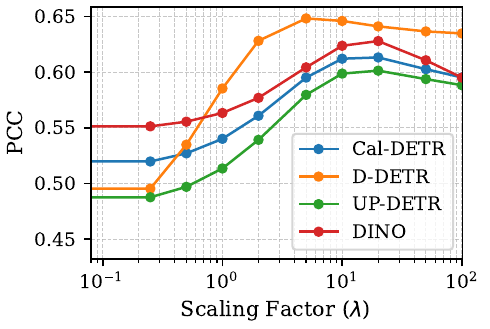}
    }
    \subfloat[\texttt{Cityscapes}]{
        \includegraphics[width=0.32\textwidth]{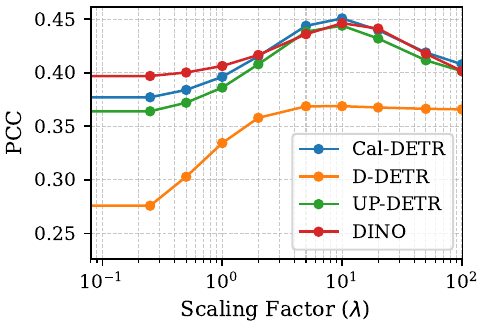}
    }
    \subfloat[\texttt{Foggy Cityscapes}]{
        \includegraphics[width=0.32\textwidth]{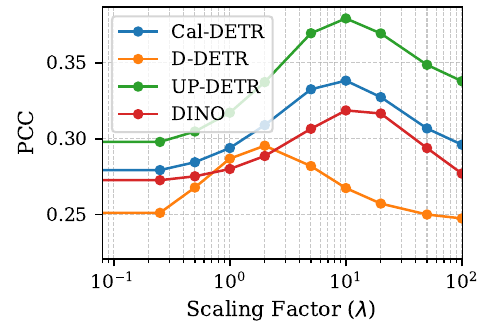}
    }
    
    \caption{Impact of the scaling factor ($\lambda$) on image-level UQ performance of $\approach$ (OCE). Pearson correlation coefficient (PCC) using various scaling factors is reported on (a) In-distribution (\texttt{COCO}), (b) Near OOD (\texttt{Cityscapes}), and (c) Far OOD (\texttt{Foggy Cityscapes}). The optimal scaling factor lies within the range of $5.0$ to $10.0$, while this range generalizes well across out-of-distribution datasets. Furthermore, it shows the efficacy of $\approach$ over $\approachpos$ (i.e., $\approach$ with $\lambda=0.0$).}

    \label{fig:scaling_ablation}
\end{figure*}

In addition, we conduct an ablation study on the scaling factor ($\lambda$), with results presented in Figure~\ref{fig:scaling_ablation}. The study reveals that the best performing scaling factor for \texttt{COCO} dataset lies between $5.0$ and $10.0$. Notably, this range remains effective even for out-of-distribution datasets such as \texttt{Cityscapes} and \texttt{Foggy Cityscapes}. Furthermore, the results demonstrate that $\approach$ with $1.0 \leq \lambda \leq 10.0$ consistently outperforms $\approachpos$ (i.e., $\lambda = 0.0$, the leftmost point on each line). That said, we also observe a rapid drop in performance when the scaling factor is excessively large, particularly on the OOD dataset, \texttt{Foggy Cityscapes}.
Thus, we do not recommend using large scaling factors or relying solely on the negative predictions.

\section{Conclusion}

The main contribution of our work is an in-depth analysis of the reliability of DETR frameworks. In particular, we reveal that DETR's predictions for a given image exhibit varying calibration quality, highlighting the importance of identifying well-calibrated predictions. To address this challenge, we introduced a systematic framework that leverages our object-level calibration error metric to discern these reliable predictions effectively. Furthermore, we proposed a novel uncertainty quantification method for estimating image-level reliability in DETR and conducted comparative studies of various post-processing schemes regarding their impact on DETR’s reliability.
We hope our efforts expand the scope of DETR applications by enabling more precise and reliable deployment, sparking further research into the area of reliable positive prediction identification.

\section*{Acknowledgments}
 The authors acknowledge the MIT SuperCloud and Lincoln Laboratory Supercomputing Center for providing computing resources that have contributed to the results reported within this paper. This work was supported in part by the MIT Summer Research Program (MSRP), the MIT-IBM Watson AI Lab, the MIT-Google Program for Computing Innovation, the MIT-Amazon Science Hub, Netflix, and Jane Street.

\bibliographystyle{IEEEtran}
\bibliography{IEEEabrv,refs}

\begin{IEEEbiography}[{\includegraphics[width=1in,height=1.25in,clip,keepaspectratio]{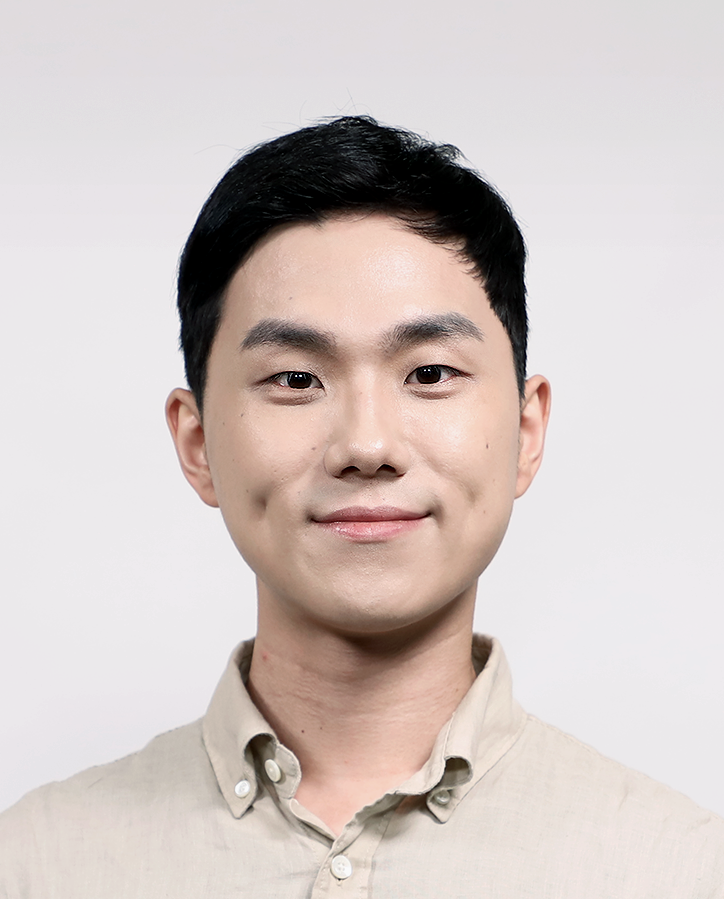}}]
{Young-Jin Park} is completing his Ph.D. in Mechanical Engineering at the Massachusetts Institute of Technology in 2026. He conducted his research in the Laboratory for Information and Decision Systems (LIDS) under the supervision of Prof. Navid Azizan. His doctoral research focuses on uncertainty quantification and instance-level reliability of deep learning models, with particular emphasis on foundation models for computer vision and language. He received his B.S. and M.S. degrees in Aerospace Engineering from KAIST in 2017 and 2019, respectively; his M.S. thesis received the Outstanding Paper Award from the Department of Aerospace Engineering. Prior to his doctoral studies, he spent three years as a Research Engineer at NAVER AI Lab, building large-scale recommender systems and demand forecasting models serving millions of users. During his Ph.D., he collaborated with the MIT-IBM Watson AI Lab and interned at Mitsubishi Electric Research Laboratories and Meta in 2024 and 2025, respectively. His work has been recognized with the Wunsch Foundation Award for excellence in graduate research at MIT, the Shangzhi Wu Fellowship, and the KAIST Presidential Fellowship.
\end{IEEEbiography}

\begin{IEEEbiography}[{\includegraphics[width=1in,height=1.25in,clip,keepaspectratio]{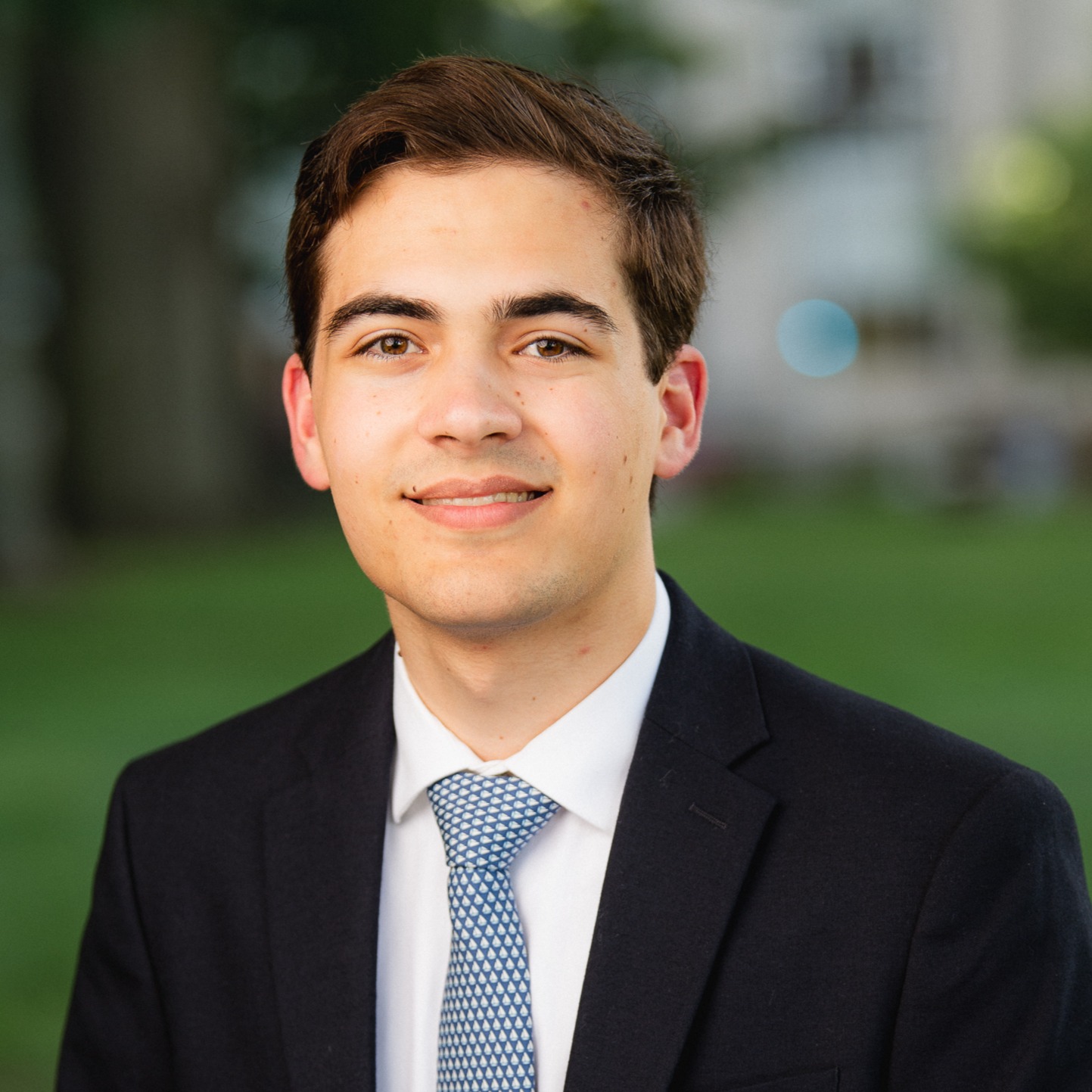}}]{Carson Sobolewski} is currently an S.M. student at the Massachusetts Institute of Technology (MIT) in the Laboratory for Information and Decision Systems (LIDS) and the Department of Aeronautics and Astronautics (AeroAstro). Previously, he received his B.S. degree in Computer Engineering from the University of Florida. In 2024, he interned at MIT, where he contributed to this project.
His research interests include uncertainty quantification and safe autonomous systems.
\end{IEEEbiography}

\begin{IEEEbiography}[{\includegraphics[width=1in,height=1.25in,clip,keepaspectratio]{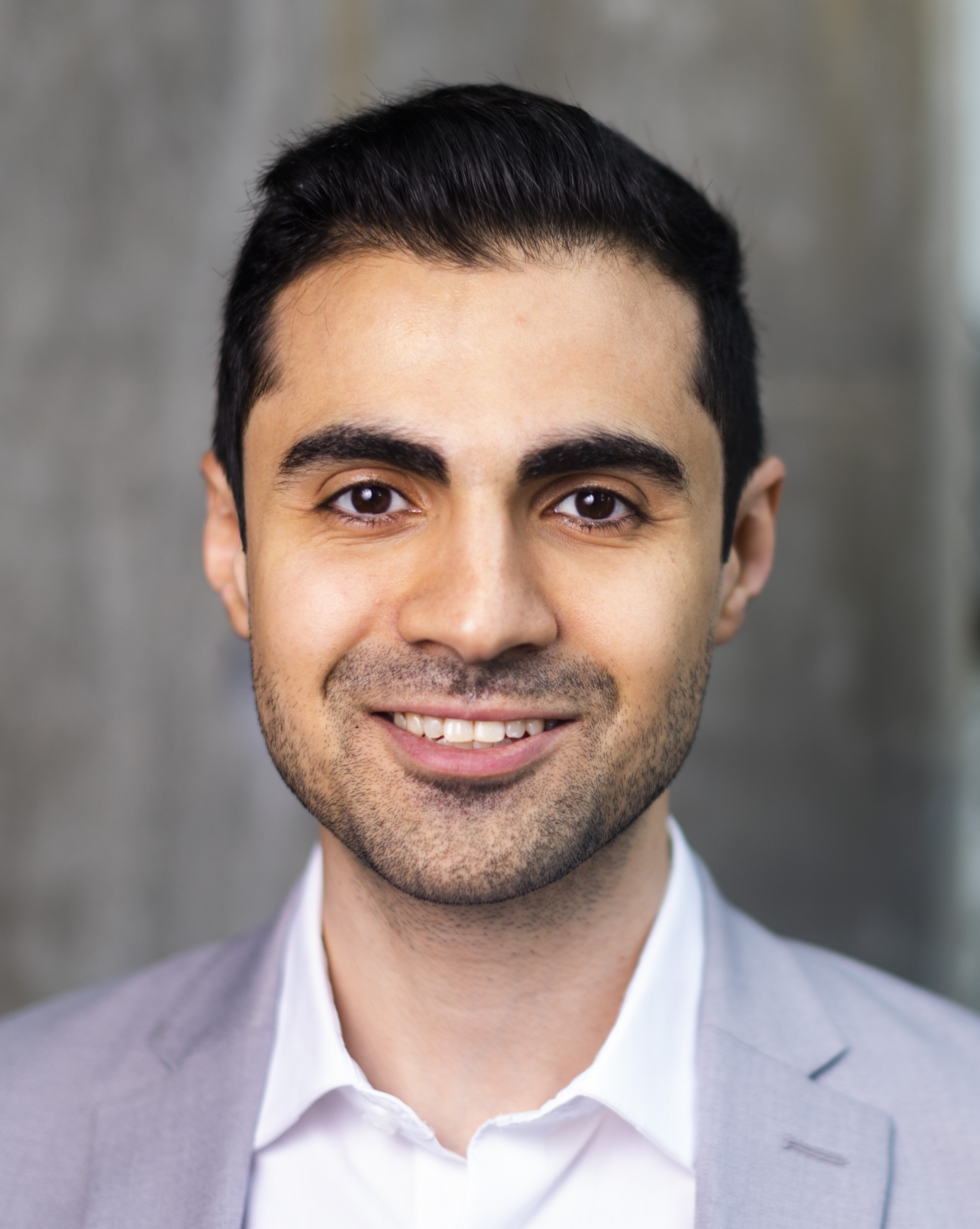}}]{Navid Azizan} is currently the Alfred H. (1929) and Jean M. Hayes Associate Professor at MIT, in LIDS, MechE, and IDSS. He obtained his PhD in Computing and Mathematical Sciences (CMS) from the California Institute of Technology (Caltech) in 2020, his MSc in electrical engineering from the University of Southern California in 2015, and his BSc in electrical engineering with a minor in physics from Sharif University of Technology in 2013. Prior to joining MIT, he completed a postdoc at Stanford University in 2021. Additionally, he was a research scientist intern at Google DeepMind in 2019.  His research interests broadly lie in machine learning, systems and control, and mathematical optimization. His work has been recognized by several awards, including the National Science Foundation CAREER Award, research awards from Google, Amazon, MathWorks, and IBM, and best paper awards and nominations at several venues, including ACM Greenmetrics, the Learning for Dynamics \& Control (L4DC), and INFORMS JFIG. He was named in the list of Outstanding Academic Leaders in Data by the CDO Magazine for two consecutive years in 2024 and 2023, received the 2020 Information Theory and Applications (ITA) “Sun” (Gold) Graduation Award, and was named an Amazon Fellow in AI in 2017 and a PIMCO Fellow in Data Science in 2018. His mentorship has been recognized with the Frank E. Perkins Award for Excellence in Graduate Advising (MIT Institute Award) in 2025 and the UROP Outstanding Mentor Award in 2023. Early in his academic journey, he was the first-place winner and a gold medalist at the 2008 National Physics Olympiad in Iran.
\end{IEEEbiography}

\clearpage

\appendices
\setcounter{page}{1}
\pagestyle{appendixstyle}

\begin{figure*}[t!]
\begin{center}
    {\large Supplementary Material for}\\[0.5em]
    {\huge Uncertainty Quantification in Detection Transformers:}\\[0.3em]
    {\huge Object-Level Calibration and Image-Level Reliability}\\[0.8em]
    {\large Young-Jin Park$^*$, Carson Sobolewski$^*$, and Navid Azizan}\\[0.2em]
    {\large Massachusetts Institute of Technology}\\[0.1em]
    {\small\tt \{youngp,csobo,azizan\}@mit.edu}\\[0.2em]
\end{center}
\vspace{0.5em}
\hrule
\end{figure*}

\section{Dataset} \label{app:setup}
For our experiments, we used the \texttt{Cityscapes} and \texttt{Foggy Cityscapes} datasets, which each have 500 images of first-person driving footage in realistic environments. \texttt{Foggy Cityscapes} has the same base images as \texttt{Cityscapes}, but with fog simulated and added to create a further out of distribution set. Since the DETR models were trained on \texttt{COCO}, the \texttt{Cityscapes} and \texttt{Foggy Cityscapes} annotations were converted to correspond to the labels of \texttt{COCO}. More specifically, the person, bicycle, car, motorcycle, bus, train, and truck classes were transferred directly. In addition, the rider class of \texttt{Cityscapes} and \texttt{Foggy Cityscapes} was mapped to the person class of \texttt{COCO}. The other classes present in \texttt{Cityscapes} and \texttt{Foggy Cityscapes} are largely focused on image segmentation, and thus were omitted (e.g. building, sky, sidewalk).
The pre-trained model weights were obtained from their respective official implementations.

\begin{revision}
\section{Detailed Theoretical Proofs}
\label{app:proofs}

\setcounter{prop}{0}

\subsection{Proof of Proposition~\ref{prop:bayes}} \label{proof:bayes}
\begin{prop}
The minimum achievable expected total loss is lower bounded by the entropy of the ground-truth class distribution, i.e., $$\mathbb{E}\left[ \Ltotal \right] \ge \entropy(p_{\text{gt}}) + \min_{\hat{b}} \mathbb{E}[\Lbox(\hat{b}, b_{\text{gt}})],$$ where $\entropy(p_{\text{gt}}) = -\sum_{c \in \cset} p_{\text{gt}}(c) \log(p_{\text{gt}}(c))$.
\end{prop}
\begin{proof}
The expected total loss is the sum of the expected loss for the optimal positive query matched to $y_{\text{gt}}$ (denoted as index $p$) and the expected losses for the $M-1$ remaining queries matched to $\emptyset$ (denoted as indices $\{n \ne p\}$):
$$
\mathbb{E}[\Ltotal] = \mathbb{E}[\mathcal{L}(\hat{y}_p, y_{\text{gt}})] + \sum_{n \ne p} \mathcal{L}(\hat{y}_n, \emptyset).
$$
From Equation~\ref{eq:rev_neg_loss}, $\mathcal{L}(\hat{y}_n, \emptyset) = -w_{\emptyset} \log(\hat{p}_n(\emptyset))$; and since $\log(\hat{p}_n(\emptyset)) \le 0$: $\mathcal{L}(\hat{y}_n, \emptyset) \ge 0$. Therefore, the total expected loss is bounded below by:
$$
\mathbb{E}[\Ltotal] \ge \mathbb{E}[\mathcal{L}(\hat{y}_p, y_{\text{gt}})].
$$
From Equation~\ref{eq:rev_pos_loss}, we have:
$$
\mathbb{E}[\mathcal{L}(\hat{y}_p, y_{\text{gt}})] = \underbrace{- \sum_{c \in \cset} p_{\text{gt}}(c) \log(\hat{p}_{p}(c))}_{\mathbb{E}[\Lcls]} + \mathbb{E}_{b_{\text{gt}} \sim \mathcal{B}} [\Lbox(\hat{b}_p, b_{\text{gt}})].
$$
We minimize the two terms on the right-hand side independently.
First, we find the minimum classification loss $\mathbb{E}[\Lcls]$. We define $S = \sum_{c \in \cset} \hat{p}_p(c) = 1 - \hat{p}_p(\emptyset)$ and $\hat{q}_p(c) = \hat{p}_p(c) / S$. Then:
$$
\begin{aligned}
\mathbb{E}[\Lcls] &= - \sum_{c \in \cset} p_{\text{gt}}(c) (\log(S) + \log(\hat{q}_p(c))) \\
&= - \log(S) + \mathrm{CE}(p_{\text{gt}}, \hat{q}_p),
\end{aligned}
$$
where $\mathrm{CE}(p_{\text{gt}}, \hat{q}_p) = -\sum_{c \in \cset} p_{\text{gt}}(c) \log(\hat{q}_p(c))$ is the cross-entropy.
By Gibbs' inequality, $\mathrm{CE}(p_{\text{gt}}, \hat{q}_p) \ge \entropy(p_{\text{gt}})$ and equality holds if and only if $\hat{q}_p = p_{\text{gt}}$.
Additionally, $-\log(S) \ge 0$ and equality holds if and only if $S=1$ (i.e., $\hat{p}_p(\emptyset)=0$).

Denoting the minimum box loss as $\min_{\hat{b}} \mathbb{E}[\Lbox(\hat{b}, b_{\text{gt}})]$, a lower bound of $\mathbb{E}_{b_{\text{gt}} \sim \mathcal{B}} [\Lbox(\hat{b}_p, b_{\text{gt}})]$, we have:
$$
\mathbb{E}[\Ltotal] \ge \entropy(p_{\text{gt}}) + \min_{\hat{b}} \mathbb{E}[\Lbox(\hat{b}, b_{\text{gt}})].
$$
\end{proof}

\subsection{Proof of Proposition~\ref{prop:specialist}} \label{proof:specialist}
\begin{prop}[Optimal Prediction Set]
\revv{Let $\{(\hat{p}^*_\qidx, \hat{b}^*_\qidx)\}_{\qidx=1}^{M}$ denote a globally optimal prediction set, i.e., that minimizes $\mathbb{E}\left[ \Ltotal \right]$. Then we have}
\begin{enumerate}
    \item for the optimal positive prediction ($\qidx=p$): $\hat{p}^*_\qidx = p_{\text{gt}}$ (i.e., $\hat{p}^*_{\qidx}(c) = p_{\text{gt}}(c)$ $\forall c \in \cset$ and $\hat{p}^*_{\qidx}(\emptyset)=0$) and the box $\hat{b}^*_\qidx$ minimizes $\mathbb{E}[\Lbox(\hat{b}_\qidx, b_{\text{gt}})]$, and
    \item for any optimal negative prediction ($\qidx \ne p$): $\hat{p}^*_{\qidx}(c)=0$ $\forall c \in \cset$ and $\hat{p}^*_{\qidx}(\emptyset)=1$.
\end{enumerate}
\end{prop}
\begin{proof}
We evaluate the expected total loss for the proposed strategy $\{(\hat{p}^*_\qidx, \hat{b}^*_\qidx)\}_{\qidx=1}^{M}$:
\begin{enumerate}
    \item For the optimal positive prediction: The strategy sets $\hat{p}^*_p = p_{\text{gt}}$ and $\hat{b}^*_p$ to minimize the expected box loss. The resulting expected loss is:
    $$
    \begin{aligned}
    \mathbb{E}[\mathcal{L}(\hat{y}^*_p, y_{\text{gt}})] &= \min_{\hat{b}} \mathbb{E}[\Lbox(\hat{b}, b_{\text{gt}})] - \sum_{c \in \cset} p_{\text{gt}}(c) \log(p_{\text{gt}}(c)) \\
    &= \min_{\hat{b}} \mathbb{E}[\Lbox(\hat{b}, b_{\text{gt}})] + \entropy(p_{\text{gt}}).
    \end{aligned}
    $$
    \item For any optimal negative prediction: The strategy sets $\hat{p}^*_n(\emptyset) = 1$. The resulting expected loss is:
    $$ \mathcal{L}(\hat{y}^*_n, \emptyset) = -w_{\emptyset} \log(1) = 0. $$
\end{enumerate}
Summing these terms, the strategy achieves the lower bound derived in Proposition~\ref{prop:bayes}:
$$ \mathbb{E}[\Ltotal] = \min_{\hat{b}} \mathbb{E}[\Lbox(\hat{b}, b_{\text{gt}})] + \entropy(p_{\text{gt}}). $$
Any deviation from this strategy results in a strictly higher loss:
\begin{itemize}
    \item If $\hat{p}_n(\emptyset) < 1$, then $-w_{\emptyset}\log(\hat{p}_n(\emptyset)) > 0$.
    \item If $\hat{b}_p$ is not the minimizer, the box loss term increases.
    \item If $\hat{p}_p \ne p_{\text{gt}}$, then either $-\log(S) > 0$ or $\mathrm{CE}(p_{\text{gt}}, \hat{q}_p) > \entropy(p_{\text{gt}})$ (by Gibbs' inequality).
\end{itemize}
Thus, the proposed prediction set is the global minimizer.
\end{proof}

\subsection{Proof of Proposition~\ref{prop:undercal}} \label{proof:undercal}
\begin{prop}[Optimal Strategy under Uncertainty]
For a query with a matching probability $P_\qidx$ to the ground-truth object whose class distribution is $p_{\text{gt}}$, the class prediction that minimizes the expected loss is $\hat{p}_\qidx^*$, where
\begin{enumerate}
    \item the prediction for any foreground class $c \in \cset$ is:
    \begin{equation*}
    \hat{p}_\qidx^*(c) = p_{\text{fg}}^*(P_\qidx) p_{\text{gt}}(c)
    .\end{equation*}
    \item the prediction for the background class is $$\hat{p}_\qidx^*(\emptyset) = 1 - p_{\text{fg}}^*(P_\qidx),$$
\end{enumerate}
where $p_{\text{fg}}^*(P_\qidx)$ is the optimal foreground probability, defined as
\begin{equation*}
p_{\text{fg}}^*(P_\qidx) = \frac{P_\qidx}{P_\qidx + w_{\emptyset}(1-P_\qidx)}.
\end{equation*}
\end{prop}
\begin{proof}
We find the optimal probability distribution $\hat{p}_\qidx$ by minimizing the expected classification objective:
$$
J(\hat{p}_\qidx) = -P_\qidx \sum_{c \in \cset} p_{\text{gt}}(c) \log(\hat{p}_\qidx(c)) - w_{\emptyset}(1-P_\qidx) \log(\hat{p}_\qidx(\emptyset)).
$$
This is a constrained optimization problem, as we must satisfy the constraint that:
$$
g(\hat{p}_\qidx) = \sum_{c \in \cset} \hat{p}_\qidx(c) + \hat{p}_\qidx(\emptyset) - 1 = 0.
$$
We form the Lagrangian $\mathcal{J}$ with a Lagrange multiplier $\lambda$:
$$
\mathcal{J} = J(\hat{p}_\qidx) - \lambda g(\hat{p}_\qidx).
$$
We find the optimum by setting the partial derivatives of $\mathcal{J}$ with respect to each probability $\hat{p}_\qidx(c)$ to zero.

For any foreground class $c \in \cset$:
$$
\frac{\partial \mathcal{J}}{\partial \hat{p}_\qidx(c)} = 0 \implies \hat{p}_\qidx(c) = -\frac{P_\qidx \cdot p_{\text{gt}}(c)}{\lambda}.
$$
    
For the background class $\emptyset$:
$$
\frac{\partial \mathcal{J}}{\partial \hat{p}_\qidx(\emptyset)} = 0 \implies \hat{p}_\qidx(\emptyset) = -\frac{w_{\emptyset}(1-P_\qidx)}{\lambda}.
$$

We solve for $\lambda$ by substituting these back into the constraint $g(\hat{p}_\qidx) = 0$:
$$
\sum_{c \in \cset} \left( -\frac{P_\qidx \cdot p_{\text{gt}}(c)}{\lambda} \right) + \left( -\frac{w_{\emptyset}(1-P_\qidx)}{\lambda} \right) = 1.
$$
Since $\sum_{c \in \cset} p_{\text{gt}}(c) = 1$ by definition:
$$
-\frac{P_\qidx}{\lambda} - \frac{w_{\emptyset}(1-P_\qidx)}{\lambda} = 1 \implies \lambda = -(P_\qidx + w_{\emptyset}(1-P_\qidx)).
$$
Finally, we substitute this $\lambda$ back to find the optimal predictions $\hat{p}_\qidx^*$:
$$
\hat{p}_\qidx^*(c) = -\frac{P_\qidx \cdot p_{\text{gt}}(c)}{-(P_\qidx + w_{\emptyset}(1-P_\qidx))} = \left( \frac{P_\qidx}{P_\qidx + w_{\emptyset}(1-P_\qidx)} \right) p_{\text{gt}}(c) ,
$$
and
$$
\hat{p}_\qidx^*(\emptyset) = -\frac{w_{\emptyset}(1-P_\qidx)}{-(P_\qidx + w_{\emptyset}(1-P_\qidx))} = 1 - \left( \frac{P_\qidx}{P_\qidx + w_{\emptyset}(1-P_\qidx)} \right).
$$
\end{proof}

\end{revision}

\begin{figure*}[t]
    \centering
    \subfloat[\texttt{COCO - 000000039769.jpg}  ]{
        \includegraphics[width=0.6\textwidth]{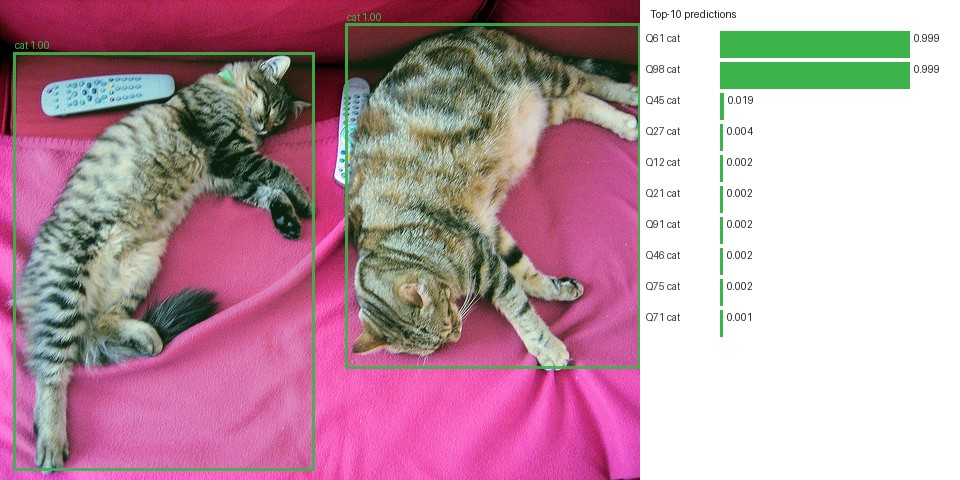}
    }
    
    \subfloat[\texttt{COCO - 000000033707.jpg} ]{
        \includegraphics[width=0.6\textwidth]{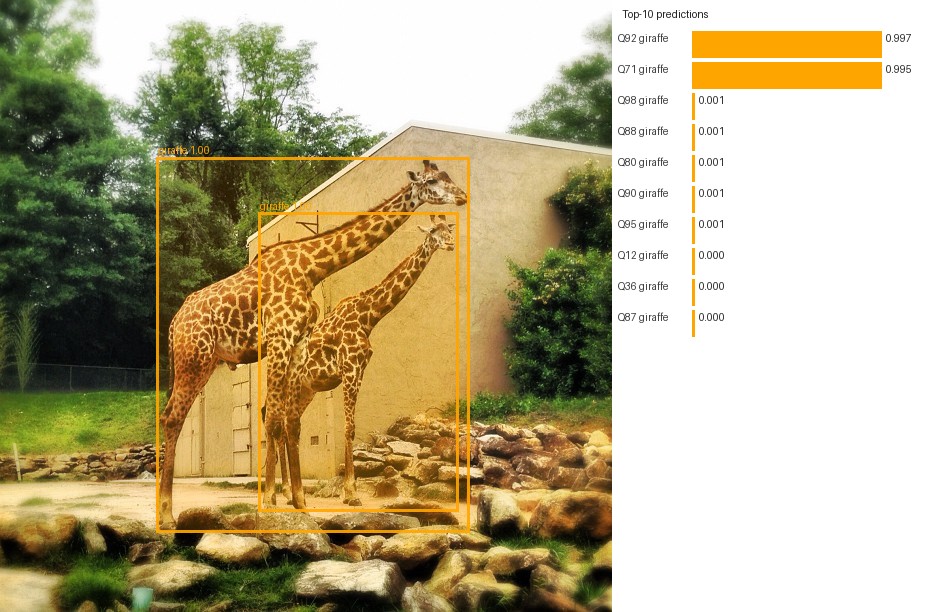}
    }
    
    \subfloat[\texttt{COCO - 000000000872.jpg} ]{
        \includegraphics[width=0.6\textwidth]{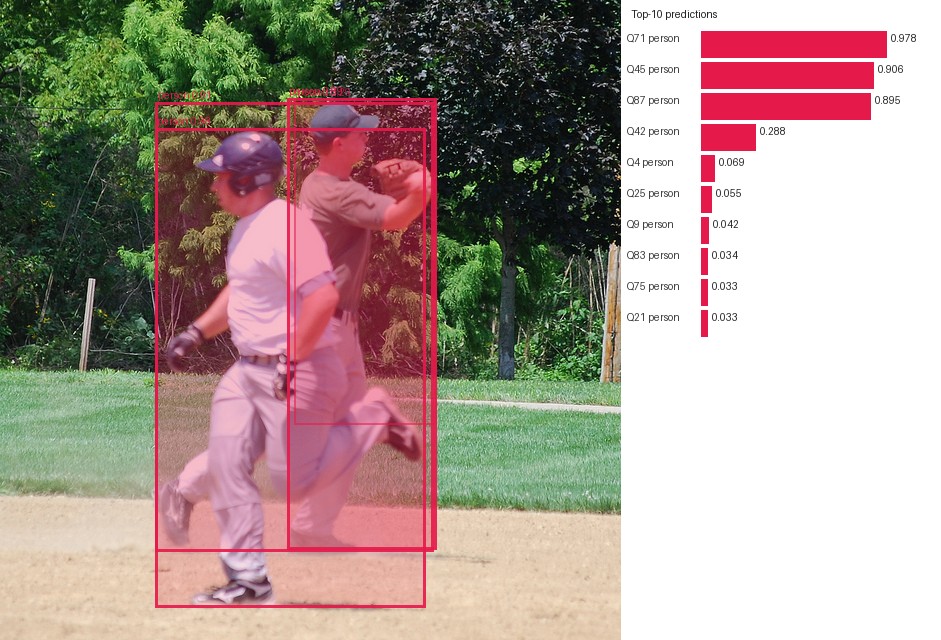}
    }
    
    \caption{\revv{The inference results with DETR on three COCO images. When instances are visually distinguishable (even if spatially overlapping), the specialist pattern holds. When instances are visually entangled (e.g., similar appearance, heavy occlusion), the model hedges with multiple mid-confidence predictions.}} \label{fig:detr_real}
\end{figure*}

\begin{figure*}[t]
    \centering
    \subfloat[\texttt{COCO - 000000039769.jpg}  ]{
        \includegraphics[width=0.6\textwidth]{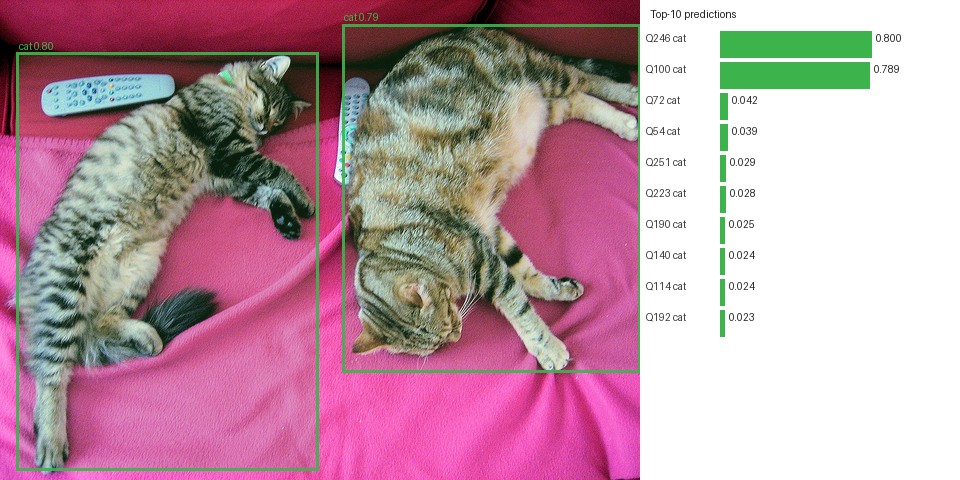}
    }
    
    \subfloat[\texttt{COCO - 000000033707.jpg} ]{
        \includegraphics[width=0.6\textwidth]{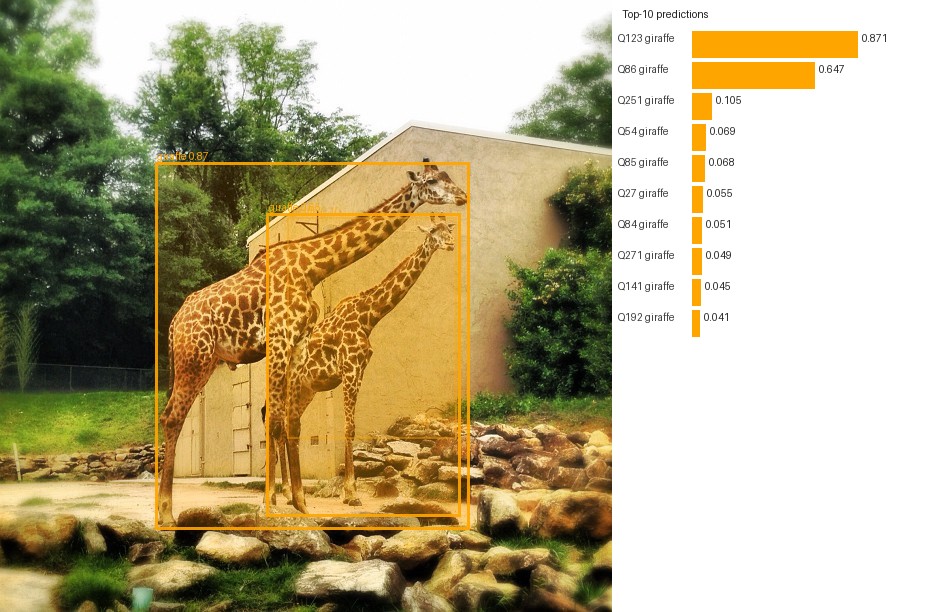}
    }
    
    \subfloat[\texttt{COCO - 000000000872.jpg} ]{
        \includegraphics[width=0.6\textwidth]{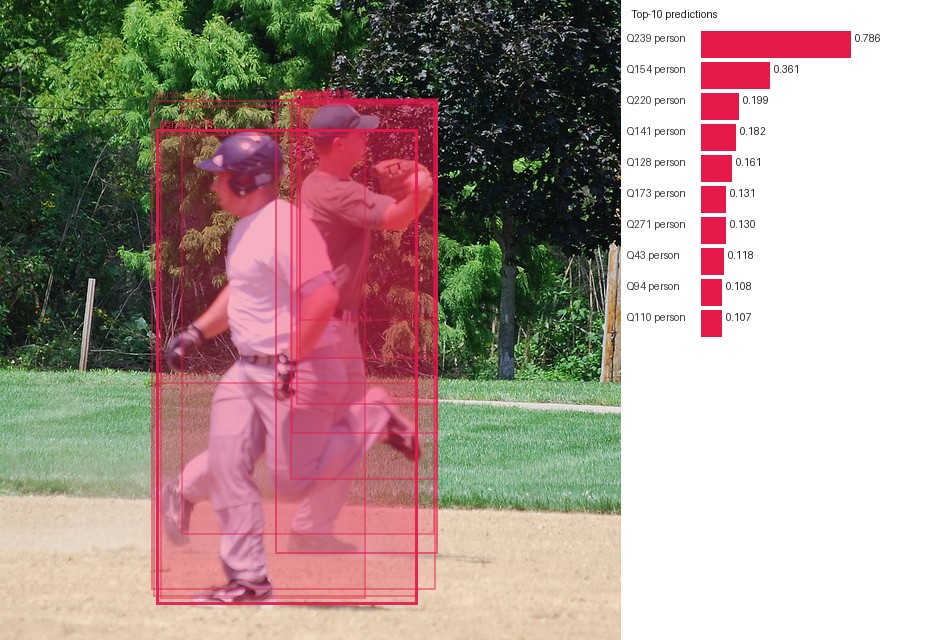}
    }
    
    \caption{\revv{The inference results with D-DETR on three COCO images. While the overall trend remains the same as in DETR (Figure~\ref{fig:detr_real}), D-DETR appears to hedge more than DETR.}}  \label{fig:ddetr_real}
\end{figure*}

\begin{figure*}[t]
    \centering
    \subfloat[\texttt{Two People (No Overlap)}  ]{
        \includegraphics[width=0.4\textwidth]{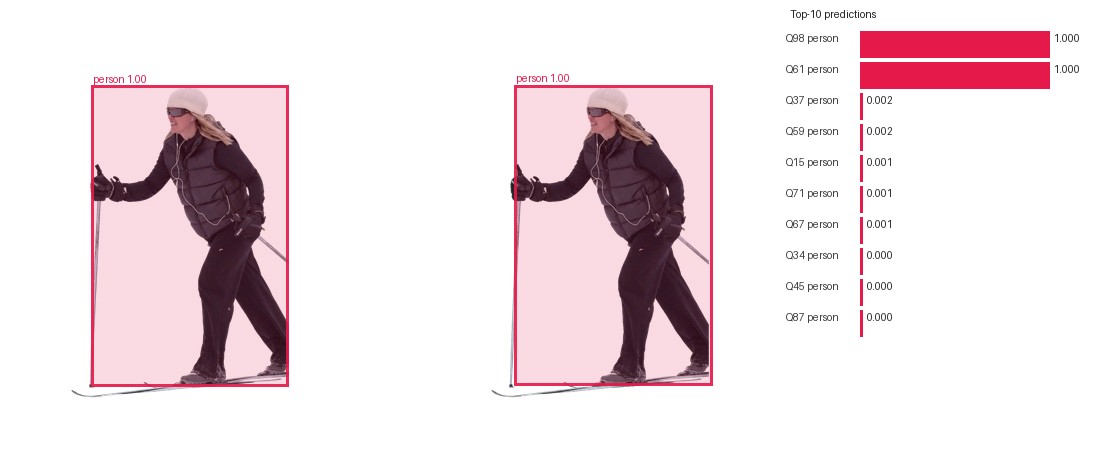}
    }
    \hspace{2em}
    \subfloat[\texttt{Two People (No Overlap)}  ]{
        \includegraphics[width=0.4\textwidth]{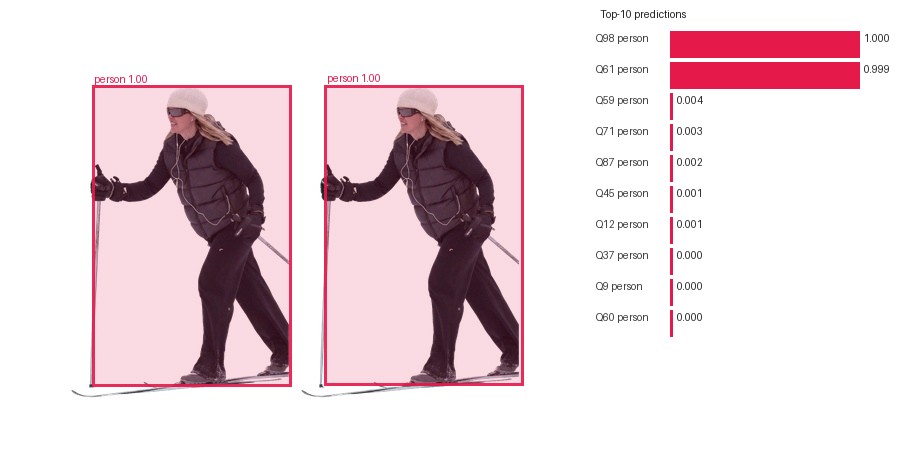}
    }
    
    \subfloat[\texttt{Two People (Small Overlap)}  ]{
        \includegraphics[width=0.4\textwidth]{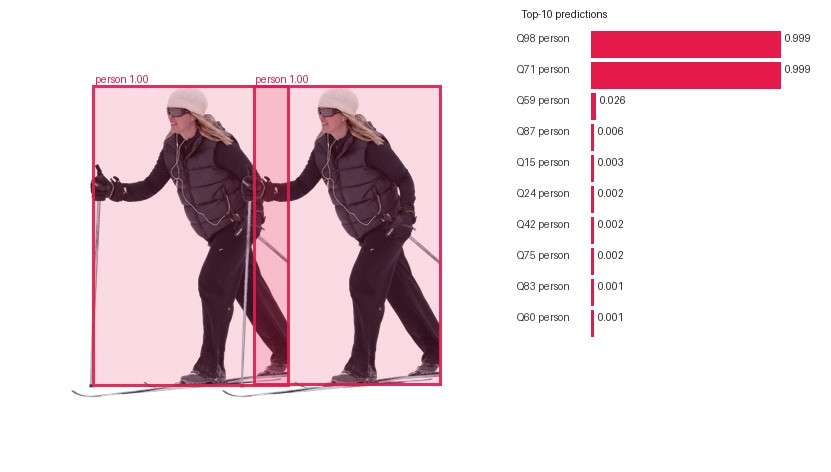}
    }
    \hspace{2em}
    \subfloat[\texttt{Two People (Small Overlap)}  ]{
        \includegraphics[width=0.4\textwidth]{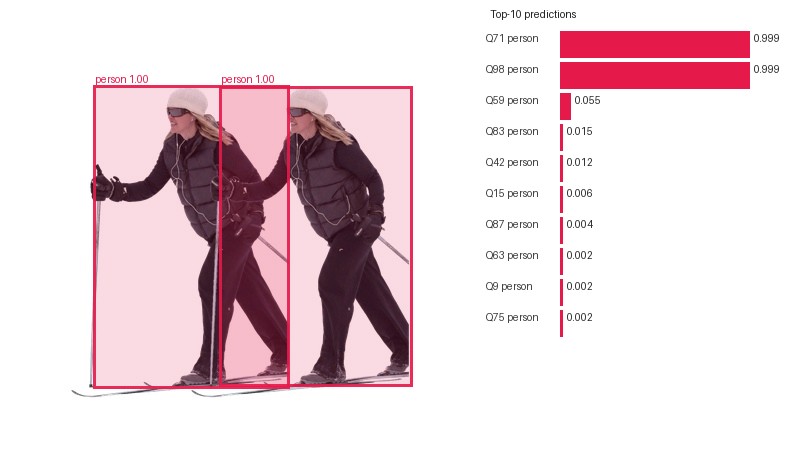}
    }

    \subfloat[\texttt{Two People (Medium Overlap)}  ]{
        \includegraphics[width=0.4\textwidth]{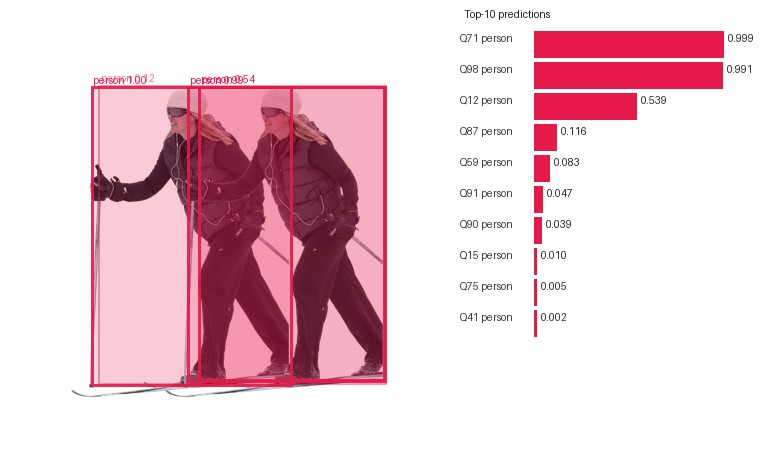}
    }
    \hspace{2em}
    \subfloat[\texttt{Two People (Medium Overlap)}  ]{
        \includegraphics[width=0.4\textwidth]{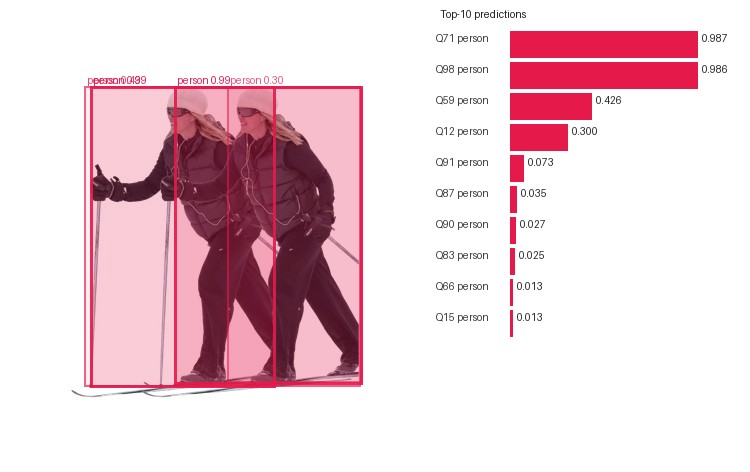}
    }
    
    \subfloat[\texttt{Two People (Large Overlap)}  ]{
        \includegraphics[width=0.4\textwidth]{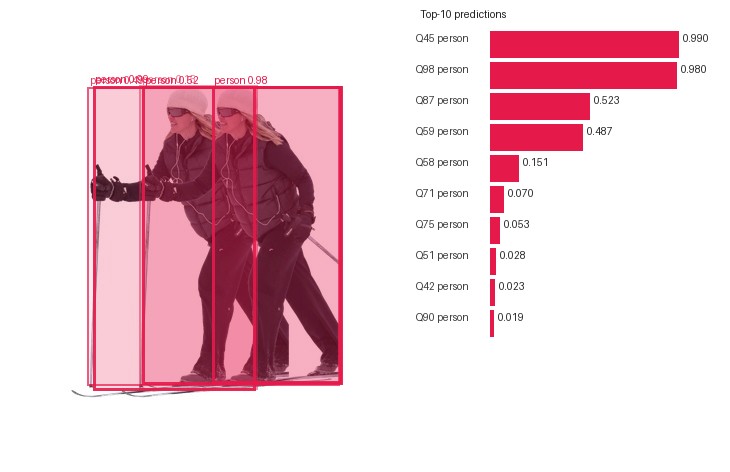}
    }
    \hspace{2em}
    \subfloat[\texttt{Two People (Large Overlap)}  ]{
        \includegraphics[width=0.4\textwidth]{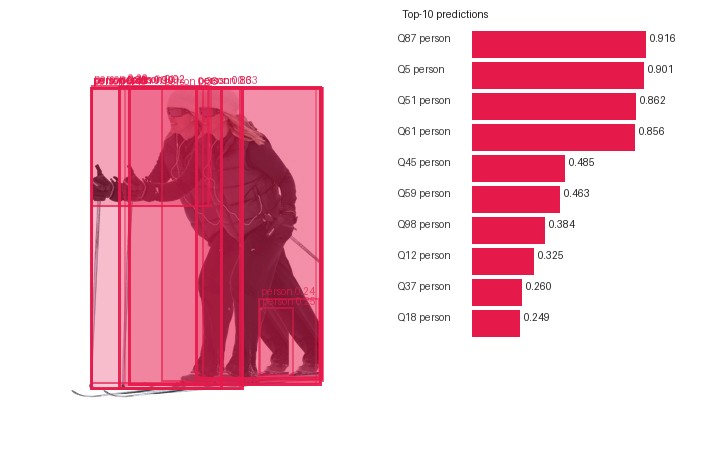}
    }
    
    \caption{\revv{Inference results of DETR on synthetic images containing two people with varying levels of overlap. The results show that the specialist pattern holds when instances are visually distinguishable, while entangled scenes lead to multiple mid-confidence predictions consistent with the hedging interpretation.}} \label{fig:detr_synthetic_person}
\end{figure*}

\begin{figure*}[t]
    \centering
    \subfloat[\texttt{Two People (No Overlap)}  ]{
        \includegraphics[width=0.4\textwidth]{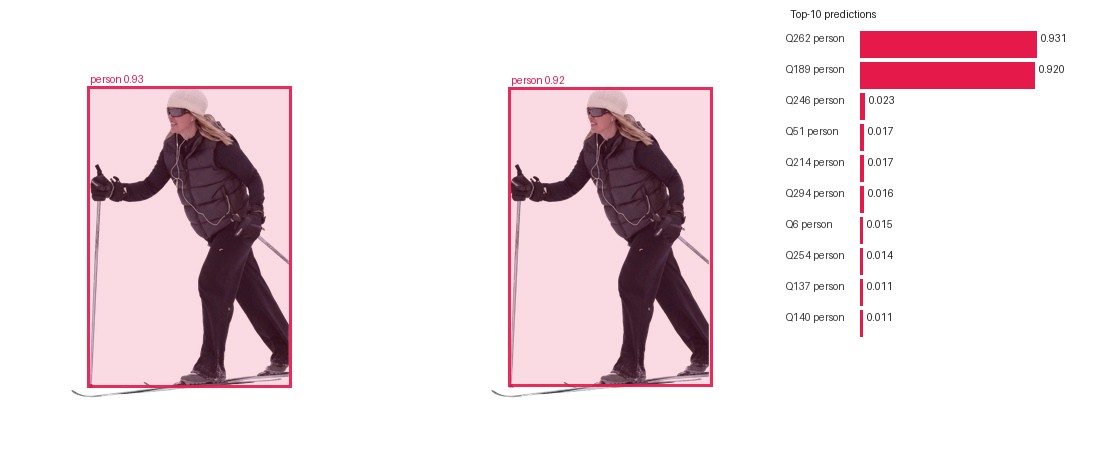}
    }
    \hspace{2em}
    \subfloat[\texttt{Two People (No Overlap)}  ]{
        \includegraphics[width=0.4\textwidth]{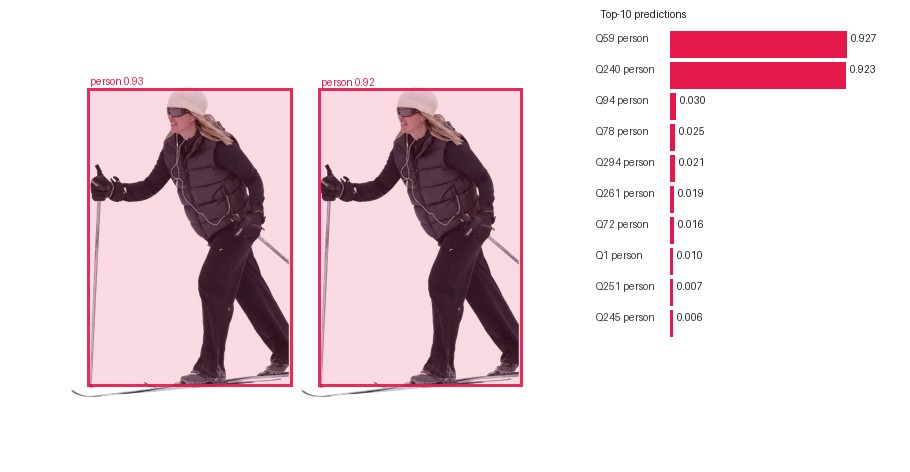}
    }
    
    \subfloat[\texttt{Two People (Small Overlap)}  ]{
        \includegraphics[width=0.4\textwidth]{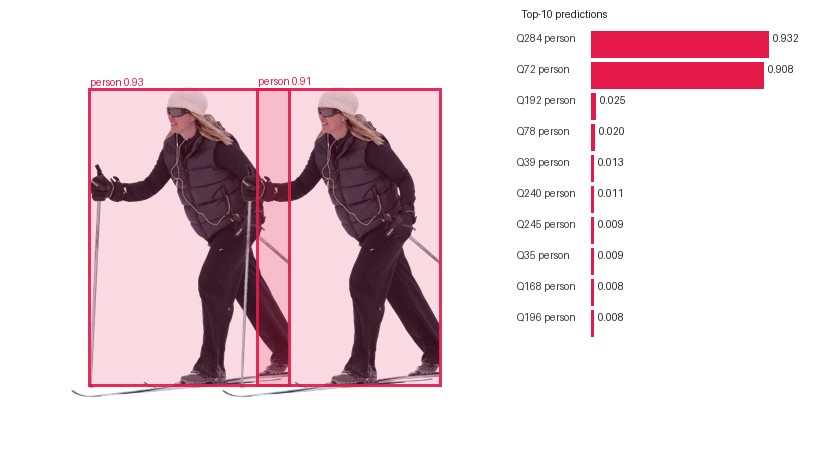}
    }
    \hspace{2em}
    \subfloat[\texttt{Two People (Small Overlap)}  ]{
        \includegraphics[width=0.4\textwidth]{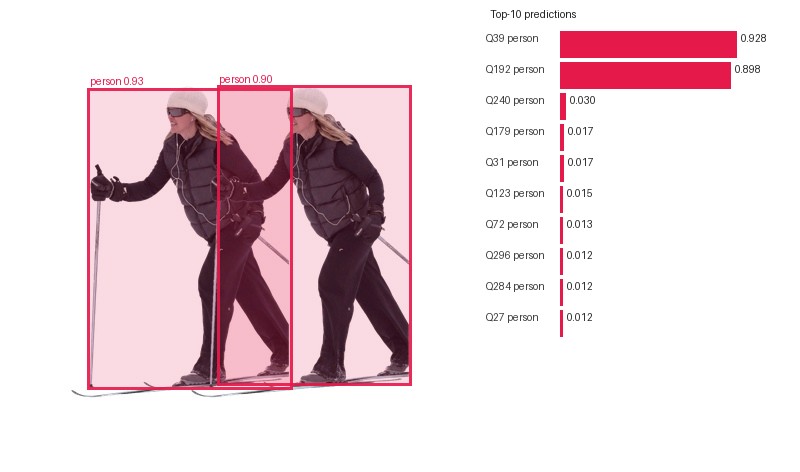}
    }

    \subfloat[\texttt{Two People (Medium Overlap)}  ]{
        \includegraphics[width=0.4\textwidth]{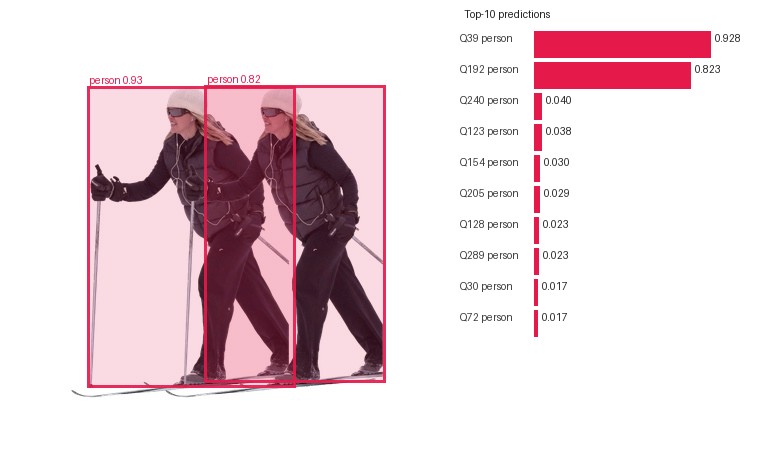}
    }
    \hspace{2em}
    \subfloat[\texttt{Two People (Medium Overlap)}  ]{
        \includegraphics[width=0.4\textwidth]{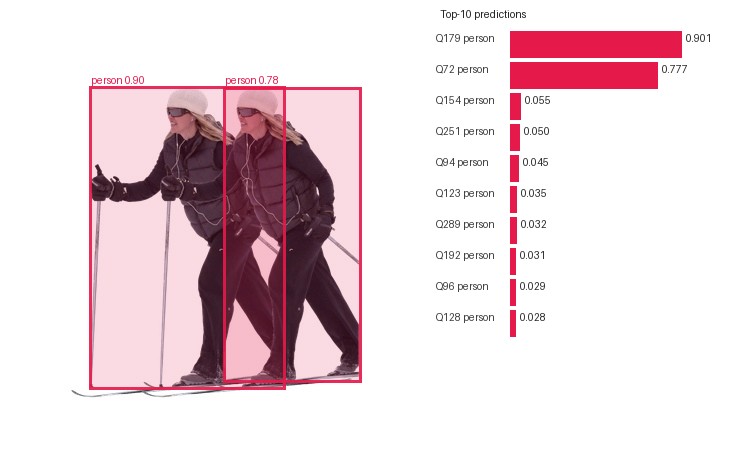}
    }
    
    \subfloat[\texttt{Two People (Large Overlap)}  ]{
        \includegraphics[width=0.4\textwidth]{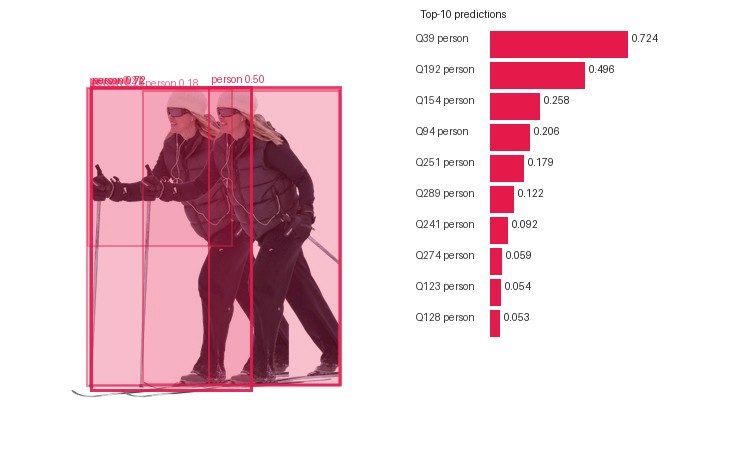}
    }
    \hspace{2em}
    \subfloat[\texttt{Two People (Large Overlap)}  ]{
        \includegraphics[width=0.4\textwidth]{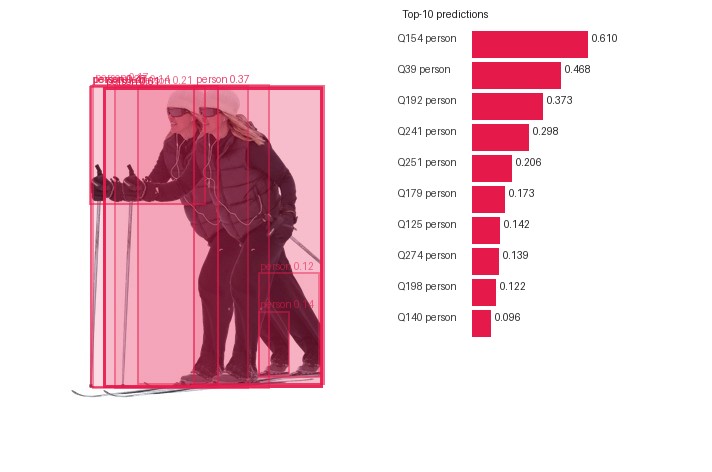}
    }
    
    \caption{\revv{Inference results of D-DETR on synthetic images containing two people with varying levels of overlap.}} \label{fig:ddetr_synthetic_person}
\end{figure*}

\begin{figure*}[t]
    \centering
    \subfloat[\texttt{Two Dogs (No Overlap)}  ]{
        \includegraphics[width=0.4\textwidth]{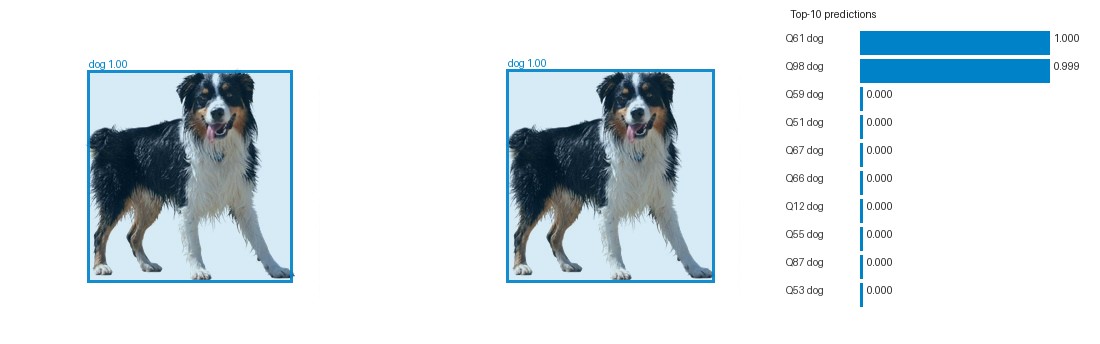}
    }
    \hspace{2em}
    \subfloat[\texttt{Two Dogs (No Overlap)}  ]{
        \includegraphics[width=0.4\textwidth]{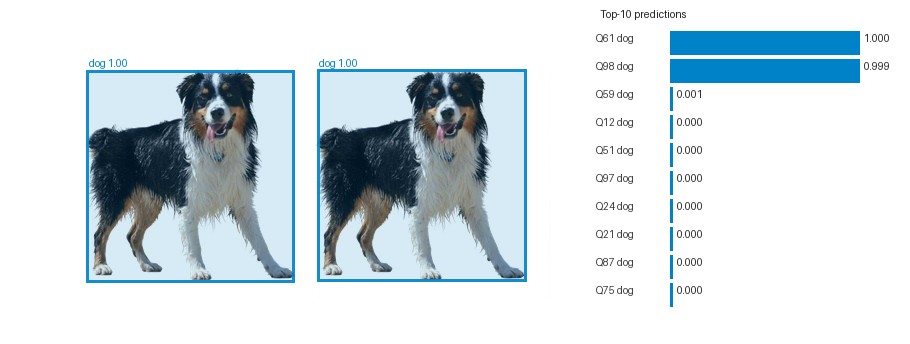}
    }
    
    \subfloat[\texttt{Two Dogs (Small Overlap)}  ]{
        \includegraphics[width=0.4\textwidth]{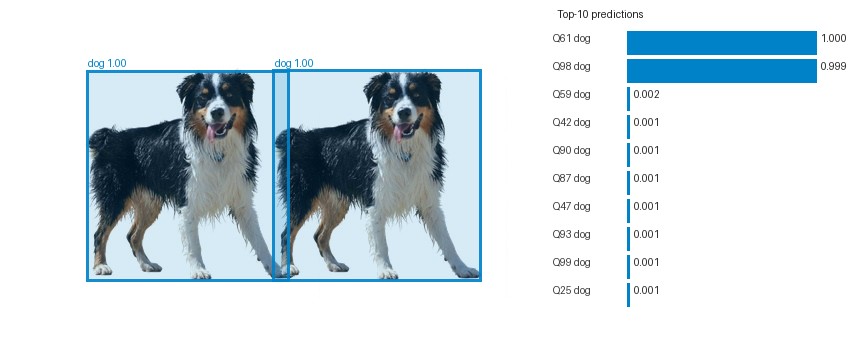}
    }
    \hspace{2em}
    \subfloat[\texttt{Two Dogs (Small Overlap)}  ]{
        \includegraphics[width=0.4\textwidth]{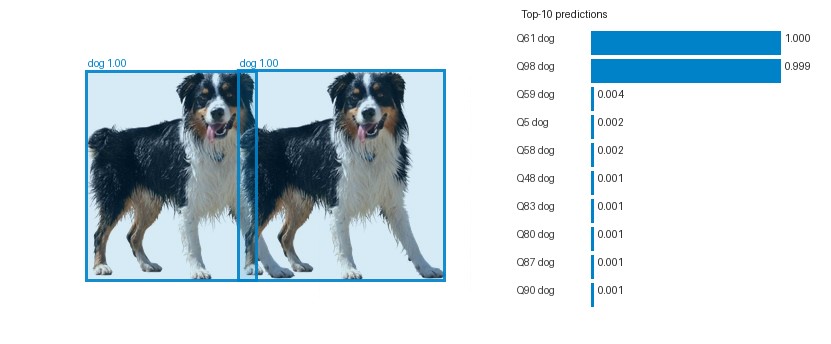}
    }

    \subfloat[\texttt{Two Dogs (Medium Overlap)}  ]{
        \includegraphics[width=0.4\textwidth]{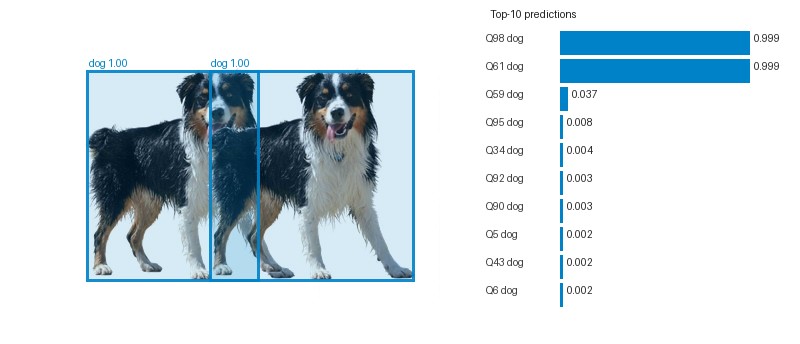}
    }
    \hspace{2em}
    \subfloat[\texttt{Two Dogs (Medium Overlap)}  ]{
        \includegraphics[width=0.4\textwidth]{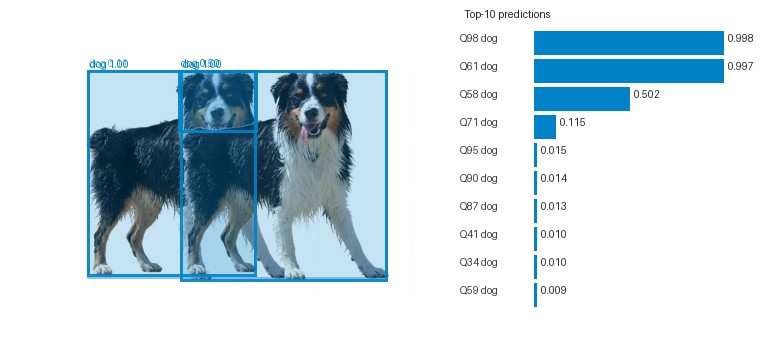}
    }
    
    \subfloat[\texttt{Two Dogs (Large Overlap)}  ]{
        \includegraphics[width=0.4\textwidth]{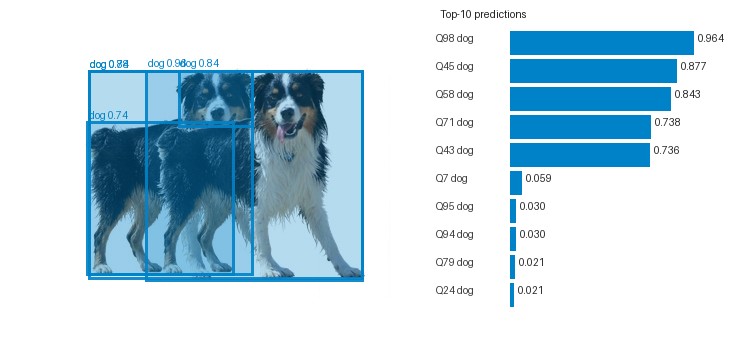}
    }
    \hspace{2em}
    \subfloat[\texttt{Two Dogs (Large Overlap)}  ]{
        \includegraphics[width=0.4\textwidth]{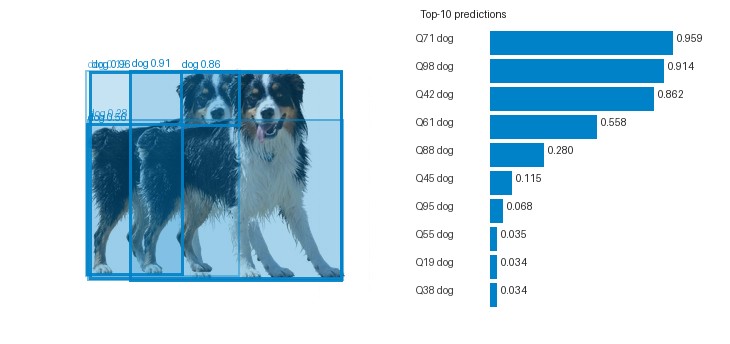}
    }
    
    \caption{\revv{Inference results of DETR on synthetic images containing two dogs with varying levels of overlap.}} \label{fig:detr_synthetic_dog}
\end{figure*}

\begin{figure*}[t]
    \centering
    \subfloat[\texttt{Two Dogs (No Overlap)}  ]{
        \includegraphics[width=0.4\textwidth]{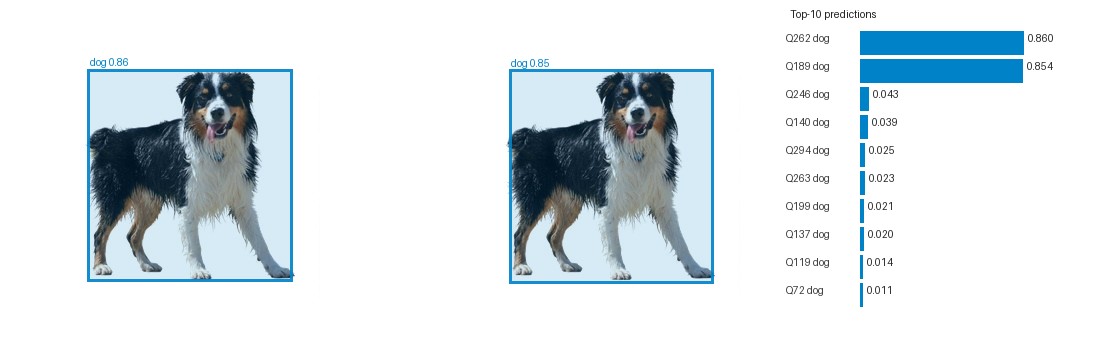}
    }
    \hspace{2em}
    \subfloat[\texttt{Two Dogs (No Overlap)}  ]{
        \includegraphics[width=0.4\textwidth]{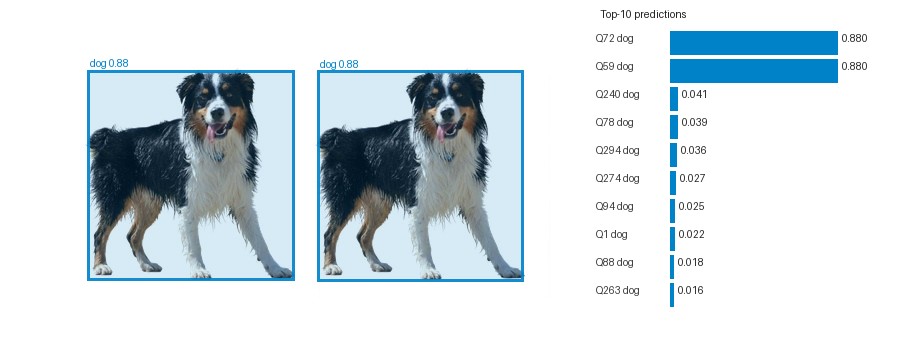}
    }
    
    \subfloat[\texttt{Two Dogs (Small Overlap)}  ]{
        \includegraphics[width=0.4\textwidth]{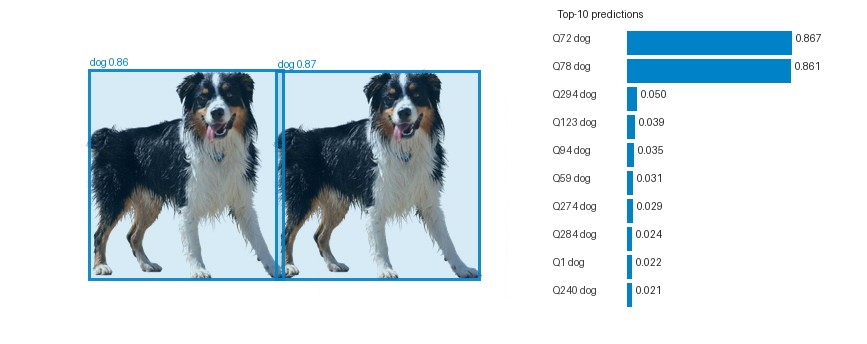}
    }
    \hspace{2em}
    \subfloat[\texttt{Two Dogs (Small Overlap)}  ]{
        \includegraphics[width=0.4\textwidth]{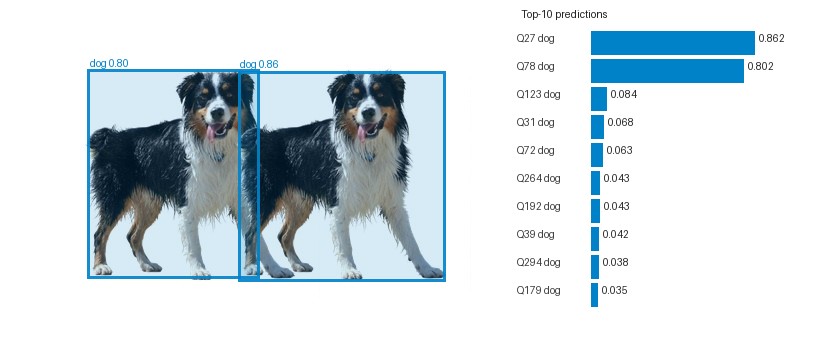}
    }

    \subfloat[\texttt{Two Dogs (Medium Overlap)}  ]{
        \includegraphics[width=0.4\textwidth]{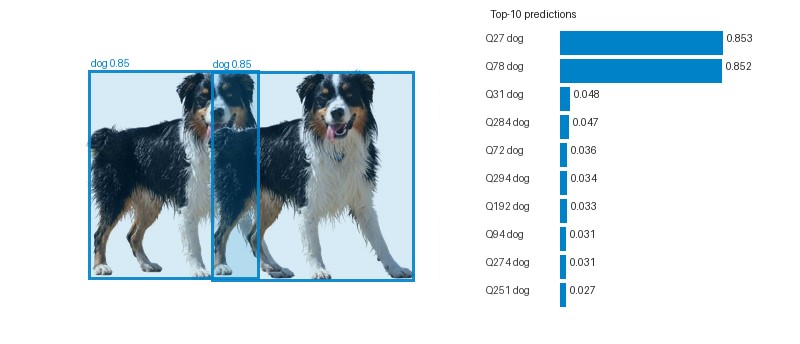}
    }
    \hspace{2em}
    \subfloat[\texttt{Two Dogs (Medium Overlap)}  ]{
        \includegraphics[width=0.4\textwidth]{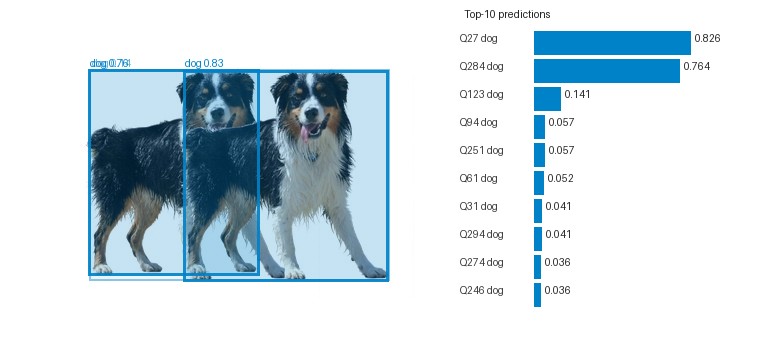}
    }
    
    \subfloat[\texttt{Two Dogs (Large Overlap)}  ]{
        \includegraphics[width=0.4\textwidth]{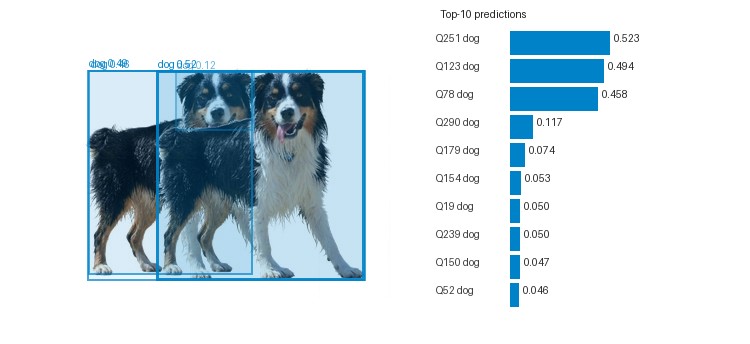}
    }
    \hspace{2em}
    \subfloat[\texttt{Two Dogs (Large Overlap)}  ]{
        \includegraphics[width=0.4\textwidth]{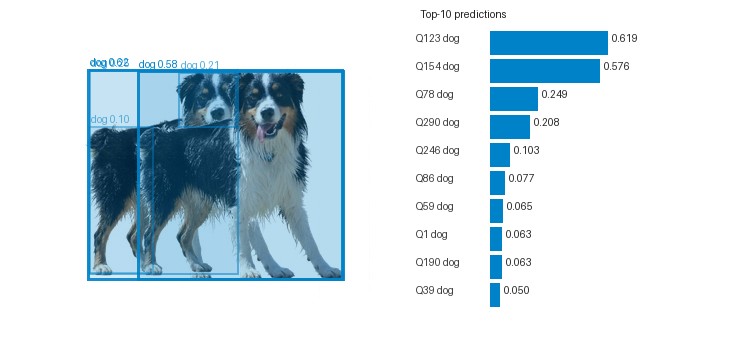}
    }
    
    \caption{\revv{Inference results of D-DETR on synthetic images containing two dogs with varying levels of overlap.}} \label{fig:ddetr_synthetic_dog}
\end{figure*}

\begin{figure*}[t]
    \centering
    \subfloat[\texttt{Person\&Dog (No Overlap)}  ]{
        \includegraphics[width=0.4\textwidth]{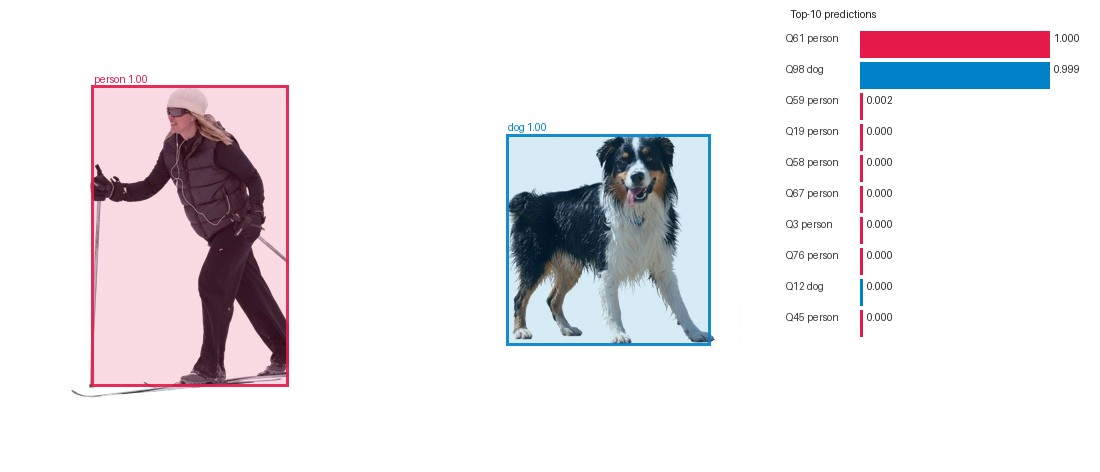}
    }
    \hspace{2em}
    \subfloat[\texttt{Person\&Dog (No Overlap)}  ]{
        \includegraphics[width=0.4\textwidth]{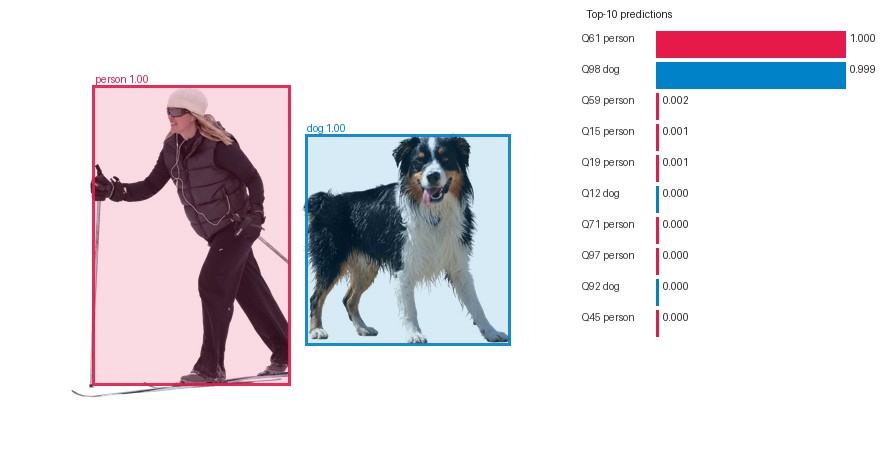}
    }
    
    \subfloat[\texttt{Person\&Dog (Small Overlap)}  ]{
        \includegraphics[width=0.4\textwidth]{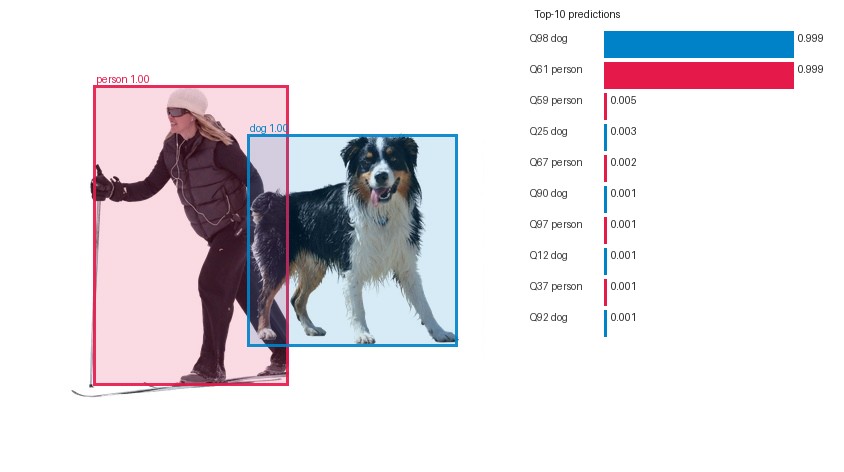}
    }
    \hspace{2em}
    \subfloat[\texttt{Person\&Dog (Small Overlap)}  ]{
        \includegraphics[width=0.4\textwidth]{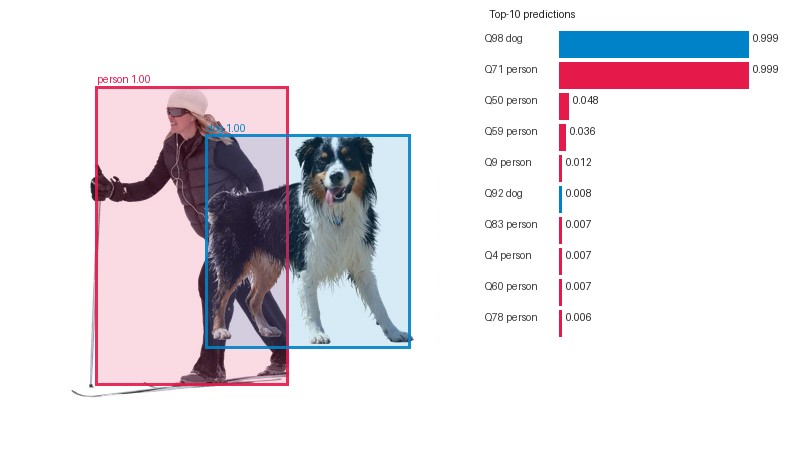}
    }

    \subfloat[\texttt{Person\&Dog (Medium Overlap)}  ]{
        \includegraphics[width=0.4\textwidth]{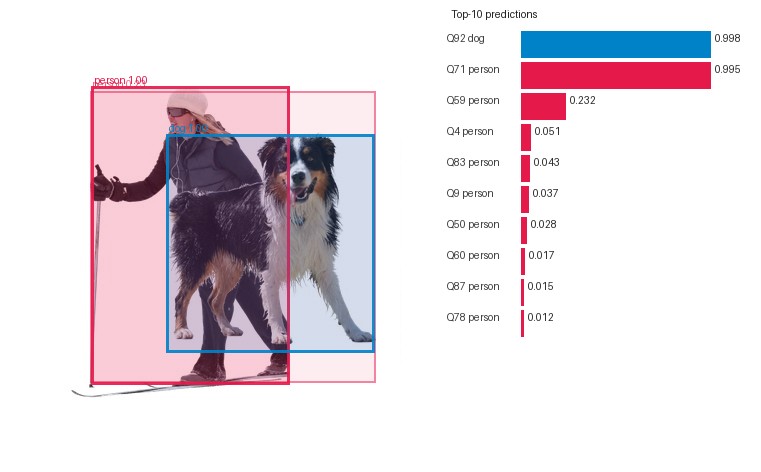}
    }
    \hspace{2em}
    \subfloat[\texttt{Person\&Dog (Medium Overlap)}  ]{
        \includegraphics[width=0.4\textwidth]{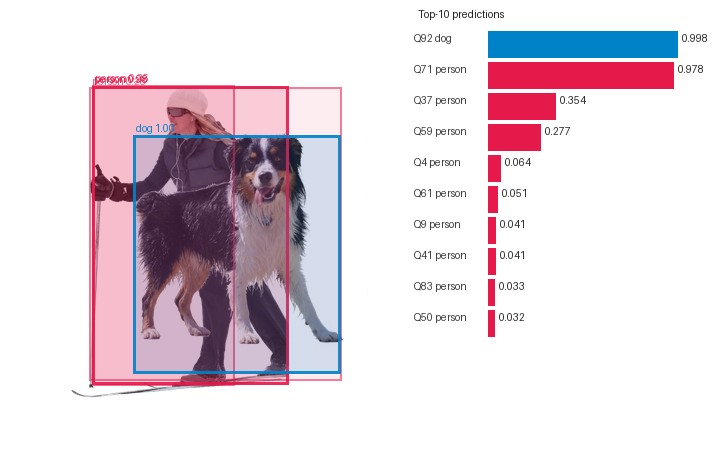}
    }
    
    \subfloat[\texttt{Person\&Dog (Large Overlap)}  ]{
        \includegraphics[width=0.4\textwidth]{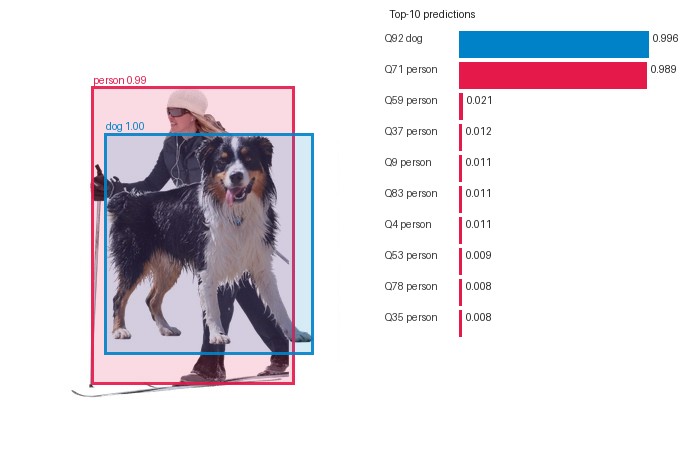}
    }
    \hspace{2em}
    \subfloat[\texttt{Person\&Dog (Large Overlap)}  ]{
        \includegraphics[width=0.4\textwidth]{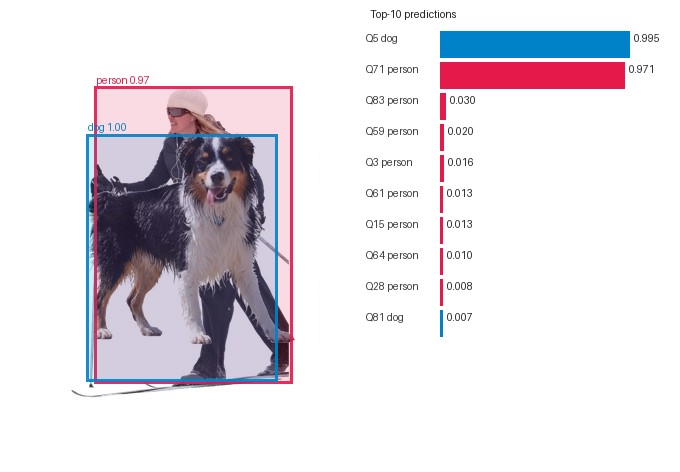}
    }
    
    \caption{\revv{Inference results of DETR on synthetic images containing a person and a dog with varying levels of overlap.}} \label{fig:detr_synthetic_person_dog}
\end{figure*}

\begin{figure*}[t]
    \centering
    \subfloat[\texttt{Person\&Dog (No Overlap)}  ]{
        \includegraphics[width=0.4\textwidth]{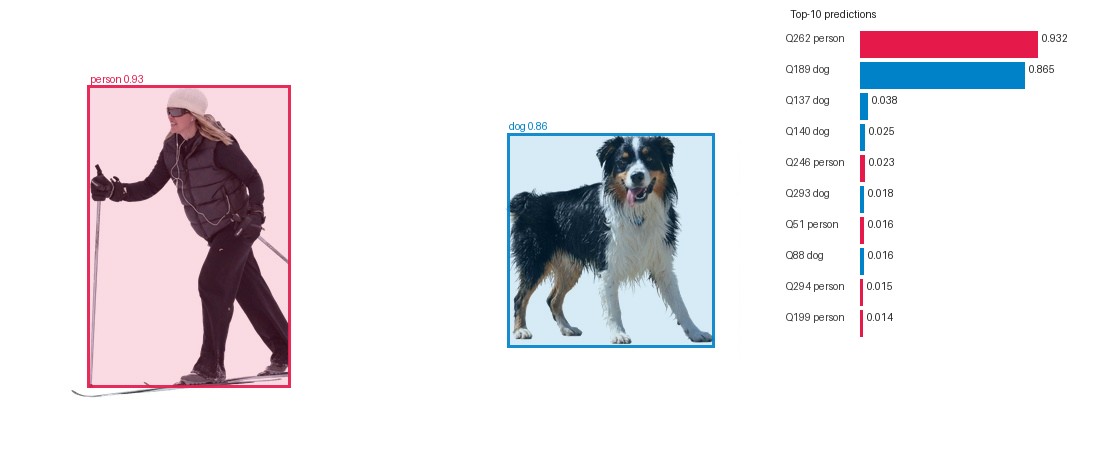}
    }
    \hspace{2em}
    \subfloat[\texttt{Person\&Dog (No Overlap)}  ]{
        \includegraphics[width=0.4\textwidth]{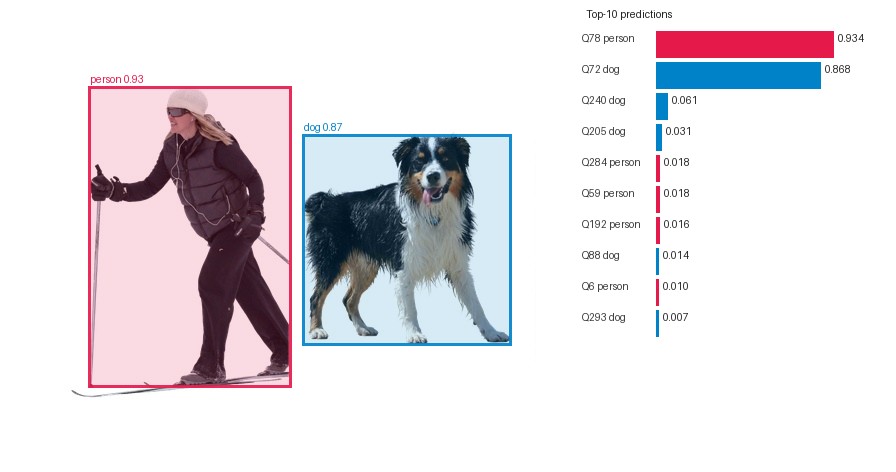}
    }
    
    \subfloat[\texttt{Person\&Dog (Small Overlap)}  ]{
        \includegraphics[width=0.4\textwidth]{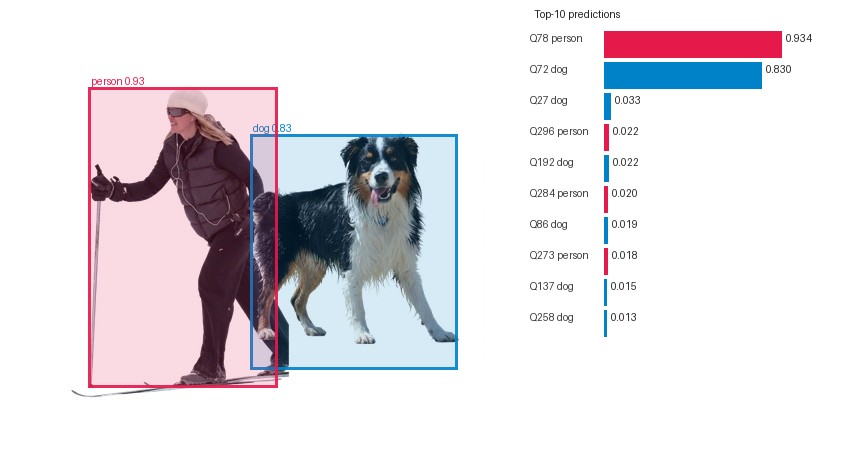}
    }
    \hspace{2em}
    \subfloat[\texttt{Person\&Dog (Small Overlap)}  ]{
        \includegraphics[width=0.4\textwidth]{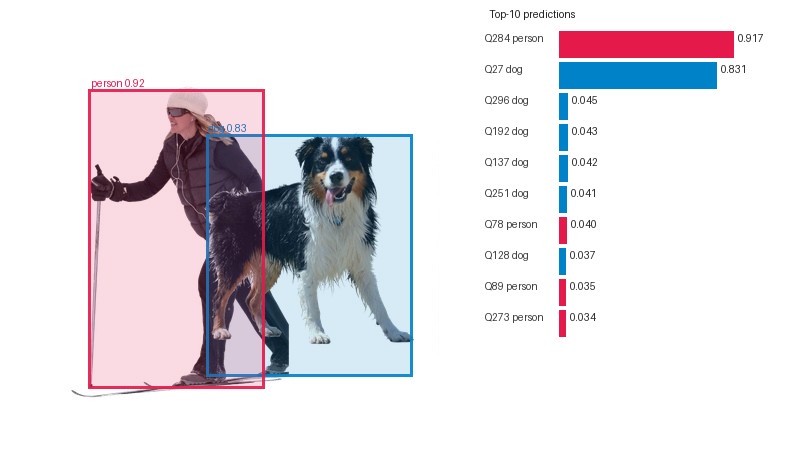}
    }

    \subfloat[\texttt{Person\&Dog (Medium Overlap)}  ]{
        \includegraphics[width=0.4\textwidth]{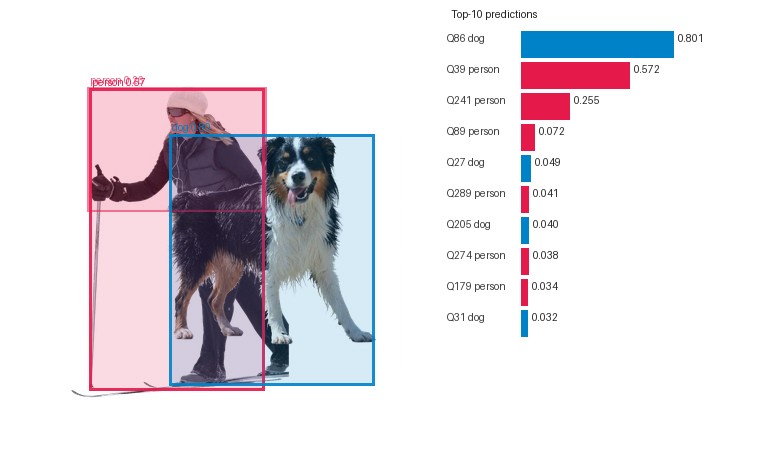}
    }
    \hspace{2em}
    \subfloat[\texttt{Person\&Dog (Medium Overlap)}  ]{
        \includegraphics[width=0.4\textwidth]{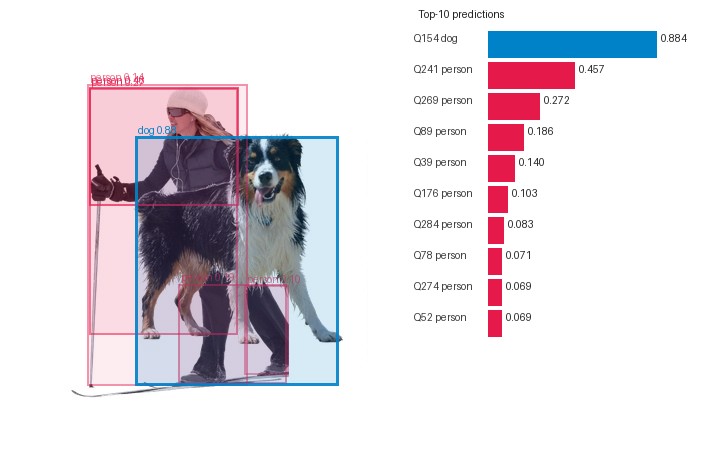}
    }
    
    \subfloat[\texttt{Person\&Dog (Large Overlap)}  ]{
        \includegraphics[width=0.4\textwidth]{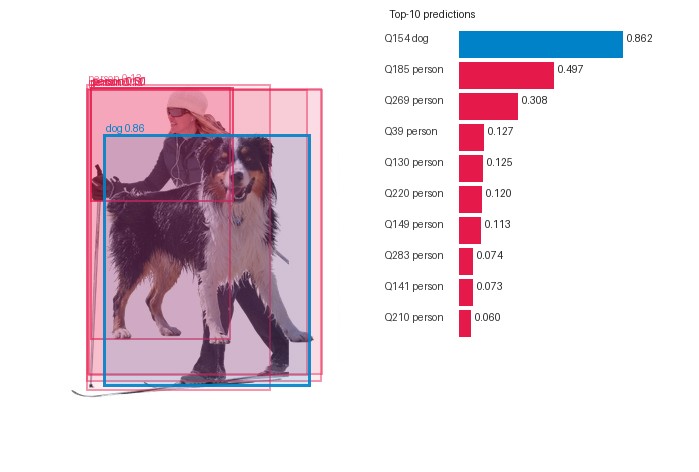}
    }
    \hspace{2em}
    \subfloat[\texttt{Person\&Dog (Large Overlap)}  ]{
        \includegraphics[width=0.4\textwidth]{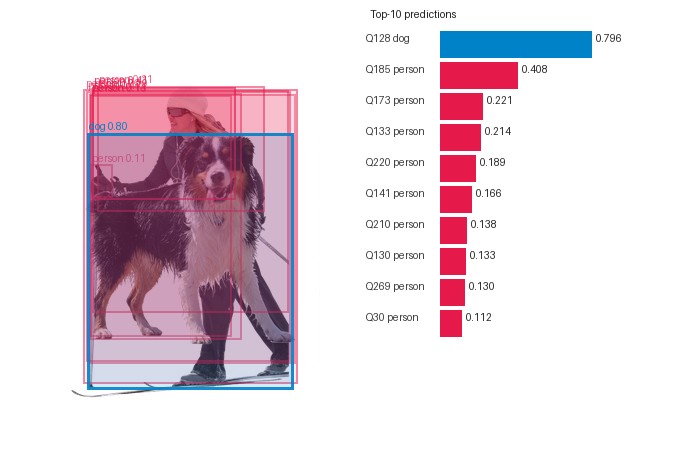}
    }
    
    \caption{\revv{Inference results of D-DETR on synthetic images containing a person and a dog with varying levels of overlap.}} \label{fig:ddetr_synthetic_person_dog}
\end{figure*}

\begin{revision2}
\subsection{Extension to Multiple Object Scenarios} \label{app:multi}
\label{proof:multi}

We now consider $1 \le K \le M$ foreground targets: $y_\yidx=(c_\yidx,b_\yidx)$ for $\yidx=1,\dots,K$, and let $y_\yidx=\emptyset$ for $\yidx>K$.
In a general stochastic setting, the Hungarian assignment $\optmatch(\yidx)$ is a random variable that depends dynamically on the specific realization of the ground-truth targets. 
To cleanly decouple the expected loss into independent terms for specific queries, we assume a \emph{non-interference} regime where this assignment becomes deterministic.

\begin{assumption}[\revvv{Non-Interference Regime}] \label{assum:non_interference}
There exist distinct optimal positive query indices $\qidx_1,\dots,\qidx_K \in \{1,\dots,M\}$ such that $\optmatch(\yidx)=\qidx_\yidx$ for $\yidx=1,\dots,K$ with probability 1.
Denoting these optimal positive queries as $Q_+=\{\qidx_1,\dots,\qidx_K\}$, the expected total loss decomposes as:
\begin{equation*}
\mathbb{E}[\Ltotal] = \sum_{\yidx=1}^{K}\mathbb{E}\!\left[\mathcal{L}(\hat y_{\qidx_\yidx},y_\yidx)\right] + \sum_{\qidx\notin Q_+}\mathcal{L}(\hat y_\qidx,\emptyset).
\end{equation*}
\end{assumption}

\begin{cor} \label{prop:bayes_multi}
Under the non-interference regime \revvv{(Assumption~\ref{assum:non_interference})},
$$ \mathbb{E}[\Ltotal] \ge \sum_{\yidx=1}^{K} \left( \entropy(p_{\mathrm{gt},\yidx}) + \min_{\hat b}\mathbb{E}[\Lbox(\hat b,b_\yidx)] \right) $$
where $\entropy(p_{\text{gt},\yidx}) = -\sum_{c \in \cset} p_{\text{gt},\yidx}(c) \log(p_{\text{gt},\yidx}(c))$.
\end{cor}
\begin{proof}
Since $\mathcal{L}(\hat y_\qidx,\emptyset)=-w_{\emptyset}\log \hat p_\qidx(\emptyset)\ge 0$,
$$
\mathbb{E}[\Ltotal] \ge \sum_{\yidx=1}^{K}\mathbb{E}\!\left[\mathcal{L}(\hat y_{\qidx_\yidx},y_\yidx)\right].
$$
For each $\yidx\le K$, by the same argument as in the proof of Proposition~\ref{prop:bayes},
$$
\mathbb{E}\!\left[\mathcal{L}(\hat y_{\qidx_\yidx},y_\yidx)\right] \ge \entropy(p_{\mathrm{gt},\yidx}) + \min_{\hat b}\mathbb{E}[\Lbox(\hat b,b_\yidx)].
$$
Summing over \(\yidx=1,\dots,K\) proves the claim.
\end{proof}

\begin{cor}[Optimal Prediction Set] \label{prop:specialist_multi}
\revv{Let $\{(\hat{p}^*_\qidx, \hat{b}^*_\qidx)\}_{\qidx=1}^{M}$ denote a globally optimal prediction set under the non-interference regime \revvv{(Assumption~\ref{assum:non_interference})}, i.e., that minimizes $\mathbb{E}\left[ \Ltotal \right]$. Then we have}
\begin{enumerate}
    \item for each optimal positive prediction ($\qidx = \qidx_\yidx \in Q_+$): $\hat{p}^*_\qidx = p_{\text{gt},\yidx}$ (i.e., $\hat{p}^*_{\qidx}(c) = p_{\text{gt},\yidx}(c)$ $\forall c \in \cset$ and $\hat{p}^*_{\qidx}(\emptyset)=0$) and the box $\hat{b}^*_\qidx$ minimizes $\mathbb{E}[\Lbox(\hat{b}_\qidx, b_\yidx)]$, and
    \item for any optimal negative prediction ($\qidx \notin Q_+$): $\hat{p}^*_{\qidx}(c)=0$ $\forall c \in \cset$ and $\hat{p}^*_{\qidx}(\emptyset)=1$.
\end{enumerate}
\end{cor}
\begin{proof}
Because the predictions $\{\hat{y}_\qidx\}_{\qidx=1}^M$ are parameterized independently, minimizing $\mathbb{E}[\Ltotal]$ is equivalent to minimizing each term in the decomposed sum individually. 
For each $\qidx_\yidx \in Q_+$, applying Proposition~\ref{prop:specialist} yields that $\mathbb{E}\!\left[\mathcal{L}(\hat y_{\qidx_\yidx},y_\yidx)\right]$ is minimized when $\hat{p}^*_{\qidx_\yidx} = p_{\mathrm{gt},\yidx}$ and $\hat{b}^*_{\qidx_\yidx}$ minimizes $\mathbb{E}[\Lbox(\hat{b}_\qidx, b_\yidx)]$.
For each $\qidx \notin Q_+$, the term $\mathcal{L}(\hat y_\qidx,\emptyset) = -w_{\emptyset}\log \hat p_\qidx(\emptyset)$ is uniquely minimized at $0$ when $\hat p_\qidx(\emptyset)=1$, which implies $\hat p_\qidx(c)=0$ for all $c \in \cset$.
\end{proof}

As noted in the main body, the situation becomes more complex when multiple objects compete strongly during matching---for example, due to spatial overlap with similar appearance---and encourages high uncertainty in matching. In other words, the matching uncertainty for each query increases, and the optimal behavior is expected to shift from specialization toward hedging. We provide qualitative evidence for this conjecture.

First, we perform inference on several \texttt{COCO} images containing two cats, two giraffes, and two people, respectively, using DETR and D-DETR. As shown in Figures~\ref{fig:detr_real} and \ref{fig:ddetr_real}, we observe that the specialist pattern holds when instances are visually distinguishable (even if spatially overlapping). On the other hand, when instances are visually entangled, the model hedges by producing multiple mid-confidence predictions.

To confirm this more systematically, we synthetically generated multiple images containing either two people (Figures~\ref{fig:detr_synthetic_person} and \ref{fig:ddetr_synthetic_person}), two dogs (Figures~\ref{fig:detr_synthetic_dog} and \ref{fig:ddetr_synthetic_dog}), or a person and a dog (Figures~\ref{fig:detr_synthetic_person_dog} and \ref{fig:ddetr_synthetic_person_dog}), with varying levels of overlap. This provides further evidence for our conjecture: when the objects do not interfere with each other and are visually distinguishable, allowing the model to make confident predictions, they follow the specialist regime. In contrast, as the overlap increases, the model tends to generate multiple predictions to hedge against the risk of incorrect matching. A more formal and deeper analysis remains an interesting direction for future work.

\end{revision2}

\section{Omitted Experimental Results} \label{app:results}

\begin{figure*}
    \centering
    \subfloat[]{\includegraphics[width=0.25\linewidth]{figures/ecdf_plot_UP-DETR.pdf}}
    \hspace{2em}
    \subfloat[]{\includegraphics[width=0.25\linewidth]{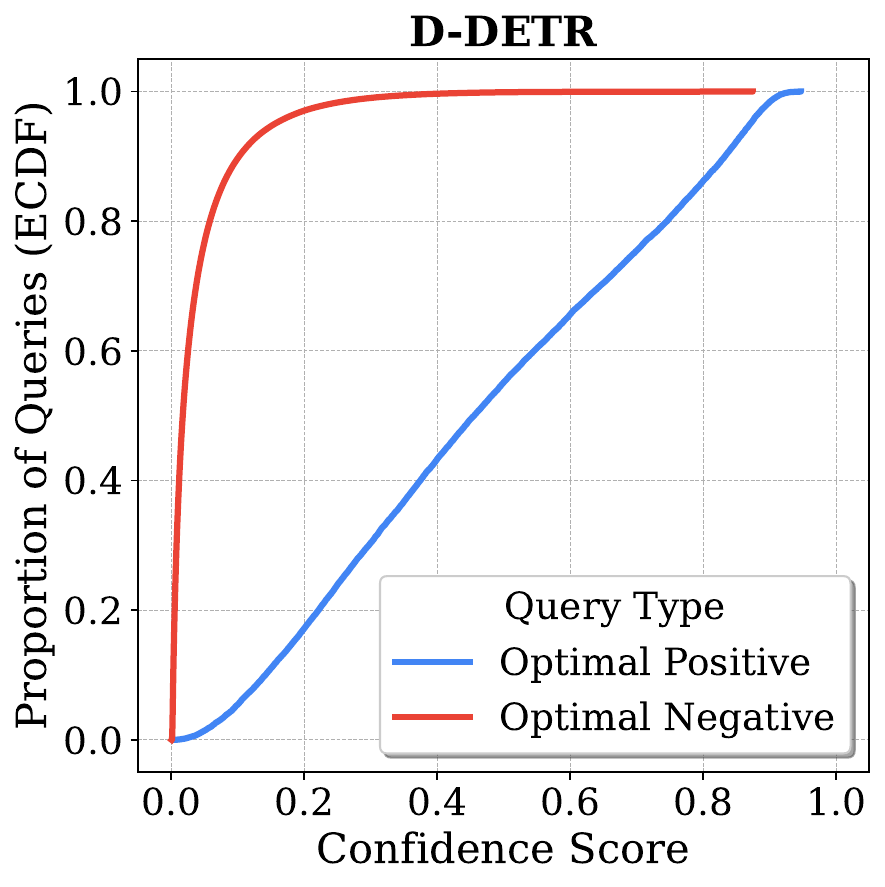}}
    \hspace{2em}
    \subfloat[]{\includegraphics[width=0.25\linewidth]{figures/ecdf_plot_Cal-DETR.pdf}}
    
    \subfloat[]{\includegraphics[width=0.25\linewidth]{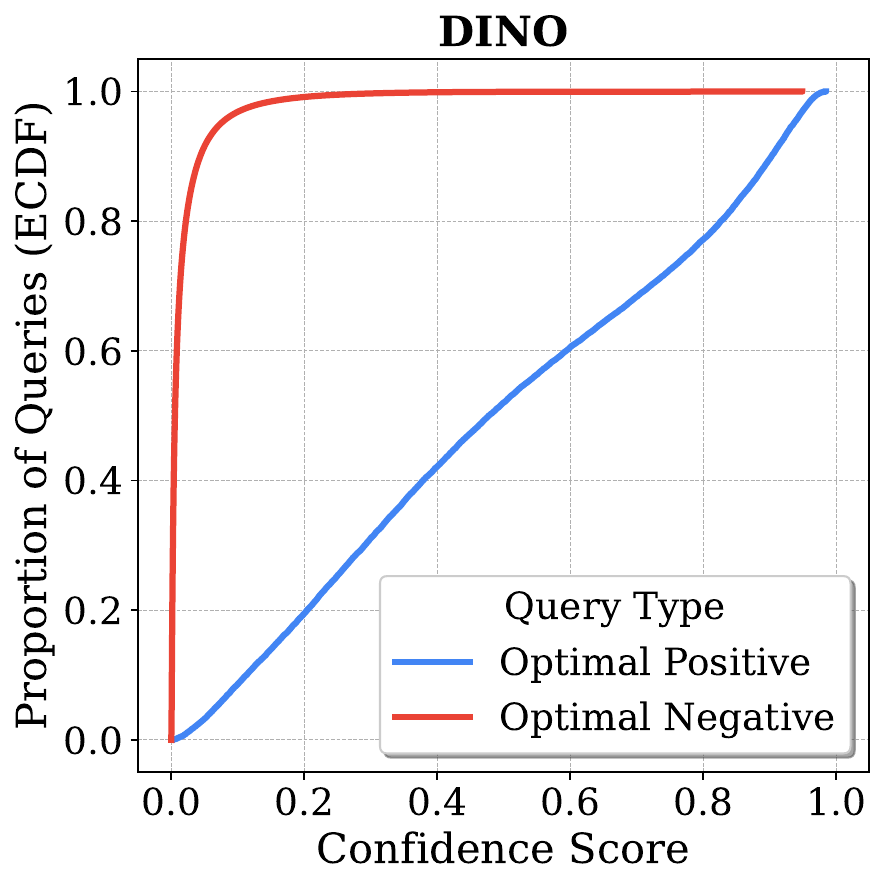}}
    \hspace{2em}
    \subfloat[]{\includegraphics[width=0.25\linewidth]{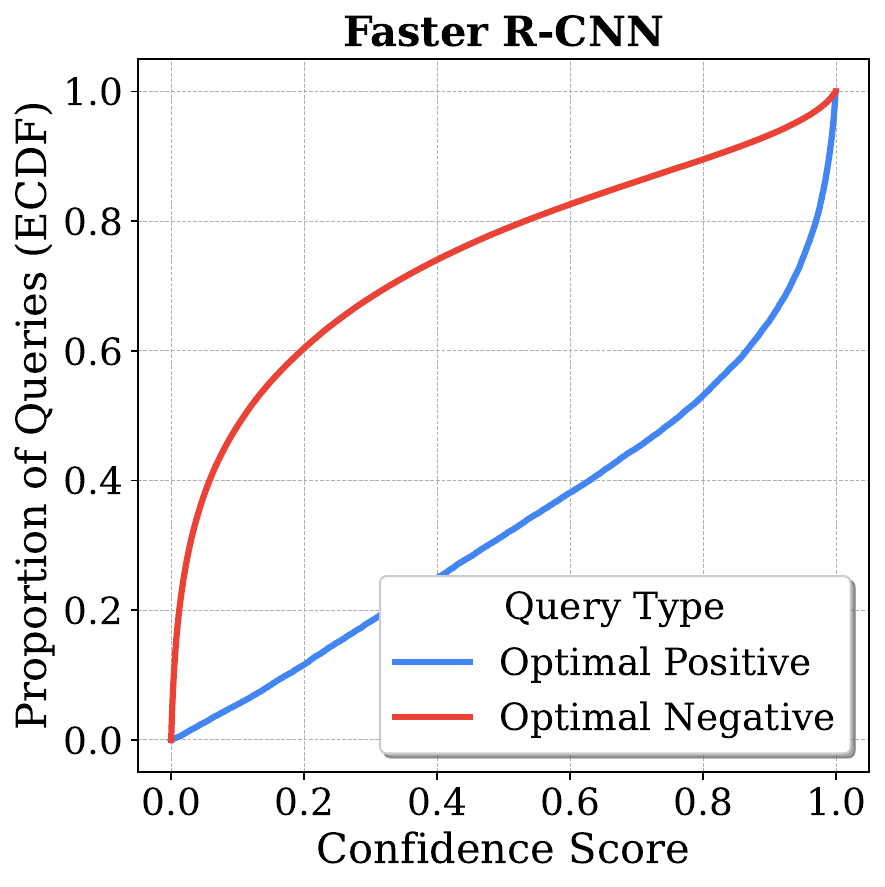}}
    \hspace{2em}
    \subfloat[]{\includegraphics[width=0.25\linewidth]{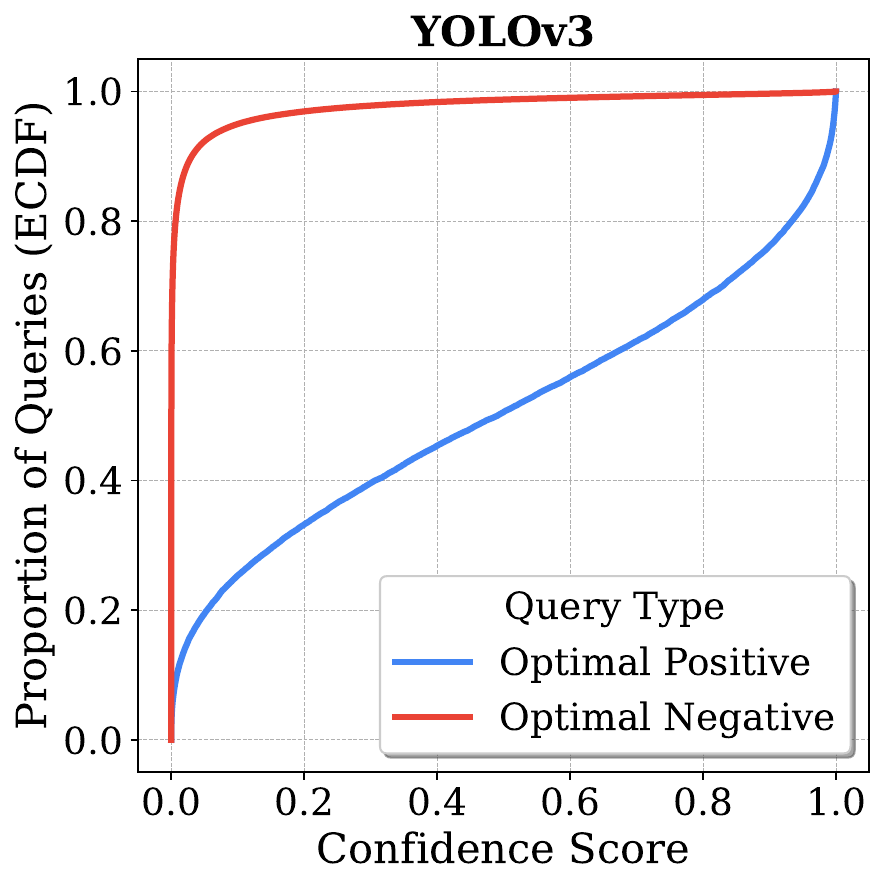}}
    
    \subfloat[]{\includegraphics[width=0.25\linewidth]{figures/ecdf_bbox_plot_UP-DETR.pdf}}
    \hspace{2em}
    \subfloat[]{\includegraphics[width=0.25\linewidth]{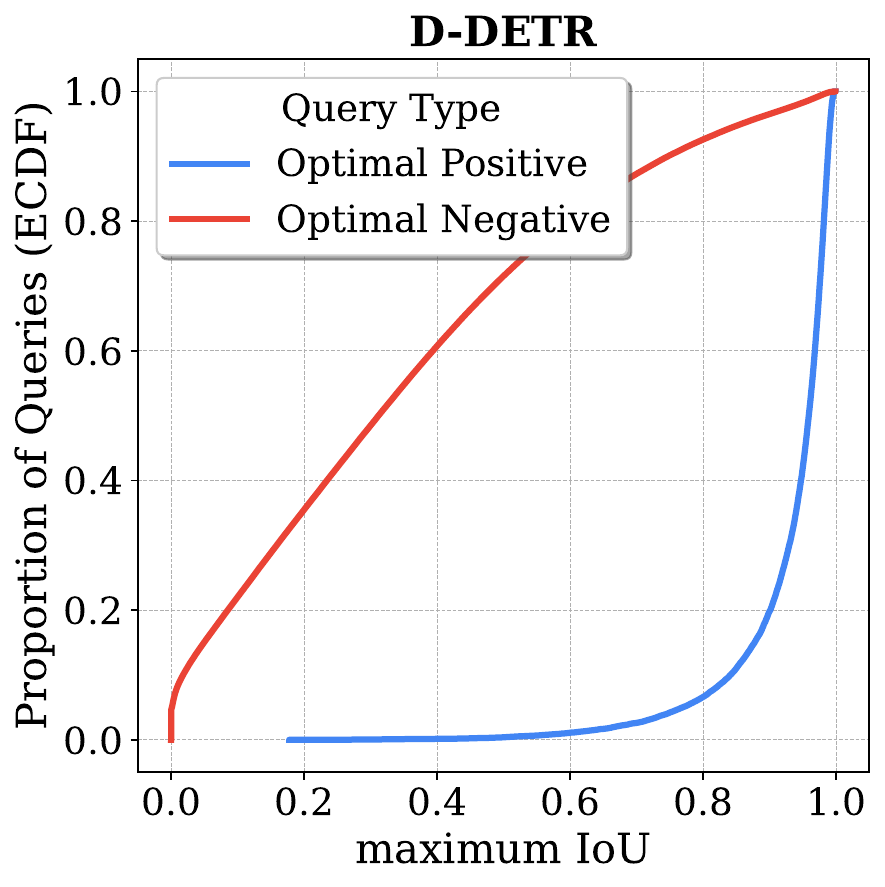}}
    \hspace{2em}
    \subfloat[]{\includegraphics[width=0.25\linewidth]{figures/ecdf_bbox_plot_Cal-DETR.pdf}}
    
    \subfloat[]{\includegraphics[width=0.25\linewidth]{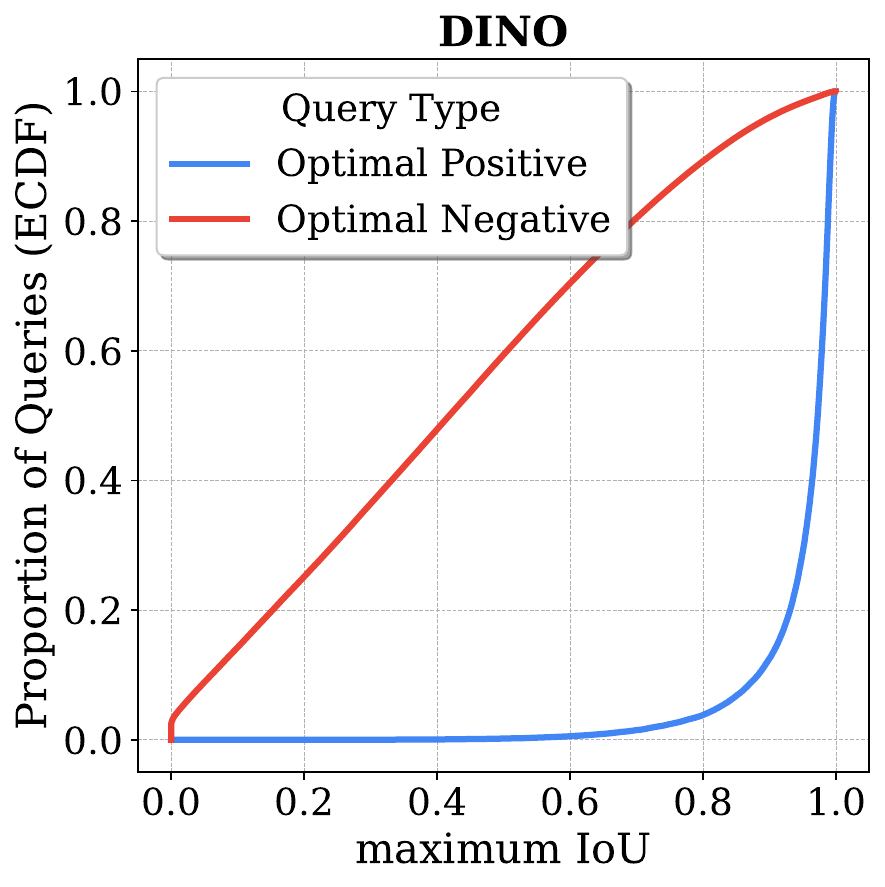}}
    \hspace{2em}
    \subfloat[]{\includegraphics[width=0.25\linewidth]{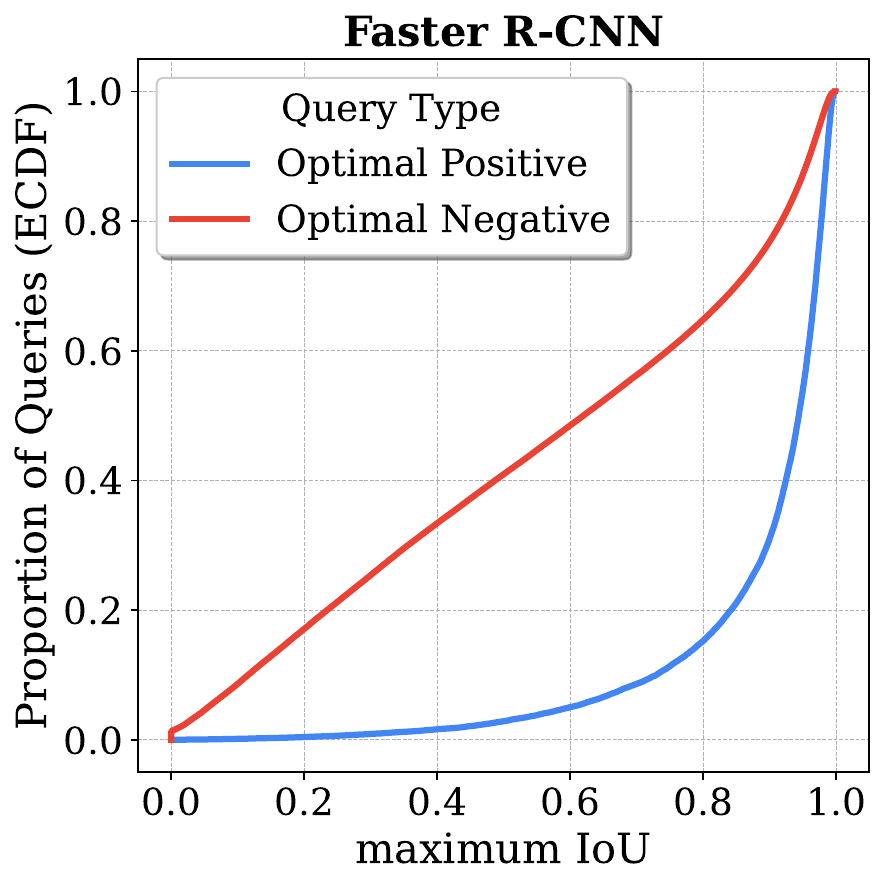}}
    \hspace{2em}
    \subfloat[]{\includegraphics[width=0.25\linewidth]{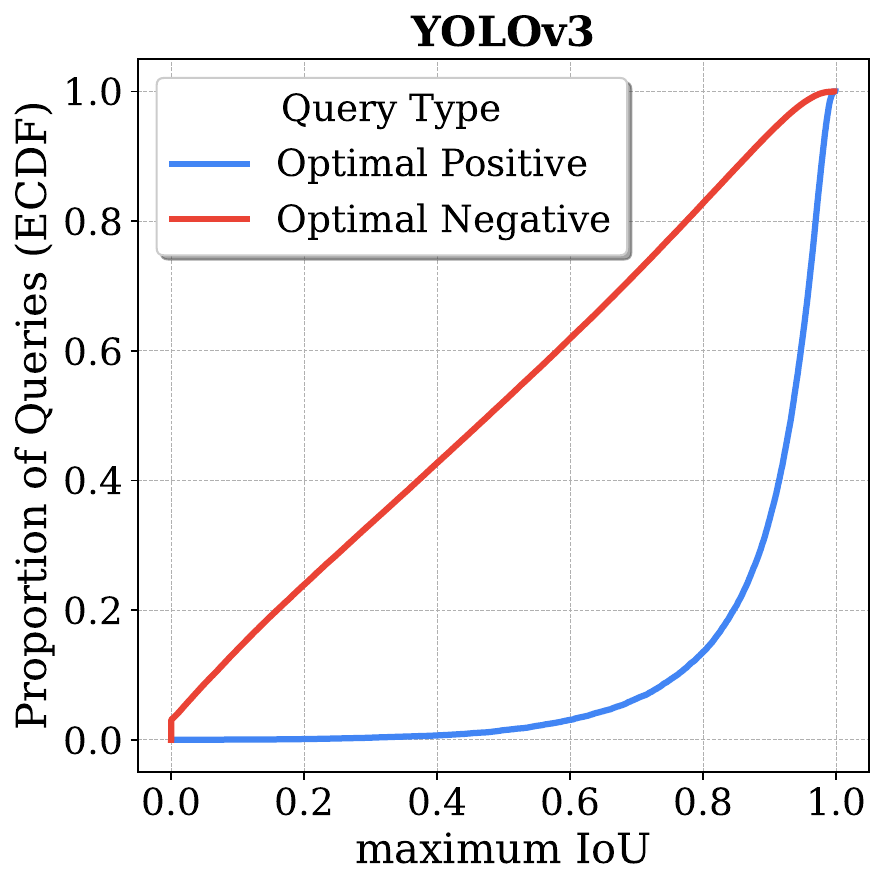}}
    
    \caption{
    \rev{Empirical cumulative distribution function (ECDF) plots for maximum confidence score and maximum IoU, comparing optimal positive predictions (blue) and optimal negative predictions (red) on the \texttt{COCO} dataset.}
    } \label{fig:ecdf_appendix}
\end{figure*}

\begin{figure*}[t!]
    \centering
    \subfloat[UP-DETR (\texttt{COCO})]{%
        \includegraphics[width=0.23\textwidth]{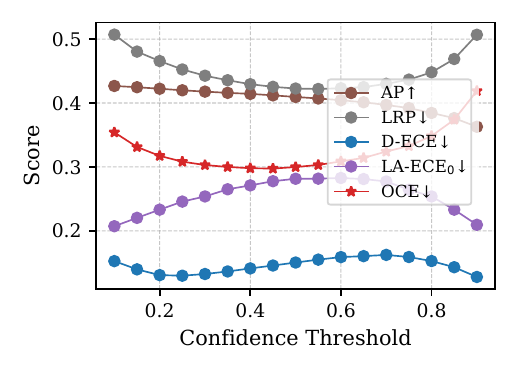}
    }
    \subfloat[D-DETR (\texttt{COCO})]{%
        \includegraphics[width=0.23\textwidth]{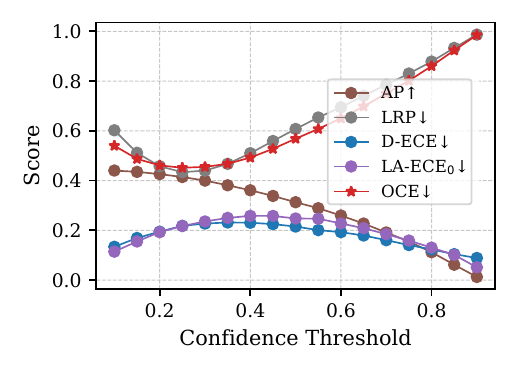}
    }
    \subfloat[Cal-DETR (\texttt{COCO})]{%
        \includegraphics[width=0.23\textwidth]{figures/fig1_cocoval_caldetr.pdf}
    }
    \subfloat[DINO (\texttt{COCO})]{%
        \includegraphics[width=0.23\textwidth]{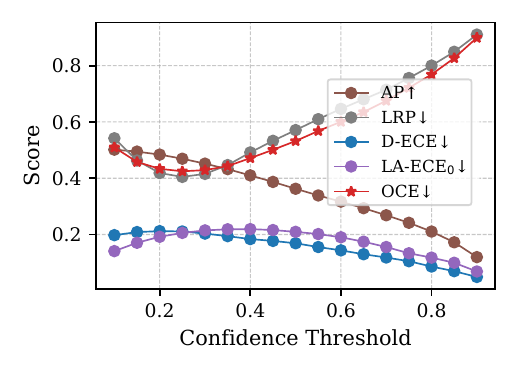}
    }
    
    \subfloat[UP-DETR (\texttt{City})]{%
        \includegraphics[width=0.23\textwidth]{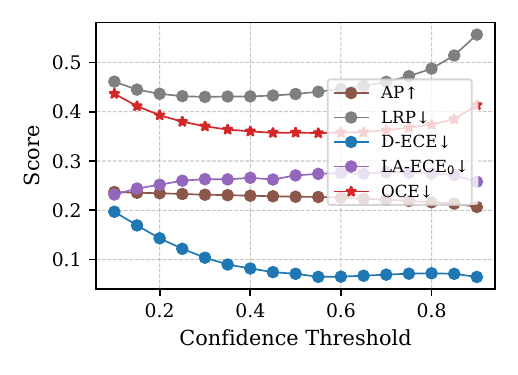}
    }
    \subfloat[D-DETR (\texttt{City})]{%
        \includegraphics[width=0.23\textwidth]{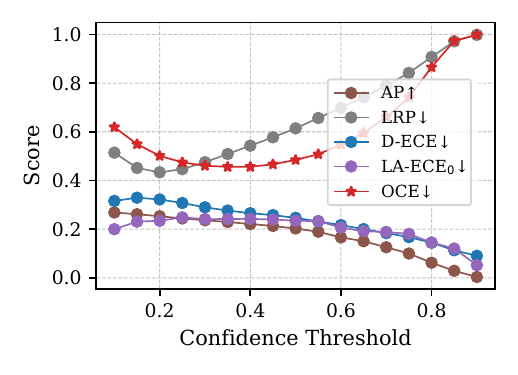}
    }
    \subfloat[Cal-DETR (\texttt{City})]{%
        \includegraphics[width=0.23\textwidth]{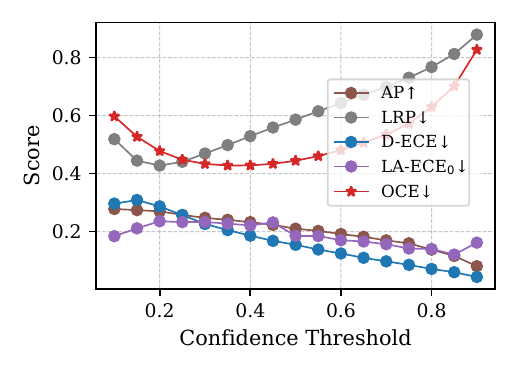}
    }
    \subfloat[DINO (\texttt{City})]{%
        \includegraphics[width=0.23\textwidth]{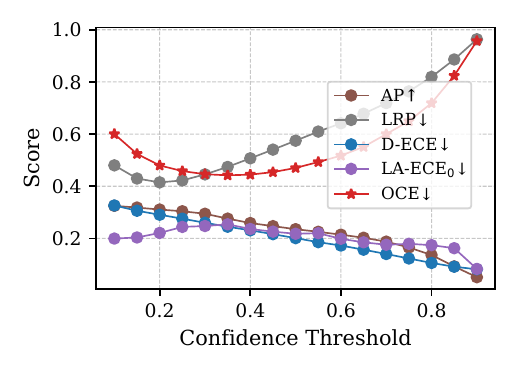}
    }
    
    \subfloat[UP-DETR (\texttt{Foggy})]{%
        \includegraphics[width=0.23\textwidth]{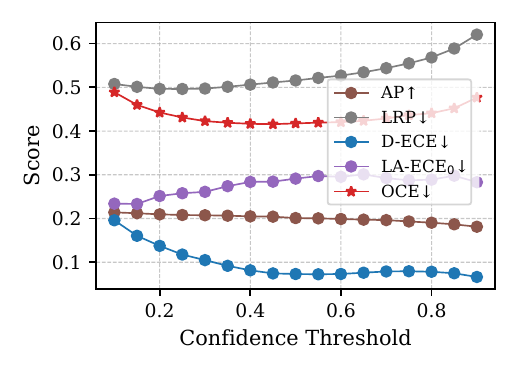}
    }
    \subfloat[D-DETR (\texttt{Foggy})]{%
        \includegraphics[width=0.23\textwidth]{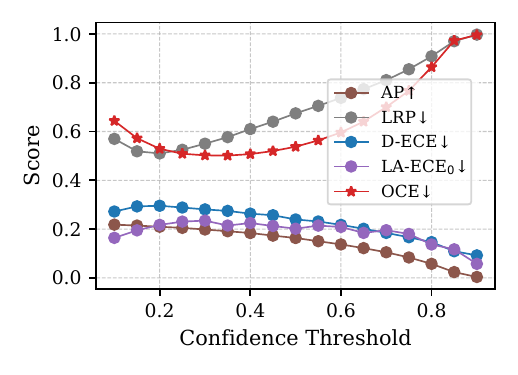}
    }
    \subfloat[Cal-DETR (\texttt{Foggy})]{%
        \includegraphics[width=0.23\textwidth]{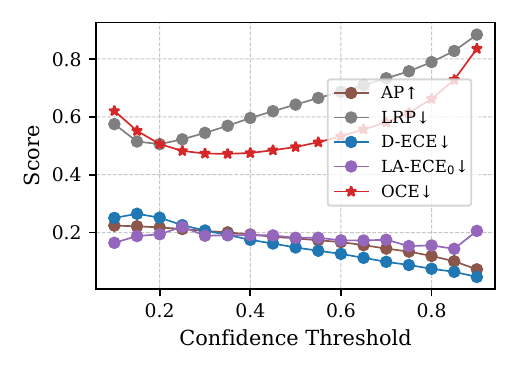}
    }
    \subfloat[DINO (\texttt{Foggy})]{%
        \includegraphics[width=0.23\textwidth]{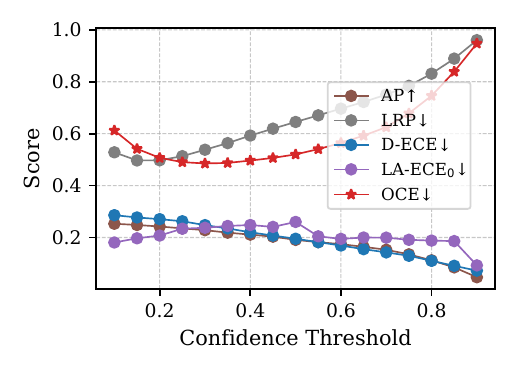}
    }
    \caption{Impact of confidence threshold selection on various performance metrics in UP-DETR, Deformable-DETR, Cal-DETR, and DINO on \texttt{COCO}, \texttt{Cityscapes}, and \texttt{Foggy Cityscapes}. Higher scores are preferable ($\uparrow$), while lower scores are preferable ($\downarrow$).}
    \label{fig:thr_appendix}
\end{figure*}

This section provides the omitted experimental results.
Figure~\ref{fig:thr_appendix} extends Figure~\ref{fig:thr_metrics} across different DETR variants and datasets.
Figure~\ref{fig:ecdf_appendix} extends Figure~\ref{fig:ecdf} across different DETR variants and datasets.

\begin{table*}[t!]
\centering\small
\caption{\revv{The Pearson correlation coefficients among OCE metrics using different aggregation methods across six models---UP-DETR, D-DETR, Cal-DETR, DINO, Faster R-CNN, and YOLOv3---and three data distributions.
Specifically, we compare the original averaging approach with the Max-IoU and IoU-weighted averaging methods described in Section~\ref{sec:oce_abl}.
The consistently strong correlations confirm that OCE is highly robust to the choice of aggregation method.}}
\resizebox{1.0\linewidth}{!}{
\begin{tabular}{l|ccc|ccc|ccc|c}
\toprule
\multirow{2}{*}{\textbf{Metrics}} & \multicolumn{3}{c}{\texttt{\textbf{COCO (in-distribution)}}} & \multicolumn{3}{c}{\textbf{\texttt{Cityscapes  (near OOD)}}} & \multicolumn{3}{c|}{\textbf{\texttt{Foggy Cityscapes (OOD)}}} & \multirow{2}{*}{\textbf{\makecell[c]{Average \\ $\pm$ Std.}}} \\
\cmidrule(lr){2-4} \cmidrule(lr){5-7} \cmidrule(lr){8-10}
 & \textbf{Averaging} & \textbf{Max-IoU} & \textbf{IoU-Weighted} & \textbf{Averaging} & \textbf{Max-IoU} & \textbf{IoU-Weighted} & \textbf{Averaging} & \textbf{Max-IoU} & \textbf{IoU-Weighted} \\
\midrule
\midrule
Averaging & 1.000 & 0.996 & 1.000 & 1.000 & 0.899 & 0.999 & 1.000 & 0.910 & 0.999 & \textcolor{black}{0.978 $\pm$ 0.042} \\
Max-IoU & 0.996 & 1.000 & 0.996 & 0.899 & 1.000 & 0.918 & 0.910 & 1.000 & 0.926 & \textcolor{black}{0.961 $\pm$ 0.045} \\
IoU-Weighted & 1.000 & 0.996 & 1.000 & 0.999 & 0.918 & 1.000 & 0.999 & 0.926 & 1.000 & \textcolor{black}{0.982 $\pm$ 0.034} \\
\bottomrule
\end{tabular}
}
\label{tab:oce_abl}
\end{table*}

\begin{figure*}[t]
    \centering
    \subfloat[\texttt{COCO}]{
        \includegraphics[width=0.32\textwidth]{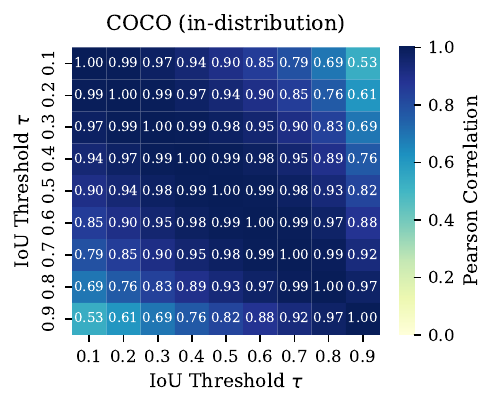}
    }
    \subfloat[\texttt{Cityscapes}]{
        \includegraphics[width=0.32\textwidth]{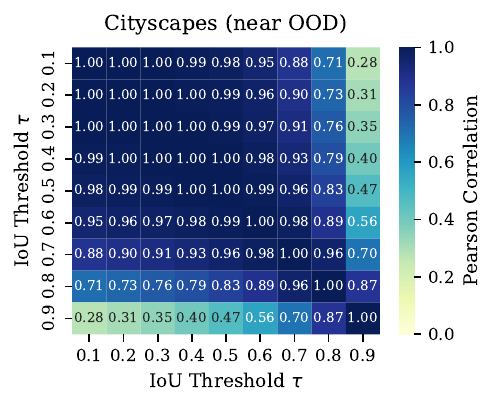}
    }
    \subfloat[\texttt{Foggy Cityscapes}]{
        \includegraphics[width=0.32\textwidth]{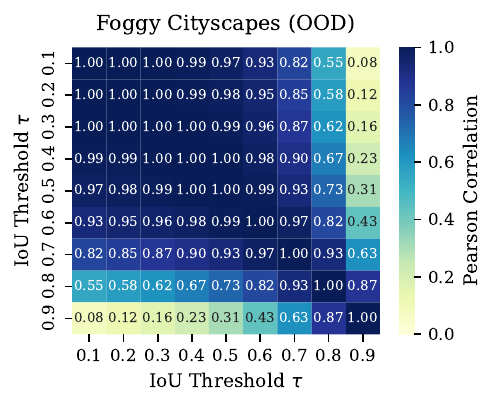}
    }
    
    \caption{\revv{Heatmaps showing inter-threshold Pearson correlation coefficients on (a) In-distribution (\texttt{COCO}), (b) Near OOD (\texttt{Cityscapes}), and (c) Far OOD (\texttt{Foggy Cityscapes}). Results indicate that medium thresholds are highly redundant, whereas divergence occurs at high $\tau$.}}

    \label{fig:oce_iou_abl}
\end{figure*}

\begin{table*}[t!]
\centering
\caption{
Comparison of post-processing schemes based on calibration quality (i.e., OCE). We evaluated three standard methods: confidence thresholding, top-$k$, and NMS (without confidence thresholding). For each scheme, the optimal hyperparameter selected from the validation set is shown in parentheses, and is applied on the test set.
the strongest correlations are highlighted in bold. Confidence thresholding achieves the lowest OCE, demonstrating its better efficacy compared to the other schemes.
The optimal thresholds are approximately $0.3$, aligning well with values commonly employed in previous studies.
}
\renewcommand{\arraystretch}{1.0}
\resizebox{1.0\linewidth}{!}{
\begin{tabular}{l|cccc|cccc|cccc}
\toprule
\multirow{2}{*}{\textbf{Methods}} & \multicolumn{4}{c}{\texttt{\textbf{COCO (in-distribution)}}} & \multicolumn{4}{c}{\textbf{\texttt{Cityscapes  (near OOD)}}} & \multicolumn{4}{c}{\textbf{\texttt{Foggy Cityscapes (OOD)}}}  \\
\cmidrule(lr){2-5} \cmidrule(lr){6-9} \cmidrule(lr){10-13}
& \textbf{UP-DETR} & \textbf{D-DETR} & \textbf{Cal-DETR} & \textbf{DINO} & \textbf{UP-DETR} & \textbf{D-DETR} & \textbf{Cal-DETR} & \textbf{DINO} & \textbf{UP-DETR} & \textbf{D-DETR} & \textbf{Cal-DETR} & \textbf{DINO} \\
\midrule
\midrule
Thresholding & \bf 0.276 (0.35) & \bf 0.450 (0.20) & \bf 0.416 (0.25) & \bf 0.436 (0.25) & \bf 0.313 (0.45) & \bf 0.430 (0.30) & \bf 0.413 (0.30) & \bf 0.402 (0.35) & \bf 0.387 (0.40) & \bf 0.488 (0.30) & \bf 0.469 (0.30) & \bf 0.458 (0.30) \\
Top-$k$ &  0.358 (20.00) &  0.485 (20.00) &  0.457 (20.00) &  0.479 (20.00) &  0.352 (20.00) &  0.451 (20.00) &  0.436 (20.00) &  0.422 (20.00) &  0.416 (20.00) &  0.500 (20.00) &  0.490 (20.00) &  0.473 (20.00) \\
NMS &  0.358 (0.90) &  0.535 (0.90) &  0.624 (0.90) &  0.521 (0.90) &  0.426 (0.90) &  0.625 (0.90) &  0.784 (0.90) &  0.583 (0.90) &  0.495 (0.90) &  0.650 (0.90) &  0.794 (0.90) &  0.609 (0.90) \\
\bottomrule
\end{tabular}
}
\label{tab:oce_separation_all}
\end{table*}

\begin{table*}[t!]
\centering
\small
\caption{
Comparison of post-processing schemes and baseline methods based on their Pearson correlation coefficients with image-level reliability when $\approachpos$ and $\approach(\lambda=1.0)$ are applied, respectively. We evaluated three standard methods: confidence thresholding, top-$k$, and NMS (without confidence thresholding). For each scheme, the optimal hyperparameter selected from the validation set is shown in parentheses and is applied on the test set.
The strongest correlations are highlighted in bold. Confidence thresholding achieves the lowest OCE, demonstrating its better efficacy compared to the other schemes.
For Deformable-DETR, Cal-DETR, and DINO, the optimal thresholds are approximately $0.3$, aligning well with values commonly employed in previous studies.
For UP-DETR, the optimal thresholds often exceed $0.5$, highlighting the potential limitation of using a fixed threshold for image-level UQ.
}
\renewcommand{\arraystretch}{1.0}
\resizebox{1.0\linewidth}{!}{
\begin{tabular}{ll|cccc|cccc|cccc}
\toprule
\multirow{2}{*}{\textbf{UQ}} & \multirow{2}{*}{\textbf{Methods}} & \multicolumn{4}{c}{\texttt{\textbf{COCO (in-distribution)}}} & \multicolumn{4}{c}{\textbf{\texttt{Cityscapes  (near OOD)}}} & \multicolumn{4}{c}{\textbf{\texttt{Foggy Cityscapes (OOD)}}}  \\
\cmidrule(lr){3-6} \cmidrule(lr){7-10} \cmidrule(lr){11-14}
& & \textbf{UP-DETR} & \textbf{D-DETR} & \textbf{Cal-DETR} & \textbf{DINO} & \textbf{UP-DETR} & \textbf{D-DETR} & \textbf{Cal-DETR} & \textbf{DINO} & \textbf{UP-DETR} & \textbf{D-DETR} & \textbf{Cal-DETR} & \textbf{DINO} \\
\midrule
\midrule
\multirow{3}{*}{$\approachpos$} & Thresholding & \bf 0.479 (0.25) & \bf 0.478 (0.20) & \bf 0.547 (0.25) & \bf 0.512 (0.25) & \bf 0.333 (0.80) & \bf 0.355 (0.25) & \bf 0.411 (0.20) & \bf 0.403 (0.35) & \bf 0.229 (0.50) & \bf 0.291 (0.25) & \bf 0.281 (0.15) & \bf 0.282 (0.35) \\
 & Top-$k$ &  0.067 (1.00) &  0.132 (1.00) &  0.252 (1.00) &  0.155 (1.00) &  \textcolor{red}{-0.037 (2.00)} &  0.019 (1.00) &  0.078 (1.00) &  0.021 (1.00) &  0.080 (1.00) &  \textcolor{red}{-0.006 (1.00)} &  0.026 (1.00) &  0.090 (1.00) \\
 & NMS &  \textcolor{red}{-0.524 (0.10)} &  \textcolor{red}{-0.508 (0.10)} &  \textcolor{red}{-0.488 (0.10)} &  \textcolor{red}{-0.510 (0.10)} &  \textcolor{red}{-0.370 (0.10)} &  \textcolor{red}{-0.324 (0.10)} &  \textcolor{red}{-0.273 (0.10)} &  \textcolor{red}{-0.338 (0.10)} &  \textcolor{red}{-0.205 (0.10)} &  \textcolor{red}{-0.251 (0.10)} &  \textcolor{red}{-0.202 (0.15)} &  \textcolor{red}{-0.204 (0.10)} \\
\midrule
\multirow{3}{*}{$\approach$} & Thresholding &  0.597 (0.55) & \bf 0.513 (0.25) & \bf 0.563 (0.25) & \bf 0.540 (0.25) &  0.408 (0.80) & \bf 0.389 (0.30) & \bf 0.422 (0.20) & \bf 0.437 (0.35) & \bf 0.312 (0.75) & \bf 0.317 (0.25) & \bf 0.288 (0.15) & \bf 0.311 (0.35) \\
 & Top-$k$ & \bf 0.629 (1.00) &  0.268 (1.00) &  0.320 (1.00) &  0.290 (1.00) & \bf 0.409 (1.00) &  0.169 (1.00) &  0.151 (1.00) &  0.220 (1.00) &  0.284 (1.00) &  0.077 (1.00) &  0.060 (1.00) &  0.174 (1.00) \\
 & NMS &  \textcolor{red}{-0.227 (0.10)} &  \textcolor{red}{-0.157 (0.90)} &  \textcolor{red}{-0.402 (0.10)} &  \textcolor{red}{-0.140 (0.90)} &  \textcolor{red}{-0.044 (0.80)} &  \textcolor{red}{-0.034 (0.75)} &  \textcolor{red}{-0.203 (0.10)} &  \textcolor{red}{-0.080 (0.80)} &  0.035 (0.75) &  \textcolor{red}{-0.035 (0.75)} &  \textcolor{red}{-0.149 (0.90)} &  \textcolor{red}{-0.075 (0.70)} \\
\bottomrule
\end{tabular}
}
\label{tab:imreli_separation_all}
\end{table*}

\section{Comparative Study on Post-Processing} \label{sec:postprocessing}

In this section, we provide a comprehensive analysis regarding the impact of post-processing on the model's overall calibration quality and UQ performance.
For the post-processing (i.e., separation) methods, we compare the following approaches: (1) applying a threshold on the confidence score, (2) selecting the Top-$k$ predictions, and (3) utilizing NMS.

Primarily, we evaluate OCE by varying the hyperparameter for each method (i.e., confidence threshold, $k$, and IoU threshold, respectively) and compare the resulting performance.
In this first analysis, to exclude the dependency of the final performance on OCE, we choose the best hyperparameter on the validation set for each setting and compare their best possible performances.
Table~\ref{tab:oce_separation_all} shows that the confidence thresholding approach outperforms the top-$k$ and NMS approaches.
The top-$k$ approach underperforms because the number of objects in an image varies significantly, making it impractical to determine a single optimal value for $k$.
Top-$k$ becomes optimal when $k=20$.
Given that the average and 95th percentile number of objects per image in the \texttt{COCO} dataset are 7 and 22, respectively, these results appear reasonable. 
Therefore, we confirm that using an excessively large number (e.g., 100) for top-$k$ is inadequate for achieving well-calibrated predictions.
On the other hand, with NMS (when applied without preceding confidence thresholding), we empirically observe that although most of the optimal positive predictions are included, a substantial number of negative predictions are retained, often outnumbering the positive ones by several times.

Similarly, we compare the image-level uncertainty quantification performance, and the results are shown in Table~\ref{tab:imreli_separation_all}.
In this experiment, we fixed the scaling factor $\lambda$ at $1.0$ to minimize its influence on selecting the optimal hyperparameter.
We reconfirm the effectiveness of confidence thresholding; for Deformable-DETR, Cal-DETR, and DINO, the optimal thresholds are approximately $0.3$, aligning well with values commonly employed in previous studies.
For UP-DETR, the optimal thresholds exceed $0.5$, highlighting the potential limitation of using a fixed threshold in image-level UQ applications.
Moreover, the correlation is often negative when NMS is applied along with the $\approachpos$ framework.
This is because, while most of the optimal positive queries are likely to be included with NMS, a substantial number of optimal negative queries remain within the final subset, often outnumbering the positive ones by several times.
As a result, applying NMS leads to an inaccurate reliability assessment.
However, many schemes, including NMS, achieve significant improvement when used with the proposed contrastive framework; even if the post-processing scheme is inaccurate, the framework can robustly perform by leveraging the negative predictions. 

\begin{figure*}[t]
    \centering
    \subfloat[Thresholding (UP-DETR)]{%
        \includegraphics[width=0.33\textwidth]{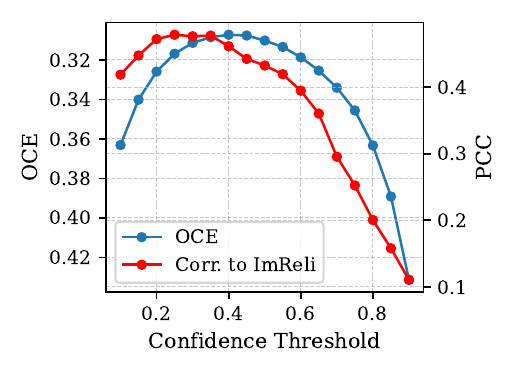}
    }
    \subfloat[Top-$k$ (UP-DETR)]{%
        \includegraphics[width=0.33\textwidth]{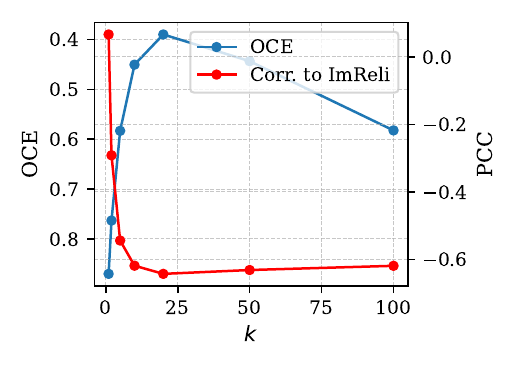}
    }
    \subfloat[NMS (UP-DETR)]{%
        \includegraphics[width=0.33\textwidth]{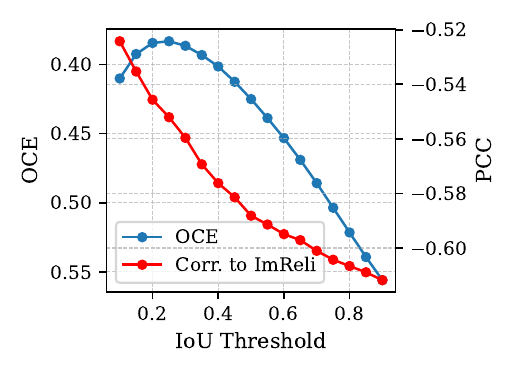}
    }
    
    \vspace{1em}
    
    \subfloat[Thresholding (D-DETR)]{%
        \includegraphics[width=0.33\textwidth]{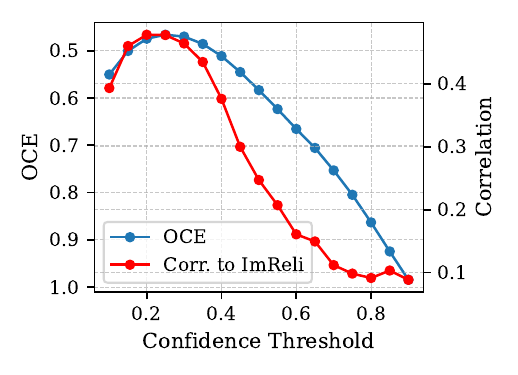}
    }
    \subfloat[Top-$k$ (D-DETR)]{%
        \includegraphics[width=0.33\textwidth]{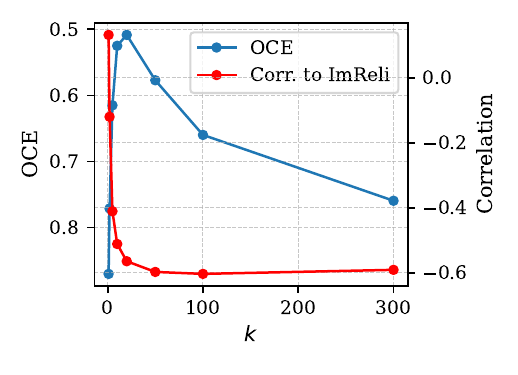}
    }
    \subfloat[NMS (D-DETR)]{%
        \includegraphics[width=0.33\textwidth]{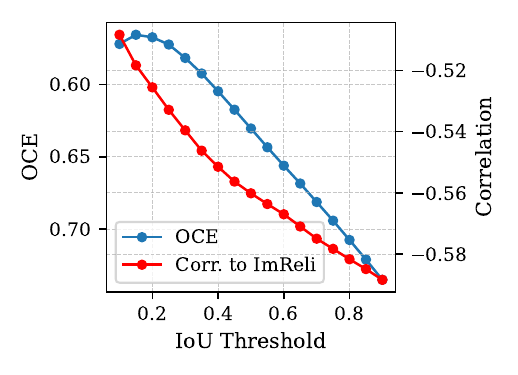}
    }
    
    \vspace{1em}
    
    \subfloat[Thresholding (Cal-DETR)]{%
        \includegraphics[width=0.33\textwidth]{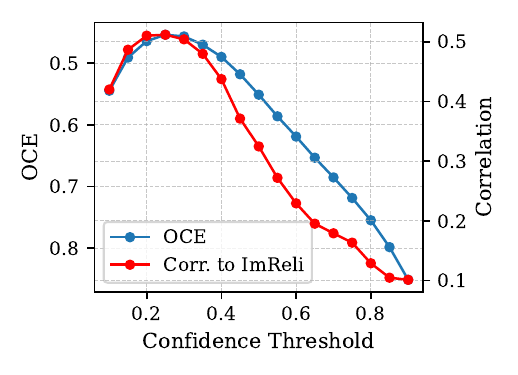}
    }
    \subfloat[Top-$k$ (Cal-DETR)]{%
        \includegraphics[width=0.33\textwidth]{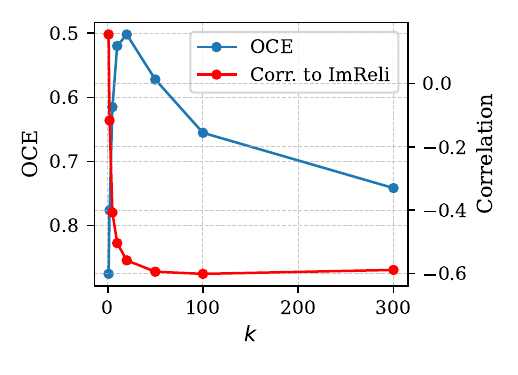}
    }
    \subfloat[NMS (Cal-DETR)]{%
        \includegraphics[width=0.33\textwidth]{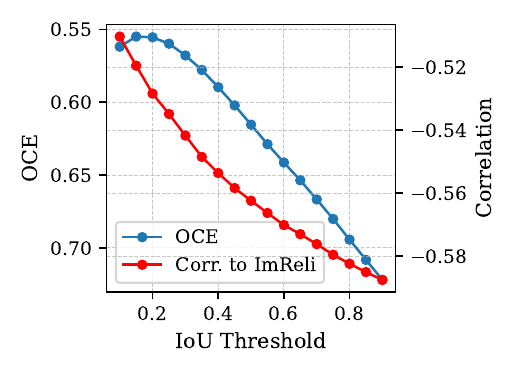}
    }
    
    \vspace{1em}
    
    \subfloat[Thresholding (DINO)]{%
        \includegraphics[width=0.33\textwidth]{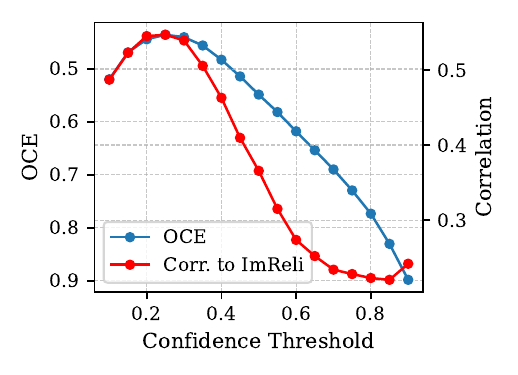}
    }
    \subfloat[Top-$k$ (DINO)]{%
        \includegraphics[width=0.33\textwidth]{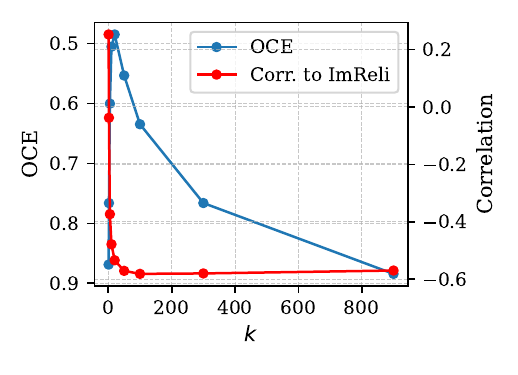}
    }
    \subfloat[NMS (DINO)]{%
        \includegraphics[width=0.33\textwidth]{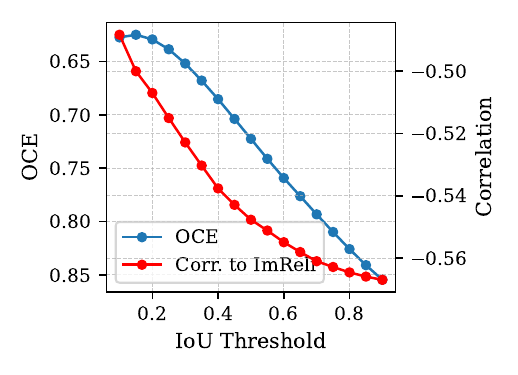}
    }
    \caption{Impact of parameter selection on OCE (y-axis inverted) and the Pearson correlation coefficient (PCC) between $\approachpos$ and image-level reliability across different DETR models on \texttt{COCO} for different post-processing schemes.}
    \label{fig:separations_all}
\end{figure*}

\begin{figure*}[t!]
    \centering
    \subfloat[Thresholding (0.1)]{%
        \includegraphics[width=0.27\textwidth]{app_figures/sample3512_thr_0.1.pdf}
    }
    \subfloat[Thresholding (0.27)]{%
        \includegraphics[width=0.27\textwidth]{app_figures/sample3512_thr_0.3.pdf}
    }
    \subfloat[Thresholding (0.9)]{%
        \includegraphics[width=0.27\textwidth]{app_figures/sample3512_thr_0.9.pdf}
    }
    
    \subfloat[Top-$1$]{%
        \includegraphics[width=0.27\textwidth]{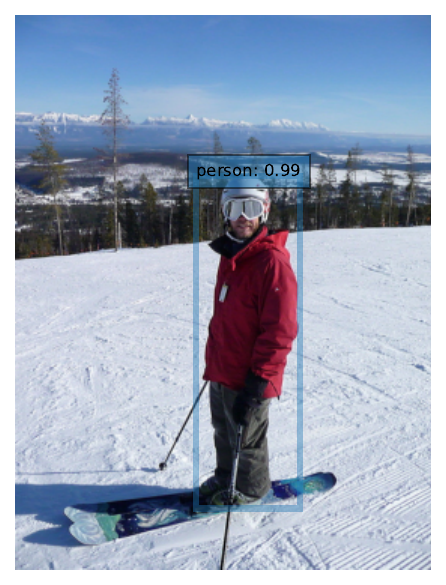}
    }
    \subfloat[Top-$20$]{%
        \includegraphics[width=0.27\textwidth]{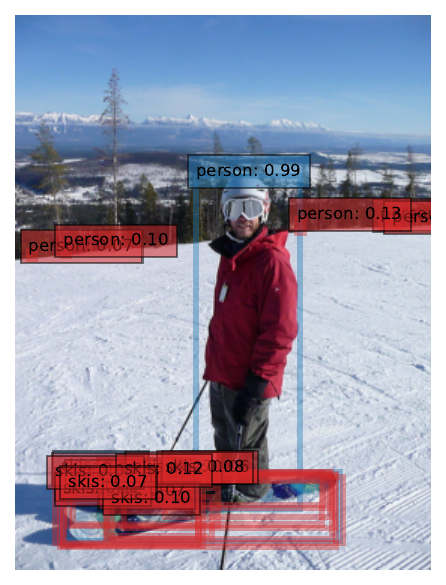}
    }
    \subfloat[Top-$100$]{%
        \includegraphics[width=0.27\textwidth]{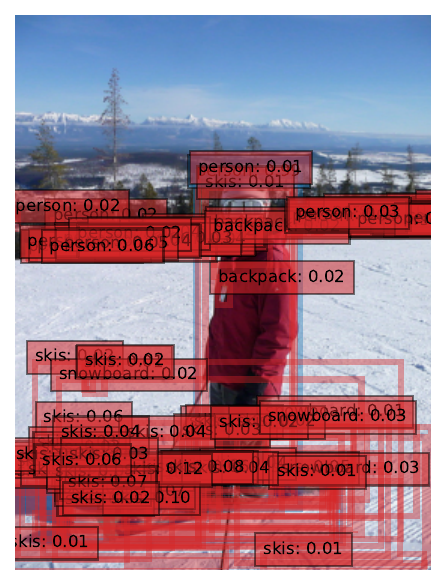}
    }
    
    \subfloat[NMS ($\tau=0.1$)]{%
        \includegraphics[width=0.27\textwidth]{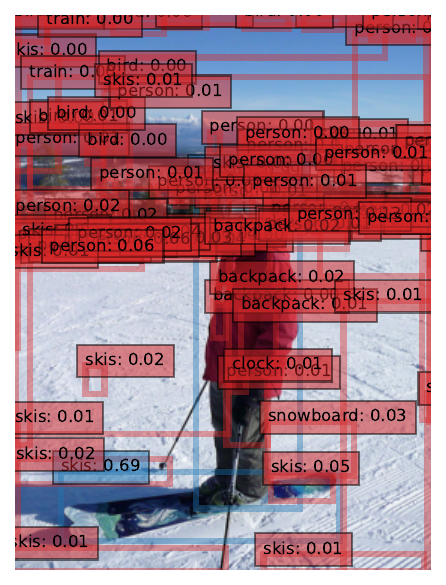}
    }
    \subfloat[NMS ($\tau=0.5$)]{%
        \includegraphics[width=0.27\textwidth]{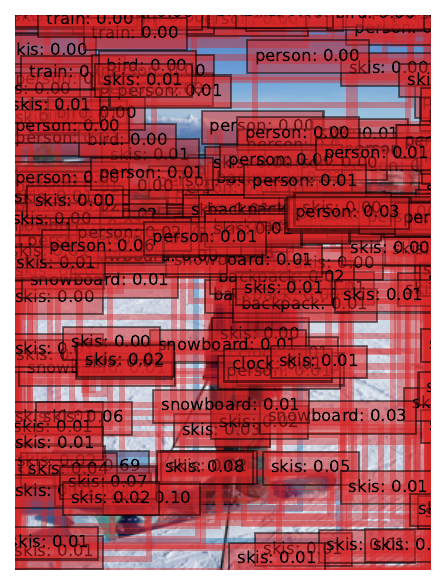}
    }
    \subfloat[NMS ($\tau=0.9$)]{%
        \includegraphics[width=0.27\textwidth]{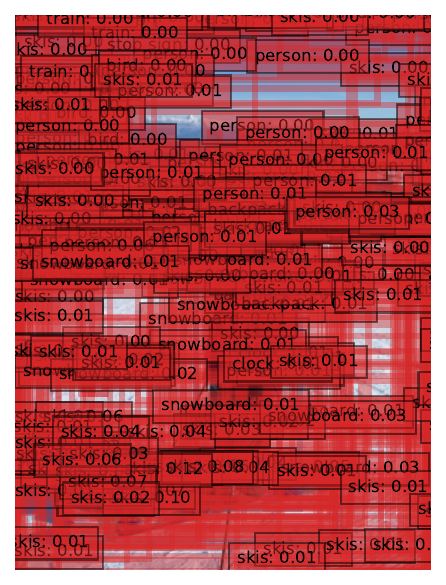}
    }
    \caption{Exemplary visualization demonstrating the impact of parameter selection on the final subset of predictions in Cal-DETR for different post-processing schemes. Optimal positive and negative predictions are highlighted with green and red boxes, respectively. As shown, the top-$k$ and NMS approaches could include too many negative predictions, degrading the calibration quality. Confidence thresholding with too low of a threshold faces a similar issue, while too high of a threshold risks omitting positive predictions. Therefore, accurately identifying a reliable set of predictions significantly affects the reliability of DETR for downstream applications.}
    \label{fig:separation_visualization}
\end{figure*}

Lastly, we plot the OCE and image-level UQ performance with different hyperparameter selections to show the sensitivity of each scheme to the choice of hyperparameter.
Figure~\ref{fig:separations_all} highlight that \emph{carefully identifying a reliable subset is crucial for achieving both high object-level calibration quality and effective image-level uncertainty quantification performance}.
We also present this with exemplary visualizations in Figure~\ref{fig:separation_visualization}.

\clearpage

\end{document}